\documentclass[final,5p,times]{elsarticle} 

\usepackage{graphicx}
\usepackage{subfig}
\usepackage{float}
\usepackage{multirow}
\usepackage{multicol}
\usepackage[hidelinks]{hyperref}
\usepackage[utf8]{inputenc}

\usepackage{lineno}
\modulolinenumbers[5]

\usepackage{hyphenat}
\journal{Expert Systems with Applications}

\biboptions{authoryear}

\begin{document}

\begin{frontmatter}

\title{Self-Driving Cars: A Survey}

\author[ufes]{Claudine Badue\corref{cor1}}
\author[ufes]{R\^anik Guidolini}
\author[ufes]{Raphael Vivacqua Carneiro}
\author[ufes]{Pedro Azevedo}
\author[ufes]{Vinicius Brito Cardoso}
\author[ifes1]{Avelino Forechi}
\author[ufes]{Luan Jesus}
\author[ufes]{Rodrigo Berriel}
\author[ifes2]{Thiago Paix\~ao}
\author[ifes2]{Filipe Mutz}
\author[ufes]{Lucas Veronese}
\author[ufes]{Thiago Oliveira-Santos}
\author[ufes]{Alberto Ferreira De Souza}

\cortext[cor1]{Corresponding Author: Claudine Badue, Departamento de Inform\'atica, Universidade Federal do Esp\'irito Santo - Campus Vit\'oria, Av. Fernando Ferrari 514, 29075-910, Goiabeiras, Vit\'oria, Esp\'irito Santo, Brazil; Email: claudine@lcad.inf.ufes.br, claudinebadue@gmail.com}

\address[ufes]{Departamento de Inform\'atica, Universidade Federal do Esp\'irito Santo, Av. Fernando Ferrari 514, 29075-910, Goiabeiras, Vit\'oria, Esp\'irito Santo, Brazil}

\address[ifes1]{Coordenadoria de Engenharia Mec\^anica, Instituto Federal do Esp\'irito Santo, Av. Morob\'a 248, 29192–733, Morob\'a, Aracruz, Esp\'irito Santo, Brazil}

\address[ifes2]{Coordenac\~ao de Inform\'atica, Instituto Federal do Esp\'irito Santo, ES-010 Km-6.5, 29173-087, Manguinhos, Serra, Esp\'irito Santo, Brazil}

\begin{abstract}

We survey research on self-driving cars published in the literature focusing on autonomous cars developed since the DARPA challenges, which are equipped with an autonomy system that can be categorized as SAE level 3 or higher. The architecture of the autonomy system of self-driving cars is typically organized into the perception system and the decision-making system. The perception system is generally divided into many subsystems responsible for tasks such as self-driving-car localization, static obstacles mapping, moving obstacles detection and tracking, road mapping, traffic signalization detection and recognition, among others. The decision-making system is commonly partitioned as well into many subsystems responsible for tasks such as route planning, path planning, behavior selection, motion planning, and control. In this survey, we present the typical architecture of the autonomy system of self-driving cars. We also review research on relevant methods for perception and decision making. Furthermore, we present a detailed description of the architecture of the autonomy system of the self-driving car developed at the Universidade Federal do Espírito Santo (UFES), named Intelligent Autonomous Robotics Automobile (IARA). Finally, we list prominent self-driving car research platforms developed by academia and technology companies, and reported in the media.

\end{abstract}


\end{frontmatter}


\tableofcontents

\section{Introduction}

Self-driving cars (also known as autonomous cars and driverless cars) have been studied and developed by many universities, research centers, car companies, and companies of other industries around the world since the middle 1980s. Important examples of self-driving car research platforms in the last two decades are the Navlab’s mobile platform \citep{item2}, University of Pavia’s and University of Parma's car, ARGO \citep{item4}, and UBM's vehicles, VaMoRs and VaMP \citep{item5}. 

In order to spur technology for the development of self-driving cars, the Defense Advanced Research Projects Agency (DARPA) organized three competitions in the last decade. The first, named DARPA Grand Challenge, was realized at the Mojave Desert, USA, in 2004, and required self-driving cars to navigate a 142 miles long course throughout desert trails within a 10 hour time limit. All competing cars failed within the first few miles. 

The DARPA Grand Challenge \citep{item6} was repeated in 2005 and required self-driving cars to navigate a 132 miles long route through flats, dry lake beds, and mountain passes, including three narrow tunnels and more than 100 sharp left and right turns. This competition had 23 finalists and 4 cars completed the route within the allotted time limit. The Stanford University's car, Stanley \citep{item7}, claimed first place, and the Carnegie Mellon University's cars, Sandstorm and H1ghlander, finished in second and third places, respectively. 

The third competition, known as the DARPA Urban Challenge \citep{item8}, was held at the former George Air Force Base, California, USA, in 2007, and required self-driving cars to navigate a 60 miles long route throughout a simulated urban environment, together with other self-driving and human driven cars, within a 6 hour time limit. The cars had to obey California traffic rules. This competition had 11 finalists and 6 cars completed the route within the allotted time limit. The Carnegie Mellon University's car, Boss \citep{item9}, claimed first place, the Stanford University's car, Junior \citep{item10}, finished in second, and the Virginia Tech's car, Odin \citep{item11}, came in third place. Even though these competitions presented challenges much simpler than those typically seen in everyday traffic, they have being hailed as milestones for the development of self-driving cars.

Since the DARPA challenges, many self-driving car competitions and trials have been performed. Relevant examples include: the European Land$-$Robot Trial (ELROB) \citep{item12}, which has being held from 2006 to the current year; the Intelligent Vehicle Future Challenge \citep{item13}, from 2009 to 2013; the Autonomous Vehicle Competition, from 2009 to 2017 \citep{item14}; the Hyundai Autonomous Challenge, in 2010 \citep{item15a}; the VisLab Intercontinental Autonomous Challenge, in 2010 \citep{item15b}; the Grand Cooperative Driving Challenge (GCDC) \citep{item16}, in 2011 and 2016; and the Proud-Public Road Urban Driverless Car Test, in 2013 \citep{item287}. At the same time, research on self-driving cars has accelerated in both academia and industry around the world. Notable examples of universities conducting research on self-driving cars comprise Stanford University, Carnegie Mellon University, MIT, Virginia Tech, FZI Research Center for Information Technology, and University of Ulm. Notable examples of companies include Google, Uber, Baidu, Lyft, Aptiv, Tesla, Nvidia, Aurora, Zenuity, Daimler and Bosch, Argo AI, Renesas Autonomy, Almotive, AutoX, Mobileye, Ambarella, Pony.ai, Idriverplus, Toyota, Ford, Volvo, and Mercedes Benz.

Although most of the university research on self-driving cars has been conducted in the United States of America, Europe and Asia, some relevant investigations have been carried out in China, Brazil and other countries. Relevant examples of self-driving car research platforms in Brazil are the Universidade Federal de Minas Gerais (UFMG)'s car, CADU \citep{item46, item47, item48, item49}, Universidade de São Paulo's car, CARINA \citep{item50, item51, item52, item53}, and the Universidade Federal do Espírito Santo (UFES)'s car, IARA \citep{item54, item55, item56, item57}. IARA was the first Brazilian self-driving car to travel autonomously tenths of kilometers on urban roads and highways. 

To gauge the level of autonomy of self-driving cars, the Society of Automotive Engineers (SAE) International published a classification system based on the amount of human driver intervention and attentiveness required by them, in which the level of autonomy of a self-driving car may range from level 0 (the car's autonomy system issues warnings and may momentarily intervene but has no sustained car control) to level 5 (no human intervention is required in any circumstance) \citep{item58}. In this paper, we survey research on self-driving cars published in the literature focusing on self-driving cars developed since the DARPA challenges, which are equipped with an autonomy system that can be categorized as SAE level 3 or higher \citep{item58}.

The architecture of the autonomy system of self-driving cars is typically organized into two main parts: the perception system, and the decision-making system \citep{item59}. The perception system is generally divided into many subsystems responsible for tasks such as autonomous car localization, static obstacles mapping, road mapping, moving obstacles detection and tracking, traffic signalization detection and recognition, a-mong others. The decision-making system is commonly partitioned as well into many subsystems responsible for tasks such as route planning, path planning, behavior selection, motion planning, obstacle avoidance and control, though this partitioning is somewhat blurred and there are several different variations in the literature \citep{item59}.

In this survey, we present the typical architecture of the autonomy system of self-driving cars. We also review research on relevant methods for perception and decision making. 

The remainder of this paper is structured as follows. In Section 2, we present an overview of the typical architecture of the autonomy system of self-driving cars, commenting on the responsibilities of the perception system, decision making system, and their subsystems. In Section 3, we present research on important methods for the perception system, including autonomous car localization, static obstacles mapping, road mapping, moving obstacles detection and tracking, traffic signalization detection and recognition. In Section 4, we present research on relevant techniques for the decision-making system, comprising the route planning, path planning, behavior selection, motion planning, obstacle avoidance and control. In Section 5, we present a detailed description of the architecture of the autonomy system of the UFES's car, IARA. Finally, in Section 6, we list prominent self-driving car research platforms developed by academia and technology companies, and reported in the media.

\section{Typical Architecture of Self-Driving Cars}

In this section, we present an overview of the typical architecture of the automation system of self-driving cars and comment on the responsibilities of the perception system, decision making system, and their subsystems.

Figure 1 shows a block diagram of the typical architecture of the automation system of self-driving cars, where the Perception and Decision Making systems \citep{item59} are shown as a collection of subsystems of different colors. The \textbf{Perception} system is responsible for estimating the State of the car and for creating an internal (to the self-driving system) representation of the environment, using data captured by on-board sensors, such as Light Detection and Ranging (LIDAR), Radio Detection and Ranging (RADAR), camera, Global Positioning System (GPS), Inertial Measurement Unit (IMU), odometer, etc., and prior information about the sensors’ models, road network, traffic rules, car dynamics, etc. The \textbf{Decision Making} system is responsible for navigating the car from its initial position to the final goal defined by the user, considering the current car’s State and the internal representation of the environment, as well as traffic rules and passengers’ safety and comfort.

\begin{figure}[htb!]
\centering
\includegraphics[width=0.4\textwidth]{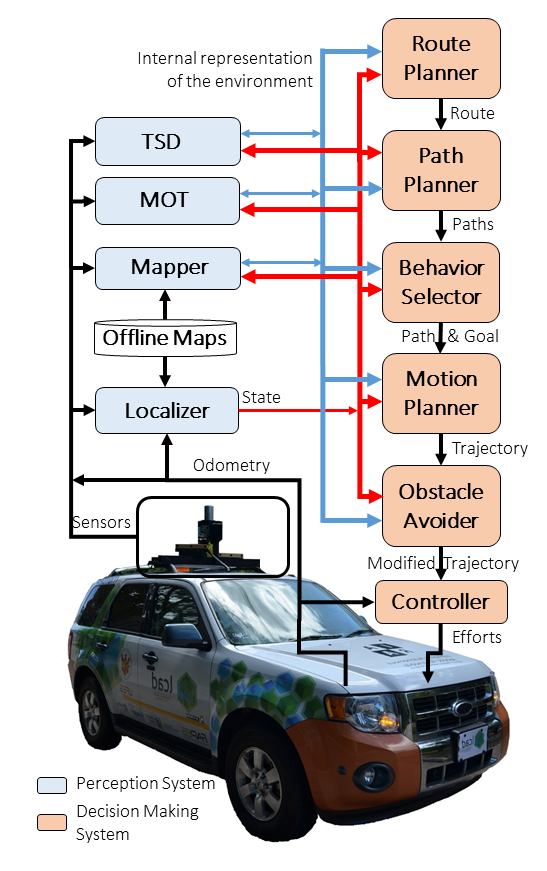}
\caption{Overview of the typical hierarchical architecture of self-driving cars. TSD denotes Traffic Signalization Detection and MOT, Moving Objects Tracking.}
\end{figure}

In order  to navigate the car throughout the environment, the Decision Making system needs to know where the self-driving car is in it. The \textbf{Localizer} subsystem (Figure 1) is responsible for estimating the car’s State (pose, linear velocities, angular velocities, etc.) in relation to static maps of the environment (see Section 3.1). These static maps, or \textbf{Offline Maps} (Figure 1), are computed automatically before the autonomous operation, typically using the sensors of the self-driving car itself, although manual annotations (i.e., the position of pedestrian crosswalks or of traffic lights) or editions (for removing non-static objects captured by the sensors) are usually required. A self-driving car may use for localization one or more Offline Maps, such as occupancy grid maps, remission maps or landmark maps. We survey the literature on methods for generating Offline Maps in Section 3.2.

Information regarding the rules and regulations on how the self-driving car may move in the roads and highways (direction of traffic, maximum speed, lane demarcation, etc.) are also essential to the Decision Making system. This information is typically embedded in road maps, which represent it in such maps using geometrical and topological properties. We survey the literature on road mapping in Section 3.3.

The Localizer subsystem receives as input the Offline Maps, Sensors’ data and the Odometry of the self-driving car, and computes as output the car’s State (Figure 1). It is important to note that, although GPS may help the localization process, it alone is not enough for proper localization in urban environments due to interferences caused by tall trees, buildings, tunnels, etc., that makes GPS positioning unreliable. We survey the literature on localization techniques in Section 3.1.

The \textbf{Mapper} subsystem receives as input the Offline Maps and the self-driving car’s State, and generates as output the online map. This online map is typically a merge of information present in the Offline Maps and an occupancy grid map computed online using sensors’ data and the current car’s State. We survey the literature on methods for computing the online map in Section 3.2. It is desirable that the online map contains only a static representation of the environment, since this may help the operation of some subsystems of the Decision Making system. 

Information regarding the pose and velocity of moving obstacles is also essential to the Decision Making system. This information enables this system to make decisions that avoid collision with moving obstacles. It also allows the removal of moving obstacles from the online map. The \textbf{Moving Objects Tracker} subsystem, or \textbf{MOT }(Figure 1), receives the Offline Maps and the self-driving car’s State, and detects and tracks, i.e., calculates the pose and velocity of, the nearest moving obstacles (e.g., other vehicles and pedestrians). We survey the literature on methods for moving objects detection and tracking in the context of self-driving cars in Section 3.4.

Horizontal (i.e. lane markings) and vertical (i.e. speed limits, traffic lights, etc.) traffic signalization must be recognized and obeyed by self-driving cars. The \textbf{Traffic Signalization Detector} subsystem, or \textbf{TSD} (Figure 1), is responsible for the detection and recognition of traffic signalization. It receives the Sensors’ data and the car’s State, and detects the position of traffic signalizations and recognizes their class or status. We survey the literature on methods for traffic signalization detection and recognition in Section 3.5.

Given a Final Goal defined in the Offline Maps by the user, the \textbf{Route Planner} subsystem computes a Route,  \( W \), in the Offline Maps, from the current self-driving car’s State to the Final Goal. A Route is a sequence of way points, i.e.  \( W= \{ w_{1}, w_{2}, \ldots, w_{ \vert W \vert } \}  \), where each way point,  \( w_{i} \), is a coordinate pair, i.e.  \( w_{i} = \left( x_{i}, y_{i} \right)  \), in the Offline Maps. We survey the literature on methods for route planning in Section 4.1.

Given a Route, the \textbf{Path Planner} subsystem computes, considering the current self-driving car’s State and the internal representation of the environment as well as traffic rules, a set of Paths,  \( P= \{ P_{1}, P_{2}, \ldots,P_{ \vert P \vert } \)  $ \} $. A Path is a sequence of poses, i.e.  \( P_{j}= \{ p_{1}, p_{2}, \ldots,p_{ \vert P \vert } \}  \), where each pose,  \( p_{i} \), is a coordinate pair in the Offline Maps and the desired car’s orientation at the position defined by this coordinate pair, i.e.  \( p_{i}= \left( x_{i}, y_{i}, \theta _{i} \right)  \). We survey the literature on methods for path planning in Section 4.2.

The \textbf{Behavior Selector} subsystem is responsible for choosing the current driving behavior, such as lane keeping, intersection handling, traffic light handling, etc. It does so by selecting a Path,  \( P_{j} \), in  \( P \), a pose, \(  p_{g} \), in  \( P_{j} \)  a few seconds ahead of the current self-driving car’s State (about 5 s), which is the decision horizon, and the desired velocity at this pose. The pair pose in  \( P_{j} \)  and associated velocity is called  \( Goal_{g}= \left( p_{g}, v_{g} \right)  \). The Behavior Selector chooses a Goal considering the current driving behavior and avoiding collisions with static and moving obstacles in the environment within the decision horizon time frame.

The \textbf{Motion Planner} subsystem is responsible for computing a Trajectory,  \( T \), from the current self-driving car’s State to the current Goal, which follows the Path defined by the Behavior Selector, satisfies car’s kinematic and dynamic constraints, and provides comfort to the passengers. A Trajectory  \( T= \{ c_{1},c_{2}, \ldots,c_{ \vert T \vert } \}  \)  may be defined as a sequence of commands,  \( c_{k}= \left( v_{k}, \varphi _{k}, \Delta t_{k} \right)  \), where  \( v_{k} \)  is the desired velocity at time  \( k \),  \(  \varphi _{k} \)  is the desired steering angle at time  \( k \), and  \(  \Delta t_{k} \)  is the duration of  \( c_{k} \); there are other forms of defining trajectories, however (see Section 4.4). A Trajectory takes the car from its current State to the current Goal smoothly and safely. We survey the literature on methods for motion planning in Section 4.4.

The \textbf{Obstacle Avoider} subsystem receives the Trajectory computed by the Motion Planner and changes it (typically reducing the velocity), if necessary, to avoid collisions. There is no much literature on methods for performing the functions of the Obstacle Avoider subsystem. We discuss some relevant literature on this subject in Section 4.5. 

Finally, the \textbf{Controller} subsystem receives the Motion Planner trajectory, eventually modified by the Obstacle Avoider subsystem, and computes and sends Effort commands to the actuators of the steering wheel, throttle and brakes in order to make the car execute the Modified Trajectory as best as the physical world allows. We survey the literature on methods for low level car control in Section 4.6.

In the following, we detail each one of these subsystems and the techniques used to implement them and their variants, grouped within perception and decision-making systems.

\section{Self-Driving Cars’ Perception}

In this section, we present research on important methods proposed in the literature for the perception system of self-driving cars, including methods for localization, offline obstacle mapping, road mapping, moving obstacle tracking, and traffic signalization detection and recognition.

\subsection{Localization}

The \textbf{Localizer} subsystem is responsible for estimating the self-driving car pose (position and orientation) relative to a map or road (e.g., represented by curbs or road marks). Most general-purpose localization subsystems are based on GPS. However, by and large, they are not applicable to urban self-driving cars, because the GPS signal cannot be guaranteed in occluded areas, such as under trees, in urban canyons (roads surrounded by large buildings) or in tunnels. 

Various localization methods that do not depend on GPS have been proposed in the literature. They can be mainly categorized into three classes: LIDAR-based, LIDAR plus camera-based, and camera-based. LIDAR-based localization methods rely solely on LIDAR sensors, which offer measurement accuracy and easiness of processing. However, despite the LIDAR industry efforts to reduce production costs, they still have a high price if compared with cameras. In typical LIDAR plus camera-based localization methods, LIDAR data is used only to build the map, and camera data is employed to estimate the self-driving car’s position relative to the map, which reduces costs. Camera-based localization approaches are cheap and convenient, even though typically less precise and/or reliable.

\subsubsection{LIDAR-Based Localization}

\citet{item60} proposed a localization method that uses offline grid maps of reflectance intensity distribution of the environment measured by a LIDAR scanner (remission grid maps, Figure 2); they have used the Velodyne HDL-64E LIDAR in their work. An unsupervised calibration method is used to calibrate the Velodyne HDL-64E’ laser beams so that they all respond similarly to the objects with the same brightness, as seen by the LIDAR. A 2-dimension histogram filter \citep{item61} is employed to estimate the autonomous vehicle position. As usual, the filter is comprised of two parts: the motion update (or prediction), to estimate the car position based on its motion, and the measurement update (or correction), to increase confidence in our estimate based on sensor data. In the motion update, the car motion is modeled as a random walk with Gaussian noise drift from a dead reckoning coordinate system (computed using the inertial update of an Applanix POS LV-420 position and orientation system) to the global coordinate system of the offline map. In the measurement step, they use, for different displacements, the similarity between the remission map computed online, with the remission map computed offline. Each displacement corresponds to one cell of the histogram of the histogram filter. To summarize the histogram into a single pose estimate, they use the center of mass of the probability distribution modeled by the histogram. The authors do not describe how they estimate the orientation, though. Their method has shown a Root Mean Squared (RMS) lateral error of 9 cm and a RMS longitudinal error of 12 cm.

\begin{figure}[htb!]
\centering
\includegraphics[width=0.4\textwidth]{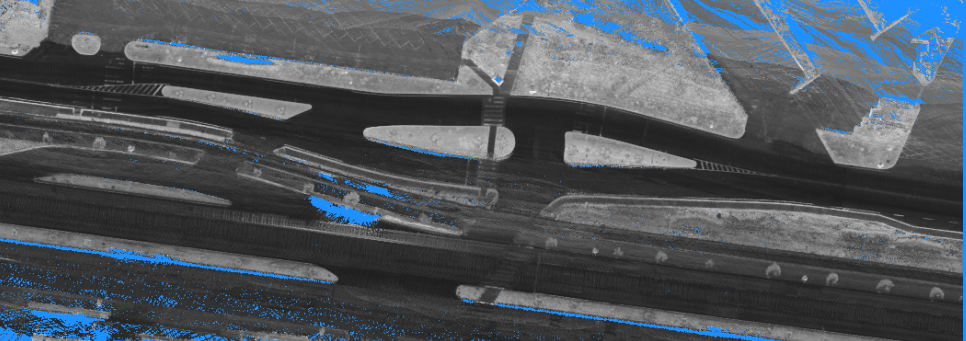}
\caption{Remission map}
\end{figure}

\citet{item62} proposed a MCL localization method that compares satellite aerial maps with remission maps. Aerial maps are downloaded offline from sources in the Internet, like OpenStreetMap, and remission maps are built online from LIDAR reflectance intensity data. The MCL algorithm is employed to estimate car’s pose by matching remission maps to aerial maps using the Normalized Mutual Information (NMI) measure to calculate the particles likelihood. The method was evaluated on a 6.5 km dataset collected by the self-driving car IARA and achieved position estimation accuracy of 0.89 m. One advantage of this method is that it does not require building a map specifically for the method.

\citet{item63} proposed a localization method based on road feature detection. Their curb detection algorithm uses ring compression analysis and least trimmed squares is used to analyze the distance between consecutive concentric measurements (or rings) formed by a multilayer LIDAR (Velodyne HDL-32E) scan. The road marking detection algorithm uses Otsu thresholding \citep{item64} to analyze LIDAR reflectance intensity data. Curb and road marking features are stored in a grid map. A Monte Carlo Localization (MCL) algorithm is employed to estimate the car pose by matching road features extracted from multilayer LIDAR measurements to the grid map. The method was evaluated on the self-driving car, CARINA \citep{item50}, and has shown lateral and longitudinal localization estimation errors of less than 0.30 m. 

\citet{item65} proposed a multilayer adaptive Monte Carlo Localization (ML-AMCL) method that operates in combination with 3D point registration algorithms. For estimating the car pose, horizontal layers are extracted from 3D LIDAR measurements and separate AMCL instances are used to align layers with a 2D projection of a 3D point cloud map built using 3D point registration algorithms. For every pose estimate, a consistency check against a series of odometry measurements is performed. Consistent pose estimates are fused to a final pose estimate. The method was evaluated on real world data and achieved position estimation errors of 0.25 m relative to the GPS reference. Their map is, however, expensive to store, since it is a 3D map.

\citet{item66} proposed a localization method based on the MCL algorithm that corrects particles’ poses by map-matching between 2D online occupancy grid-maps and 2D offline occupancy grid-maps, as illustrated in Figure 3. Two map-matching distance functions were evaluated: an improved version of the traditional Likelihood Field distance between two grid-maps, and an adapted standard Cosine distance between two high-dimensional vectors. An experimental evaluation on the self-driving car IARA demonstrated that the localization method is able to operate at about 100 Hz using the Cosine distance function, with lateral and longitudinal errors of 0.13 m and 0.26 m, respectively.

\begin{figure*}[htb!]
\centering
\subfloat[]{\includegraphics[width=0.4\textwidth]{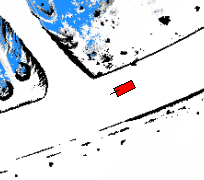}}
\subfloat[]{\includegraphics[width=0.4\textwidth]{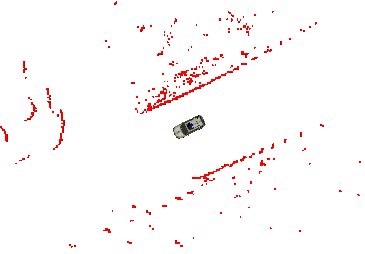}}
\caption{Localization method proposed by \citet{item66}. (a) Offline occupancy grid-map – the red rectangle is the car’s localization, black cells contain obstacles, white cells are obstacle-free, and blue cells are regions untouched by sensors during mapping. (b) Online occupancy grid-map. The online map is matched against the offline map to compute the self-driving car’s precise localization.}
\end{figure*}

\citet{item67} proposed a probabilistic localization method that models the world as a multiresolution map of mixture of Gaussians. Their Gaussian mixture maps represent the height and reflectance intensity distribution of the environment measured by LIDAR scanners (Velodyne HDL-32E). An Extended Kalman filter (EKF) localization algorithm is used to estimate the car’s pose by registering 3D point clouds against the Gaussian mixture multiresolution-maps. The method was evaluated on two self-driving cars in adverse weather conditions and presented localization estimation errors of about 0.15 m.

\subsubsection{LIDAR plus Camera-Based Localization}

Some methods use LIDAR data to build a map, and camera data to estimate the localization of the self-driving car relative to this map. \citet{item68} proposed a localization method that matches stereo images to a 3D point-cloud map. The map was generated by a mapping company 
and it is composed of geometric data (latitude, longitude and altitude) and LIDAR reflectance intensity data acquired from odometer, RTK-GPS, and 2D LIDAR scanners. Xu et al. transform the 3D-points of the map from the real-world coordinate system to the camera coordinate system, and extract depth and intensity images from them. A MCL algorithm is used to estimate the car localization by matching stereo depth and intensity images taken from the car’s camera to depth and intensity images extracted from the 3D point-cloud map. The method was evaluated on real world data and presented localization estimation errors between 0.08 m and 0.25 m.

\citet{item69} proposed a method for autonomous vehicle localization that matches ground panoramic-images to satellite images captured in different seasons of the year. In their method, LIDAR data is classified into ground/non-ground categories. Next, ground images captured by a panoramic camera in the autonomous vehicle are segmented into ground/non-ground regions using the LIDAR data, and then warped to obtain a bird’s-eye view. The satellite image is also segmented into ground/non-ground regions using k-means clustering. A MCL is then used to estimate the pose by matching bird’s-eye images to the satellite images. The method was evaluated on the NavLab11 autonomous vehicle and achieved position estimation errors between 3 m and 4.8 m.

\subsubsection{Camera-Based Localization}

Some methods rely mainly on camera data to localize self-driving cars. \citet{item70} proposed a localization method based on visual odometry and road maps. They use the OpenStreetMap, extracting from it all crossings and all drivable roads (represented as piece-wise linear segments) connecting them in the area of interest. They, then, build a graph-based representation of this road map and a probabilistic model of how the car traverses this graph. Using this probabilistic model and visual odometry measurements, they estimate the car displacement relative to the road map. 

A recursive Bayesian filtering algorithm is used to perform inferences in the graph by exploiting its structure and the model of how the car moves, as measured by the visual odometry. This algorithm is able to pinpoint the car’s position in the graph by increasing the probability that the current pose lies in a point of the graph that is correlated with latest car movements (distance traveling straight and recent curves) and by decreasing the probability that it is in a point that is not correlated. The localization method was evaluated on the KITTI visual odometry dataset and was able to localize the autonomous vehicle, after 52 s of driving, with an accuracy of 4 m on an 18 km\textsuperscript{2} map containing 2,150 km of drivable roads. 

Some methods use camera data to build a feature map. \citet{item71} describe the localization methods used by the self-driving car Bertha to drive autonomously on the Bertha Benz Memorial Route. Two complementary vision based localization techniques were developed, named Point Feature based Localization (PFL) and Lane Feature based Localization (LFL). In PFL, the current camera image is compared with images of a sequence of camera images that is acquired previously during mapping using DIRD descriptors extracted from them. A global location estimate is recovered from the global position of the images captured during mapping. In LFL, the map, computed semi-automatically, provides a global geometric representation of road marking features (horizontal road signalization). The current camera image is matched against the map by detecting and associating road marking features extracted from a bird’s-eye view of the camera image with horizontal road signalization stored in the map. Location estimates obtained by PFL and LFL are, then, combined by a Kalman filter (the authors do not provide an estimate of the combined localization error). Localization methods similar to LFL were proposed by \citet{item72}, \citet{item73}, and \citet{item74}. 

Some methods employ camera data to construct a feature map, but adopt alternative types of features. \citet{item75} proposed a localization method based on textual feature detection. Off-the-shelf text extraction techniques are used to identify text labels in the environment. A MCL algorithm is employed to integrate multiple observations. The method was evaluated on real world data and presented location estimation errors between 1 m and 25 m. \citet{item76} proposed the use of pole-like landmarks as primary features, because they are distinctive, long-term stable, and detectable by stereo cameras. Furthermore, they allow a memory efficient map representation. The feature detection is performed mainly by a stereo camera. Localization is performed by a MCL algorithm coupled with a Kalman filter for robustness and sensor fusion. The method was evaluated on an autonomous vehicle and achieved position estimation errors between 0.14 m and 0.19 m.

Some methods employ neural networks to localize self-driving cars \citep{item77, item78}. They consist of correlating camera images and associated global positions. In the mapping phase, the neural network builds a representation of the environment. For that, it learns a sequence of images and global positions where images were captured, which are stored in a neural map. In the localization phase, the neural network uses previously acquired knowledge, provided by the neural map, to estimate global positions from currently observed images. These methods present error of about some meters and have difficulty in localizing self-driving cars on large areas.

\subsection{Offline and Online Mapping of Unstructured Environments}

The offline and online \textbf{Mapper} subsystems are responsible for computing maps of the environment where the self-driving car operates. These subsystems are fundamental for allowing them to navigate on unstructured environments without colliding with static obstacles (e.g., signposts, curbs, etc.).

Representations of the environment are often distinguished between topological \citep{item79, item80, item81} and metric \citep{item82, item54, item83}. Topological representations model the environment as a graph, in which nodes indicate significant places (or features) and edges denote topological relationships between them (e.g., position, orientation, proximity, and connectivity). The resolution of these decompositions depends on the structure of the environment. 

Metric representations usually decompose the environment into regularly spaced cells. This decomposition does not depend on the location and shape of features. Spatial resolution of metric representations tends to be higher than that of topological representations. This makes them the most common space representation. For a review on the main vision-based methods for creating topological representations, readers are referred to \citet{item84}. Here, we discuss the most important methods for computing metric representations, which can be further subdivided into grid representations with regular spacing resolution and varied spacing resolution.

\subsubsection{Regular Spacing Metric Representations}

For self-driving cars, one of the most common representations of the environment is the Occupancy Grid Map (OGM), proposed by \citet{item85}. An OGM discretizes the space into fixed size cells, usually of the order of centimeters. Each cell contains the probability of occupation of the region associated with it. The occupancy probability is updated independently for each cell using sensor data. 3D sensor measurements that characterize the environment can be projected onto the 2D ground plane for simplicity and efficiency purposes. The assumption of independence of the occupancy probability of each cell makes the OGM algorithm implementation fast and easy \citep{item61}. However, it generates a sparse space representation, because only those cells touched by the sensor are updated \citep{item86}. Figure 4 shows examples of OGMs computed by the self-driving car IARA. In the maps of this figure, shades of grey represents the occupancy probability of cells, with black being the maximum occupancy probability and white being the minimum occupancy probability, and blue represents cells that were not observed by sensors yet.

\begin{figure*}[htb!]
\centering
\subfloat[]{\includegraphics[width=0.3\textwidth]{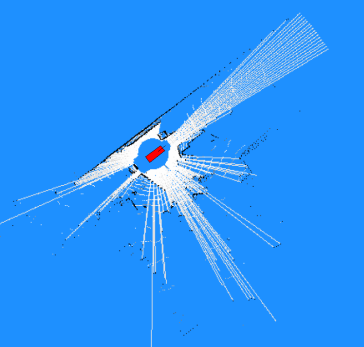}}~~~~
\subfloat[]{\includegraphics[width=0.3\textwidth]{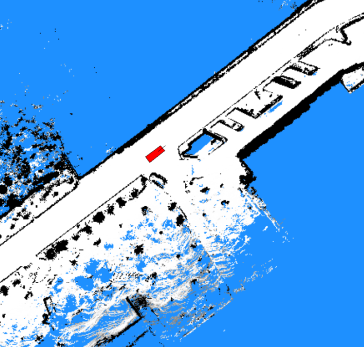}}~~~~
\subfloat[]{\includegraphics[width=0.3\textwidth]{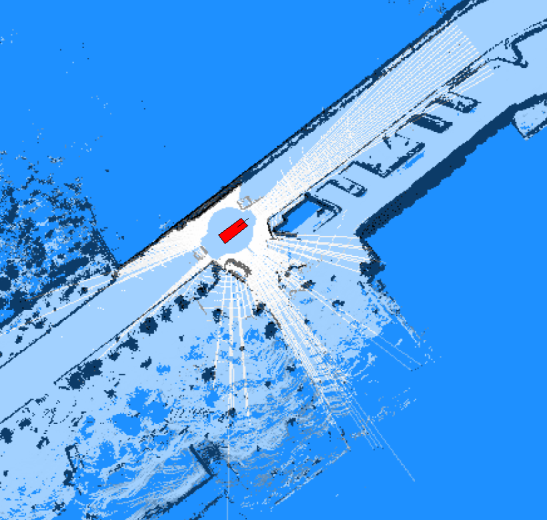}}
\caption{Examples of OGMs computed by the self-driving car IARA. Shades of grey represent the occupancy probability of map cells and blue represents cells that were not observed by sensors yet. Black represents maximal occupancy probability, while white represents minimal occupancy probability. (a) The instantaneous OGM that is computed online using a single LIDAR point cloud. (b) The offline OGM that is constructed offline using data obtained by various sensors and stores the most consolidated information that IARA has about the world. (c) The online map that is used by the motion planner subsystem. It is updated online by merging the previous ones, where the dark blue represents the cells in common in both maps and the light blue represents the cells that exist in the first but not in the second one.}
\end{figure*}

\citet{item87} presented the GraphSLAM algorithm. GraphSLAM is an offline Simultaneous Localization And Mapping (SLAM) algorithm, which extracts from sensor data a set of soft constraints, represented by a sparse graph. It is able to obtain the map of the environment and the robot path that was followed during sensor data capture by resolving these constraints into a globally consistent estimate of robot poses. The constraints are related to the properties of the data captured by the sensors (e.g. distance and elapsed time between two consecutive odometry measurements) and are generally nonlinear but, in the process of resolving them, they are linearized and the resulting least squares problem is solved using standard optimization techniques. Once the robot poses are estimated, the map can be built from sensor data using probabilistic algorithms for OGM computation.

\citet{item54}  employed GraphSLAM and data captured by an odometer, a 3D LIDAR Velodyne HDL-32E, IMU, and a low-cost GPS to build OGMs and for self-driving cars. In their work, GPS data is used to identify and remove systematic errors in odometer data. GPS data is also employed to detect revisited regions (loop closure), and a Generalized Iterative Closest Point (GICP) algorithm \citep{item88}  is employed to estimate displacements between poses in subsequent visits. Data from GPS, IMU, calibrated odometer, and loop closures are used to introduce restrictions to the GraphSLAM algorithm, which calculates the most likely set of poses given sensor data. Finally, 3D LIDAR data and their poses, calculated by GraphSLAM, are used (offline) to build an OGM. Besides the GraphSLAM bundle adjustment optimization, there are other algorithms for offline mapping, such as gradient descent, conjugate gradient and Levenberg Marquardt \citep{item61}.

During operation, a self-driving car requires an offline map for localization. It may also require an online map for applications where the environment contains moving obstacles. These two maps can be merged for improved operation safety \citep{item89}.

SLAM can be computed online using FastSLAM \citep{item90}. FastSLAM was originally designed for landmark maps but further extended for OGMs. The original FastSLAM uses a particle filter algorithm to estimate the self-driving car’s path and an EKF algorithm for the landmarks’ positions, while the extended FastSLAM employs a particle filter to estimate both the car’s path and the occupancy probability of OGM’s cells. There are other algorithms for online grid mapping \citep{item91,item92}.

\subsubsection{Varied Spacing Metric Representations}

An alternative metric representation of the environment is the Octree map, proposed by \citep{item82}, which stores information with varied 3D resolutions. Compared to OGMs with varied 3D resolutions, OctoMaps (Octree-based Maps) store only the space that is observed and, consequently, are more memory efficient. They are, however, computationally intensive \citep{item93}, which makes them unsuitable for self-driving cars, considering the hardware currently available. In addition, the typical self-driving car can be modeled as a parallelepiped, or a series of interconnected parallelepipeds, and the map may only contain information pertained to objects of the environment that are obstacles for the movements of these parallelepipeds.

Another alternative metric representation is a hybrid map proposed by \citet{item94}, which stores occupancy and distance measurements with varied resolutions. For this, measurements are stored in grid cells of increasing size from the center of the car. Thus, computational efficiency is gained by having a high resolution in the proximity of the sensor and a lower resolution as the distance from the sensor grows. This follows characteristics of some sensors with respect to distance and measurement density (e.g., angular resolution of laser sensors).

Traditional methods for computing OGMs are constrained by the assumption that the probability of occupancy of each cell of the map is modeled as an independent random variable \citep{item61}. Intuitively, this assumption is not true, because real environments have some inherent structure. An alternative metric representation is the Gaussian Process Occupancy Map (GPOM) proposed by \citet{item95}. A GPOM uses a Gaussian Process (GP) to learn the structure of the environment from sparse sensor measurements in a training dataset and, subsequently, estimate the probability of occupancy of cells that were not directly intercepted by the sensor. Experiments have shown that localization errors are more than three times lower as those estimated by a particle filter localization and OGM \citep{item96}. However, this inference carries a high computational cost of  \( O \left( N^{3} \right)  \), where  \( N \)  is the number of samples in the training dataset, and, therefore, it is not directly applicable to large-scale scenarios of self-driving cars \citep{item86}. 

Another alternative metric representation is the Hilbert map proposed by \citet{item97}. Hilbert maps represent the probability of occupancy with a linear discriminative model that operates on a high-dimensional feature vector and projects observations into a Hilbert space. The model can be trained and updated online using Stochastic Gradient Descent (SGD). Experiments showed that a Hilbert map presents accuracy comparable to a Gaussian Process Occupancy Map (GPOM), but it has time complexity linear with the number of samples in the training dataset.

A further alternative metric representation is the Discrete Cosine Transform (DCT) map proposed by \citet{item83}. 
A DCT map assigns to each point of the space a LIDAR decay rate, which models the local permeability of the space for laser rays. In this way, the map can describe obstacles of different laser permeability, from completely opaque to completely transparent. A DCT map is represented in the discrete frequency domain and converted to a continuously differentiable field in the position domain using the continuous extension of the inverse DCT. DCT maps represent LIDAR data more accurately than OGM, GPOM, and Hilbert map regarding the same memory requirements. Nonetheless, the above continuous metric representations are still slower than OGM and not widely applicable to large-scale and real-time self-driving scenarios yet.

\subsection{Road Mapping}

The mapping techniques presented above can be used to map unstructured environments and make it possible the operation of self-driving cars in any kind of flat terrain. However, for autonomous operation in roads and highways, where there are rules consubstantiated in markings in the floor and other forms of traffic signalization, self-driving cars need road maps. 
The \textbf{Road Mapper} subsystem is responsible for managing information about roads and lanes, and for making them available in maps with geometrical and topological properties.

Road maps can be created manually from aerial images. However, the high cost of maintaining them through manual effort makes this approach unviable for large scale use. Because of that, methods for automated generation of road maps from aerial images have been proposed. 

Here, we discuss the most important road map representations and the methods used for road map creation.

\subsubsection{Metric Representations}

A simple metric representation for a road map is a grid map, which discretizes the environment into a matrix of fixed size cells that contain information about the roads. However, a grid representation might require a wasteful use of memory space and processing time, since usually most of the environment where self-driving cars operate is not composed of roads, but buildings, free space, etc. 

\citet{item98} proposed a metric road map (Figure 5), a grid map, where each 0.2 m $ \times $  0.2 m-cell contains a code that, when nonzero, indicates that the cell belongs to a lane. Codes ranging from 1 to 16 represent relative distances from a cell to the center of the lane, or the type of the different possible lane markings (broken, solid, or none) present in the road. To save memory space, these maps are stored in a compacted form, where only non-zero cells are present. The authors used Deep Neural Networks (DNNs) to infer the position and relevant properties of lanes with poor or absent lane markings. The DNN performs a segmentation of LIDAR remission grid maps into road grid maps, assigning the proper code (from 1 to 16) to each map cell. A dataset of tens of kilometers of manually marked road lanes was used to train the DNN, allowing it achieve an accuracy of 83.7$\%$, which proved to be sufficient for real-world applications.

\begin{figure*}[htb!]
\centering
\subfloat[]{\includegraphics[height=0.376\textwidth]{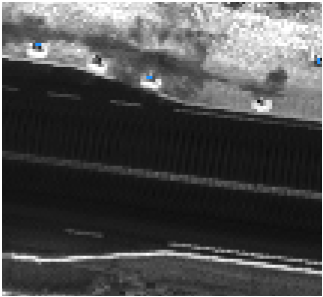}}~~~~
\subfloat[]{\includegraphics[height=0.376\textwidth]{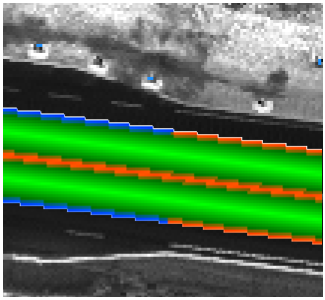}}
\caption{Metric road map. (a) Crop of a remission grid map. (b) Same as (a) with superimposed road grid map. The value of each cell of the road grid map represents different classes of features. Red cells represent a Solid Line close to the lane boundary, blue cells represent a Broken Line close to the lane boundary, different shades of green represent different distances from the map cell to the center of the lane.}
\end{figure*}

\subsubsection{Topological Representations}

Sequences of waypoints, representing the center of the lanes of the roads of interest, are an alternative simple topological road map representation. They can be defined manually, or captured semi automatically or automatically from OGM, aero-photogrammetry or satellite images. For the 2005 DARPA Grand Challenge, DARPA used a road map representation named Route Data Definition File (RDDF) \citep{item99}, 
which is a formatted file that contains waypoint coordinates and other associated information (latitude, longitude, lateral boundary offset, and course speed) that specified the path for the operation of the competing self-driving cars.

A more sophisticated road map representation depicts the environment as a graph-like model, in which vertices denote places and edges denote topological relationships between them. Topological maps can hold more complex information, including multiple lanes, lane crossings, and lane mergers. For the 2007 DARPA Urban Challenge, it was proposed the Route Network Definition File (RNDF) \citep{item100}, 
which is a topological map defined as a formatted file that specifies road segments for the operation of the competing self-driving cars. In a RNDF, the road network includes one or more segments, each of which comprises one or more lanes. A segment is characterized by the number of lanes, street name, and speed limit. A lane is characterized by the width of the lane, lane markings, and a set of waypoints. Connections between lanes are characterized by exit and entry waypoints. 

\citet{item9} used a graph model of the DARPA RNDF for the self-driving car Boss (Carnegie Mellon University’s car that claimed first place in the 2007 DARPA Urban Challenge). Each node in the graph denotes a waypoint and directional edges denote lanes that connect the node to all other waypoints it can reach. Costs are assigned to the edges based on a combination of several factors, including expected time to traverse the lane associated to the edge, length of the lane, and complexity of the environment. The authors used manual annotation of road shapes extracted from aerial imagery in order to create a road map for the self-driving car Boss. The obtained local road shapes were accurate; however, global positions were not so accurate due to the image resolution and the global registration method.

\citet{item101} proposed the OpenStreetMap (OSM). OSM is a collaborative project to create a free editable map of the world. 
Its creation and growth have been motivated by restrictions on use or availability of map information across much of the world and by the advent of inexpensive portable satellite navigation devices. OSM models the environment with topological maps using three primitive elements, namely: nodes, ways and relations. Nodes denote geographical points, ways denote lists of nodes (polylines), and relations consist of any number of members that may be of any of the three types and have specified roles. Other road properties, like the driving direction and the number of lanes, are given as properties of the elements. 

\citet{item102}  proposed a highly detailed topological road map, called lanelet map, for the self-driving car Bertha. The lanelet map includes both geometrical and topological features of road networks, such as roads, lanes, and intersections, using atomic interconnected drivable road segments, called lane-lets, illustrated in Figure 6. The geometry of a lanelet is defined by a left and a right bound, each one corresponding to a list of points (polyline). This representation implicitly defines the width and shape of each lane and its driving orientation. The adjacency of lanelets composes a weighted directed graph, in which each lanelet denotes a vertex and the length of a lanelet denotes the weight of its outgoing edges. Other elements describe constraints, such as speed limits and traffic regulations, like crossing and merging rights. The authors adopted manual annotation of all elements and properties of their lanelet maps for the self-driving car Bertha. Virtual top-view images were used as a foundation for the manual annotation of lanelets using the OSM format and the Java OSM editor. The lanelet map was successfully tested throughout an autonomous journey of 103 km on the historic Bertha Benz Memorial Route.

\begin{figure}[htb!]
\centering
\includegraphics[width=0.4\textwidth]{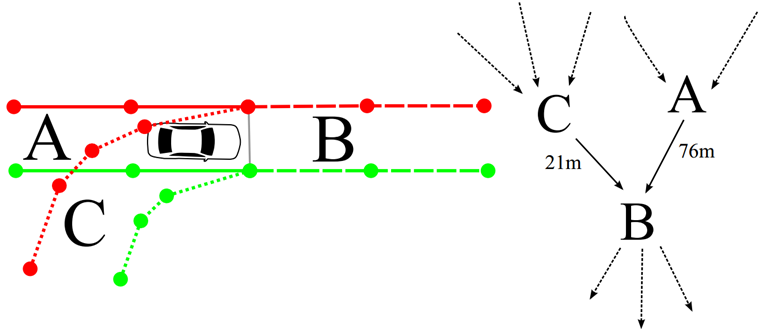}
\caption{A graph model of a lanelet map used by the self-driving car Bertha \citep{item102}. Red and green dots denote the polylines of left and right bounds of lanelets A, B and C, respectively. The graph shows the merge of A and C into B.}
\end{figure}

\citet{item103}  proposed the Road Tracer method that seeks to produce a road map directly from the output of a CNN. It uses an iterative graph construction process that adds individual road segments, one at a time, and uses the CNN to decide on the next segment to be added. The test on aerial images covering 24 km\textsuperscript{2} of 15 cities achieved an average error of 5$\%$  on a junction-by-junction matching metric.

High-Definition (HD) maps are a new generation of topological maps that are being used by some existing experimental self-driving cars. HD maps have high-precision (centimeter-level precision) and contain rich information, such as lane positions, road boundaries, and road curvatures. Since the cost incurred to create HD maps is significant, there are platforms available to provide HD maps as a service. 
\citet{item104}  assessed and ranked the top vendors, namely Google, HERE, TomTom, and Apple.

Besides the above-mentioned map generation methods, many other methods have been proposed for automated generation of road maps from aerial images. \citet{item105}   used higher-order Conditional Random Fields (CRF) to model the structures of the road network by segmenting images into superpixels and adding paths to connect these superpixels. 
\citet{item106} used Convolutional Neural Networks (CNNs) to obtain road segments from satellite images. They improved the predictive performance by using unsupervised learning methods for initializing the feature detectors and taking advantage of the local spatial coherence of the output.

There are several techniques for lane detection in images captured by top-view or front-facing cameras. For reviews on these topics, readers are referred to \citet{item107} and \citet{item108}.  

\citet{item109} use ground grid maps, in which each cell represents the probability of a ground location with high reflectance intensity, for the BMW’s self-driving car. The ground grid maps are used to extract road boundaries using a second-degree polynomial model. Lane localization is used in conjunction with a HD map in order to obtain a higher level of understanding of the environment. The map consists of two layers: a semantic geometric layer and a localization layer. The semantic geometric layer contains the lane geometry model and the high-level semantic information, such as lane connectivity. The localization layer contains lane markings and road boundaries, which, together with GPS and car odometry, can be used to match the car onto the map. 

\citet{item110} also used LIDAR remission data to detect lane markings and camera images, in case lane makings are not well defined. The lane markings on the road used during experiments was made to look good with headlight at night by using a special paint with good reflectance intensity. With this approach, road markings could also be detected with LIDAR, even in the case of changes in illumination due to rain or shadows. The lane marking detection technique based on camera image was run only in vulnerable situations (e.g., backlight and low light). This approach was successfully tested in a course 2 km in Seoul, Korea.

\subsection{Moving Objects Tracking}

The \textbf{Moving Objects Tracker (MOT)} subsystem (also known as Detector and Tracker of Moving Obstacles – DATMO) is responsible for detecting and tracking the pose of moving obstacles in the environment around the self-driving car. This subsystem is essential for enabling self-driving cars to make decisions about how to behave to avoid collision with potentially moving objects (e.g., other vehicles and pedestrians). 

Moving obstacles’ positions over time are usually estimated from data captured by ranging sensors, such as LIDAR and RADAR, or stereo and monocular cameras. Images from monocular cameras are useful to provide rich appearance information, which can be explored to improve moving obstacle hypotheses. To cope with uncertainty of sensor measurements, Bayes filters (e.g., Kalman and particle filter) are employed for state prediction. 

Various methods for MOT have been proposed in the literature. They can be mainly categorized into six classes: traditional, model based, stereo vision based, grid map based, sensor fusion based, and deep learning based. Here, we present the most recent and relevant ones published in the last ten years. For earlier works, readers are referred to \citet{item111}, \citet{item112} and    \citet{item113}.

\subsubsection{Traditional Based MOT}

Traditional MOT methods follow three main steps: data segmentation, data association, and filtering \citep{item111}. In the data segmentation step, sensor data are segmented using clustering or pattern recognition techniques. In the data association step, segments of data are associated with targets (moving obstacles) using data association techniques. In the filtering phase, for each target, a position is estimated by taking the geometric mean of the data assigned to the target. Position estimates are usually updated by Kalman or particle filters. 
\citet{item114} propose a traditional method for detection and tracking of moving vehicles using 3D LIDAR sensor. The 3D LIDAR point cloud is segmented into clusters of points using the Euclidean distance. Obstacles (clusters) observed in the current scan sensor are associated with obstacles observed in previous scans using a nearest neighbor algorithm. States of obstacles are estimated using a particle filter algorithm. Obstacles with velocity above a given threshold are considered moving vehicles. 
\citet{item115} build a cube bounding box for each cluster and use box dimensions for distinguishing whether a cluster is a vehicle or not. Data association is solved by an optimization algorithm. A Multiple Hypothesis Tracking (MHT) algorithm is employed to mitigate association errors.
\citet{item116} use images captured by a monocular camera to filter out 3D LIDAR points that do not belong to moving objects (pedestrians, cyclists and vehicles). Once filtered, object tracking is performed based on a segment matching technique using features extracted from images and 3D points. 

\subsubsection{Model Based MOT }

Model-based methods directly infer from sensor data using physical models of sensors and geometric models of objects, and employing non-parametric filters (e.g., particle filters) \citep{item111}. Data segmentation and association steps are not required, because geometric object models associate data to targets. 
\citet{item117} present the model-based method for detection and tracking of moving vehicles adopted by the self-driving car, Junior \citep{item10}(Stanford University’s car that finished in second place in the 2007 DARPA Urban Challenge). Moving vehicle hypotheses are detected using differences over LIDAR data between consecutive scans. Instead of separating data segmentation and association steps, new sensor data are incorporated by updating the state of each vehicle target, which comprises vehicle pose and geometry. This is achieved by a hybrid formulation that combines Kalman filter and Rao-Blackwellized Particle Filter (RBPF). The work of  
\citet{item117} was revised by \citet{item118} that propose to combine RBPF with Scaling Series Particle Filter (SSPF) for geometry fitting and for motion estimate throughout the entire tracking process. The geometry became a tracked variable, which means that its previous state is also used to predict the current state. 
\citet{item119} propose a model-based MOT method that aims at finding the most likely set of tracks (trajectories) of moving obstacles, given laser measurements over a sliding window of time. A track is a sequence of object shapes (L-shape, I-shape and mass point) produced over time by an object satisfying the constraints of a measurement model and motion model from frame to frame. Due to the high computational complexity of such a scheme, they employ a Data Driven Markov chain Monte Carlo (DD-MCMC) technique that enables traversing efficiently in the solution space to find the optimal solution. DD-MCMC is designed to sample the probability distribution of a set of tracks, given the set of observations within a time interval. At each iteration, DD-MCMC samples a new state (set of tracks) from the current state following a proposal distribution. The new candidate state is accepted with a given probability. To provide initial proposals for the DD-MCMC, dynamic segments are detected from laser measurements that fall into free or unexplored regions of an occupancy grid map and moving obstacle hypotheses are generated by fitting predefined object models to dynamic segments. 
\citet{item120} adopt a similar method to the model-based one, but they do not assume prior categories for moving objects. A Bayes filter is responsible for joint estimation of the pose of the sensor, geometry of static local background, and dynamics and geometry of objects. Geometry information includes boundary points obtained with a 2D LIDAR. Basically, the system operates by iteratively updating tracked states and associating new measurements to current targets. Hierarchical data association works in two levels. In the first level, new observations (i.e., cluster of points) are matched against current dynamic or static targets. In the second level, boundary points of obstacles are updated.

\subsubsection{Stereo Vision Based MOT }

Stereo vision based methods rely on color and depth information provided by stereo pairs of images for detecting and tracking moving obstacles in the environment. 
\citet{item121} propose a method for obstacle detection and recognition that uses only synchronized video from a forward-looking stereo camera. The focus of their work is obstacle tracking based on the per-frame output of pedestrian and car detectors. For obstacle detection, they employ a Support Vector Machine (SVM) classifier with Histogram of Oriented Gradients (HOG) features for categorizing each image region as obstacle or non-obstacle. For obstacle tracking, they apply a hypothesize-and-verify strategy for fitting a set of trajectories to the potentially detected obstacles, such that these trajectories together have a high posterior probability. The set of candidate trajectories is generated by Extended Kalman Filters (EKFs) initialized with obstacle detections. Finally, a model selection technique is used to retain only a minimal and conflict-free set of trajectories that explain past and present observations. 
\citet{item122} describe the architecture of the modified Mercedes-Benz S-Class S500, Bertha, which drove autonomously on the historic Bertha Benz Memorial Route. For MOT, dense disparity images are reconstructed from stereo image pairs using Semi-Global Matching (SGM). All obstacles within the 3D environment are approximated by sets of thin and vertically oriented rectangles called super-pixels or stixels. Stixels are tracked over time using a Kalman filter. Finally, stixels are segmented into static background and moving obstacles using spatial, shape, and motion constraints. The spatio-temporal analysis is complemented by an appearance-based detection and recognition scheme, which exploits category-specific (pedestrian and vehicle) models and increases the robustness of the visual perception. The real-time recognition consists of three main phases: Region Of Interest (ROI) generation, obstacle classification, and object tracking. 
\citet{item123} compute a disparity map from a stereo image pair using a semi-global matching algorithm. Assisted by disparity maps, boundaries in the image segmentation produced by simple linear iterative clustering are classified into coplanar, hinge, and occlusion. Moving points are obtained during ego-motion estimation by a modified RANdom SAmple Consensus (RANSAC) algorithm. Finally, moving obstacles are extracted by merging super-pixels according to boundary types and their movements.

\subsubsection{Grid Map Based MOT}

Grid map based methods start by constructing an occupancy grid map of the dynamic environment \citep{item111}. The map construction step is followed by data segmentation, data association, and filtering steps in order to provide object level representation of the scene. 
\citet{item124}  propose a grid-based method for detection and tracking of moving objects using stereo camera. The focus of their work is pedestrian detection and tracking. 3D points are reconstructed from a stereo image pair. An inverse sensor model is used to estimate the occupancy probability of each cell of the grid map based on the associated 3D points. A hierarchical segmentation method is employed to cluster grid cells into segments based on the regional distance between cells. Finally, an Interactive Multiple Model (IMM) method is applied to track moving obstacles. \citet{item125}  use an octree-based 3D local occupancy grid map that divides the environment into occupied, free, and unknown voxels. After construction of the local grid map, moving obstacles can be detected based on inconsistencies between observed free and occupied spaces in the local grid map. Dynamic voxels are clustered into moving objects, which are further divided into layers. Moving objects are classified into known categories (pedestrians, bikes, cars, or buses) using geometric features extracted from each layer. \citet{item126}  leverage a 2.5D occupancy grid map to model static background and detect moving obstacles. A grid cell stores the average height of 3D points whose 2D projection falls into the cell space domain. Motion hypotheses are detected from discrepancies between the current grid and the background model.

\subsubsection{Sensor Fusion Based MOT }

Sensor fusion-based methods fuse data from various kinds of sensors (e.g., LIDAR, RADAR, and camera) in order to explore their individual characteristics and improve environment perception. \citet{item127} present the sensor fusion-based method for detection and tracking of moving vehicles adopted by the self-driving car Boss \citep{item9} (Carnegie Mellon University’s car that finished in first place in the 2007 DARPA Urban Challenge). The MOT subsystem is divided into two layers. The sensor layer extracts features from sensor data that may be used to describe a moving obstacle hypothesis according to either a point model or a box model. The sensor layer also attempts to associate features with currently predicted hypotheses from the fusion layer. Features that cannot be associated to an existing hypothesis are used to generate new proposals. An observation is generated for each feature associated with a given hypothesis, encapsulating all information that is necessary to update the estimation of the hypothesis state. Based on proposals and observations provided by the sensor layer, the fusion layer selects the best tracking model for each hypothesis and estimates (or updates the estimation of) the hypothesis state using a Kalman Filter. \citet{item128}  describe the new MOT subsystem used by the new experimental self-driving car of the Carnegie Mellon University. The previous MOT subsystem, presented by \citet{item127}, was extended for exploiting camera data, in order to identify categories of moving objects (e.g., car, pedestrian, and bicyclists) and to enhance measurements from automotive-grade active sensors, such as LIDARs and RADARs. \citet{item129} use scan lines that can be directly obtained from 2D LIDARs, from the projection of 3D LIDARs onto a 2D plane, or from the fusion of multiple sensors (LIDAR, RADAR and camera). Scan lines are transformed into world coordinates and segmented. Line and corner features are extracted for each segment. Segments are associated with existing obstacles and kinematics of objects are updated using a Kalman filter. \citet{item130}  merge tracks of moving obstacles generated from multiple sensors, such as RADARs, 2D LIDARs, and a 3D LIDAR. 2D LIDAR data is projected onto a 2D plane and moving obstacles are tracked using Joint Probabilistic Data Association Filter (JPDAF). 3D LIDAR data is projected onto an image and partitioned into moving obstacles using a region growing algorithm. Finally, poses of tracks are estimated or updated using Iterative Closest Points (ICP) matching or image-based data association. \citet{item131}  describe the context-aware tracking of moving obstacles for distance keeping used by the new experimental self-driving car of the Carnegie Mellon University. Given the behavioral context, a ROI is generated in the road network. Candidate targets inside the ROI are found and projected into road coordinates. The distance-keeping target is obtained by associating all candidate targets from different sensors (LIDAR, RADAR, and camera). \citet{item132}  fuse LIDAR and camera data to improve the accuracy of pedestrian detection. They use prior knowledge of a pedestrian height to reduce false detections. They estimate the height of the pedestrian according to pinhole camera equation, which combines camera and LIDAR measurements. 

\subsubsection{Deep Learning Based MOT }

Deep learning based methods use deep neural networks for detecting positions and geometries of moving obstacles, and tracking their future states based on current camera data. \citet{item133}  propose a neural-based method for detection of moving vehicles using the Overfeat Convolutional Neural Network (CNN) \citep{item134}  and monocular input images with focus on real-time performance. The Overfeat CNN aims at predicting location and range distance (depth) of cars in the same driving direction of the ego-vehicle using only the rear view of them. \citet{item135} address moving obstacle tracking for a closely related application known as $``$follow the leader$"$, which is relevant mainly for convoys of self-driving cars. The tracking method is built on top of the Generic Object Tracking Using Regression Networks (GOTURN) \citep{item136}. GOTURN is a pre-trained deep neural network capable of tracking generic objects without further training or object-specific fine-tuning. Initially, GOTURN receives as input an image and a manually delimited bounding box of the leader vehicle. It is assumed that the object of interest is in the center of the bounding box. Subsequently, for every new image, GOTURN gives as output an estimate of the position and geometry (height and width) of the bounding box. The leader vehicle position is estimated using LIDAR points that fall inside the bounding box and are considered to be vehicle.

\subsection{Traffic Signalization Detection}

The \textbf{Traffic Signalization Detector {TSD}} subsystem is responsible for detecting and recognizing signs defined in the traffic rules so that the car can take correct decisions according to the traffic law. There are many tasks related to traffic signalization, and in this review, we explore three main topics: traffic lights, traffic signs, and pavement markings in the environment around the self-driving car. Each of these topics are described in detail in the next subsections. 

\subsubsection{Traffic Light Detection and Recognition}

Traffic light detection and recognition involve detecting the position of one or more traffic lights in the environment around the car (e.g., represented in an image) and recognizing their states (red, green, and yellow). Various methods for traffic light detection and recognition have been proposed in the literature. Here, we review only the most recent and relevant ones. For a more comprehensive review, readers are referred to Jensen et al. \citep{item137}.

Methods for traffic light detection and recognition can be mainly categorized into two classes: model-based and learning-based. Traffic lights have a well-defined structure in terms of color and shape information. A common traffic light has three bulbs (one for each state: red, green and yellow) and a well-defined form. Therefore, earlier, most of the approaches for traffic light detection and recognition were model-based. These approaches relied on hand-crafted feature engineering, which tried to leverage information humans have about the color and shape of the object to build a model capable of detecting and/or recognizing it. Methods that used color \citep{item138,item139} and shape \citep{item140, item141, item142} information were not robust when assumptions were not strictly observed. To increase their robustness, a combination of different features (e.g., color, shape, and structure) was proposed \citep{item143, item144, item145}. For example, \citet{item144} proposed a multi-feature system that combines both color (using color segmentation), shape/structure (using black box detection), and geographic information (by only using the system when known traffic lights are expected). Their system, however, suffer from the high number of hyper-parameters common on model-based approaches, which usually implicates the need of recalibration under certain circumstances. The authors performed the experiments on an in-house private data set and stated that failures were due to over-exposure, occlusions, non-standard installation of traffic lights, and several other situations that are not unusual in real-world cases. This combination, in the context of model-based approaches, showed not to be sufficient. Therefore, researchers began to introduce learning-based approaches. 

In learning-based approaches, features were still hand-crafted, but detection and/or recognition processes were changed from rule-based to learning-based. Cascade classifiers \citep{item146} were probably the first attempt to learning-based ap-proaches. Eventually, popular combinations of HoG and Gabor features with classifiers (such as SVM \citep{item147}, AdaBoost \citep{item148}, and JointBoost \citep{item149}) were also investigated. More recently, end-to-end methods (i.e., without the need of hand-crafted features) outperformed most of the model-based ones. \citet{item50} employed GPS data and a traffic light location database to identify a region of interest in the image, and a Convolutional Neural Network (CNN)  to recognize the traffic light state. Furthermore, state-of-the-art general object detectors \citep{item151,item152,item153} were successfully applied to the detection of traffic lights (often without recognizing their states). These general-purpose deep object detectors (or simply deep detectors, as they are often called), comprehensively, do not provide a breakdown on the performance of the traffic light detection and recognition task. Even though, unlike the model-based approaches, these deep detectors tend to be more robust to over-exposure, color distortions, occlusions, and others. A more complete discussion of these deep detectors applied to the traffic light detection can be found in \citet{item154}. There, the authors apply the YOLO \citep{item153} on the LISA \citep{item155} dataset and achieve 90.49$\%$  AUC when using LISA's training set. However, the performance drops to 58.3$\%$  when using training data from another dataset. Despite of the fact, it is still an improvement over previous methods, and it demonstrates that there is still a lot to be done. Learning-based approaches, especially those using deep learning, require large amounts of annotated data. Only recently large databases with annotated traffic lights are being made publicly available, enabling and powering learning-based approaches \citep{item155, item157,item158}.  

Despite of the advances on traffic light detection and recognition research, little is known about what is being used by research self-driving cars. Perhaps, one of the main reasons for this is that there were no traffic lights in the 2007 DARPA Urban Challenge. First place and second place finishers of this challenge (the Carnegie Mellon University’s team with their car Boss \citep{item9} and the Stanford University’s team with their car Junior \citep{item10}, respectively) recognized that traffic lights contribute to the complexity of urban environments and that they were unable to handle them at that time. In 2010, the Stadtpilot project’s presented their autonomous vehicle Leonie on public roads of Braunschweig, Germany \citep{item160}. Leonie used information about traffic light positions from a map and Car-to-X (C2X) communication to recognize traffic light states. However, during demonstrations a co-driver had to enter the traffic light state when C2X was not available. In 2013, the Carnegie Mellon University tested their self-driving car on public roads for over a year \citep{item161}. The Carnegie Mellon’s car used cameras to detect traffic lights and employed vehicle-to-infrastructure (V2I) communication to retrieve information from DSRC-equipped traffic lights. In 2013, Mercedes-Benz tested their robotic car Bertha \citep{item122} on the historic Bertha Benz Memorial Route in Germany. Bertha used vision sensors and prior (manual) information to detect traffic lights and recognize their states \citep{item146}. However, they state that the recognition rate needs to be improved for traffic lights at distances above 50m. 

\subsubsection{Traffic Sign Detection and Recognition}

Traffic sign detection and recognition involve detecting the locations of traffic signs in the environment and recognizing their categories (e.g., speed limit, stop, and yield sign.). For reviews on methods for traffic sign detection and recognition, readers are referred to \citet{item162} and \citet{item163}. 

Earlier, most of the approaches for traffic sign detection and recognition were model-based \citep{item164, item165} and used simple features (e.g., colors, shapes, and edges). Later, learning-based approaches (such as SVM \citep{item166}, cascade classifiers \citep{item167}, and LogitBoost \citep{item168}) started leveraging simple features, but evolved into using more complex ones (e.g., patterns, appearance, and templates). However, these approaches commonly tended to not generalize well. In addition, they usually needed fine-tuning of several hyper-parameters. Furthermore, some methods worked only with recognition and not with detection, perhaps because of the lack of data. Only after large databases were made available (such as the well-known German Traffic Sign Recognition (GTSRB) \citep{item169} and Detection (GTSDB) \citep{item170} Benchmarks, with 51,839 and 900 frames, respectively) that learning-based approaches \citep{item170, item171} could finally show their power, although some of them were able to cope with fewer examples \citep{item172}. With the release of even larger databases (such as STSD \citep{item173} with over 20,000 frames, LISA \citep{item155} with 6,610 frames, BTS \citep{item171} 25,634 frames for detection and 7,125 frames for classification, and Tsinghua-Tencent 100K \citep{item174} with 100,000 frames), learning-based approaches improved and achieved far better results when compared to their model-based counterparts. Some of these datasets report the number of frames including frames with background only. Following the rise of deep learning in general computer vision tasks, convolutional neural networks \citep{item174} are the state-of-the-art for traffic sign detection and recognition, achieving up to 99.71$\%$  and 98.86$\%$  of F1 score on the recognition task for the GTSRB and BTS respectively. 

Once more, little can be said about what is being employed for traffic sign detection and recognition by research self-driving cars. Again, maybe, one of the main drivers behind this is that only the stop-sign had to be detected and recognized in the 2007 DARPA Urban Challenge, since the map had detailed information on speed limits and intersection handling \citep{item175}. Some researchers (such as those of Bertha \citep{item122}) still prefer to rely on annotations about speed limit, right-of-way, among other signs. Other researchers \citep{item50} stated that their self-driving cars can handle traffic signs but did not provide information about their method.

\subsubsection{Pavement Marking Detection and Recognition}

Pavement marking detection and recognition involve detecting the positions of pavement marking and recognizing their types (e.g., lane markings, road markings, messages, and crosswalks). Most researches deal with only one type of pavement marking at a time and not with all of them at the same time. This may happen because there is neither a widely used database nor a consensus on which set of symbols researchers should be focused on when dealing with pavement marking detection and recognition.

One important pavement marking is the lane definition in the road. Earlier, most of the methods for lane marking detection were model- or learning-based \citep{item176}. Shape and color were the most usual features, and straight and curved lines (e.g., parabolic \citep{item177} and splines \citep{item178, item179}) were the most common lane representations. \citet{item179} propose a complete system for performing ego-lane analysis. Among the features of the systems, the authors claim to be able of detecting lanes and their attributes, crosswalks, lane changing events, and some pavement markings. The authors also release datasets for evaluating these types of systems. Deep learning is another popular method that has gained popularity lately and approaches like \citep{item180} have become showed very good results. \citet{item180}  propose (i) to use two laterally-mounted down-facing cameras and (ii) to model the lateral distance estimation as a classification problem in which they employ a CNN to tackle the task. In this setting, they argue to achieve sub-centimeter accuracy with less than 2 pixels of Mean Absolute Error (MAE) in a private database. For a review on this type of methods, readers are referred to \citet{item108}. 

Many of the methods for lane marking detection were also attempted for road marking detection. They commonly use geometric and photometric features \citep{item181}. Furthermore, various methods for road marking detection and recognition use Inverse Perspective Mapping (IPM), which reduces perspective effect and, consequently, makes the problem easier to solve and improves the accuracy of results. More recently, several methods \citep{item182, item183, item184} employed Maximally Stable Extremal Regions (MSER) for detecting regions of interest (i.e., regions that are likely to contain a road marking) and convolutional networks for recognizing road markings. \citet{item183}  propose the combination of IPM, MSER, and DBSCAN-based algorithm to perform the detection of the road markings and the combination of PCANet with either SVM or linear regression for the classification. While they achieve up to 99.1$\%$  of accuracy when evaluating the classification task alone, it drops to 93.1$\%$  of accuracy when the performance of detection and recognition is reported together.

In the context of road markings, messages are often handled separately. Some methods for message detection and recognition \citep{item184} treat different messages as different categories (i.e., they firstly detect positions of messages in the scene and then recognize their categories), while most of the methods firstly recognize letters and then writings using OCR-based approaches \citep{item185, item186}. The former is usually more robust to weather and lighting conditions, but the latter can recognize unseen messages. 

Still in the setting of road markings, pedestrian crossings are often investigated separately. Most of the methods for crosswalk detection exploit the regular shape and black-and-white pattern that crosswalks usually present \citep{item187, item188}. Therefore, in many practical applications, this task is set aside in favor of robust pedestrian detectors. For a review on these methods, readers are referred to \citet{item189}. Together with the review, \citet{item189} present a deep learning-based system to detect the presence of crosswalks in images. The authors provide pre-trained models that can be directly applied for this task. 

\section{Self-Driving Cars’ Decision Making}

In this section, we survey relevant techniques reported in the literature for the decision-making system of self-driving cars, comprising the route planning, behavior selection, motion planning, and control subsystems.

\subsection{Route Planning}

The \textbf{Route Planner} subsystem is responsible for computing a route,  \( W \), through a road network, from the self-driving car’s initial position to the final position defined by a user operator. A Route is a sequence of way points, i.e.  \( W= \{ w_{1}, w_{2}, \ldots,w_{ \vert W \vert } \}  \), where each way point,  \( w_{i} \), is a coordinate pair, i.e.  \( w_{i}= \left( x_{i}, y_{i} \right)  \), in the Offline Maps. If the road network is represented by a weighted directed graph, whose vertices are way points, edges connect pairs of way points, and edge weights denote the cost of traversing a road segment defined by two way points, then the problem of computing a route can be reduced to the problem of finding the shortest path in a weighted directed graph. However, for large road networks, time complexity of classical shortest path algorithms, such as Dijkstra \citep{item190} and A$\ast$  \citep{item191}, make them impractical.

In the last decade, there have been significant advances in the performance of algorithms for route planning in road networks. Newly developed algorithms can compute driving directions in milliseconds or less, even at continental scales. For a review on algorithms for route planning in road networks that are applicable to self-driving cars, readers are referred to \citet{item192}.

Techniques for route planning in road networks provide different trade-offs in terms of query time, preprocessing time, space usage, and robustness to input changes, among other factors. Such techniques can be categorized into four classes \citep{item192}: (i) goal-directed, (ii) separator-based, (iii) hierarchical, and (iv) bounded-hop. These classes may also be combined. 

\subsubsection{Goal-Directed Techniques}

Goal-directed techniques guide the search from the source vertex to the target vertex by avoiding scans of vertices that are not in the direction of the target vertex. A$\ast$  is a classic goal-directed shortest path algorithm. It achieves better performance than the Dijkstra’s algorithm by using a lower-bound distance function on each vertex, which causes vertices that are closer to the target to be scanned earlier. A$\ast$, Landmarks, and Triangle inequality (ALT) algorithm \citep{item193} enhances A$\ast$  by picking a small set of vertices as landmarks. During the preprocessing phase, distances between all landmarks and all vertices are computed. During the query phase, a valid lower-bound distance for any vertex is estimated, using triangle inequalities involving landmarks. The query performance and correctness depend on the wise choice of vertices as landmarks. 

Another goal-directed algorithm is Arc Flags \citep{item194}. During the preprocessing phase, the graph is partitioned into cells with a small number of boundary vertices and a balanced (i.e., similar) number of vertices. Arc flags for a cell \textit{i} are computed by growing backwards a shortest path tree from each boundary vertex, setting the \textit{i}-th flag for all arcs (i.e., edges) of the tree. During the query phase, the algorithm prunes arcs that do not have the flag set for the cell that contains the target vertex. The arc flags method has high preprocessing times, but the fastest query times among goal-directed techniques. 

\subsubsection{Separator-Based Techniques}

Separator-based techniques are based on either vertex or arc separators. A vertex (or arc) separator is a small subset of vertices (or arcs) whose removal decomposes the graph into several balanced cells. The vertex separator-based algorithm uses vertex separators to compute an overlay graph. Shortcut arcs are added to the overlay graph such that distances between any pair of vertices from the full graph are preserved. The overlay graph is much smaller than the full graph and is used to accelerate the query algorithm. The High-Performance Multilevel Routing (HPML) algorithm \citep{item195} is a variant of this approach that significantly reduces query time, but at the cost of increasing space usage and preprocessing time, by adding many more shortcuts to the graph across different levels.

The arc separator-based algorithm uses arc separators to decompose the graph into balanced cells, attempting to minimize the number of cut arcs, which connect boundary vertices of different cells. Shortcuts are added to the overlay graph in order to preserve distances between boundary vertices within each cell. Customizable Route Planning (CRP) algorithm \citep{item196} is a variant of this approach, which was designed to meet requirements of real-world road networks, such as handling turn costs and performing fast updates of the cost function. Its preprocessing has two phases. The first phase computes the multilevel partition and the topology of the overlays. The second phase computes the costs of clique arcs by processing the cells bottom-up and in parallel. Queries are processed as bidirectional searches in the overlay graph.

\subsubsection{Hierarchical Techniques}

Hierarchical techniques exploit the inherent hierarchy of road networks, in which main roads such as highways compound a small arterial subnetwork. When the source and target vertices are distant, the query algorithm only scans vertices of the subnetwork. The preprocessing phase computes the importance of vertices or arcs according to the actual shortest path structure. The Contraction Hierarchies (CH) algorithm \citep{item197} is a hierarchical technique that implements the idea of creating shortcuts to skip vertices with low importance. It repeatedly executes vertex contraction operations, which remove from the graph the least important vertex and create shortcuts between each pair of neighboring vertices, if the shortest path between them is unique and contains the vertex to be removed (i.e., contracted). CH is versatile, thus serving as a building block for other point-to-point algorithms and extended queries.

The REACH algorithm \citep{item198} is a hierarchical technique that, during the preprocessing phase, computes centrality measures (reach values) on vertices and, during query phase, uses them to prune Dijkstra-based bidirectional searches. Let \textit{P} be a shortest path that contains the vertex  \( w_{v} \), which goes from the source vertex  \( w_{s} \)  to the target vertex  \( w_{t} \). The reach value of  \( w_{v} \)  with respect to \textit{P} is \textit{r}( \( w_{v} \), \textit{P}) = min$ \{ $ distance( \( w_{s} \),  \( w_{v} \) ), distance( \( w_{v} \),  \( w_{t} \) )$ \} $.

\subsubsection{Bounded-Hop Techniques}

Bounded-hop techniques precompute distances between pairs of vertices by adding virtual shortcuts to the graph. Since precomputing distances among all pairs of vertices is prohibitive for large networks, bounded-hop techniques aim to get the length of any virtual path with very few hops. Hub Labeling (HL) \citep{item199} is a bounded-hop algorithm that, during the preprocessing phase, computes a label \textit{L( \( w_{u} \) )} for each vertex  \( w_{u} \)  of the graph, which consists of a set of hub vertices of  \( w_{u} \)  and their distances from  \( w_{u} \). These labels are chosen such that they obey the cover property: for any pair ( \( w_{s} \),  \( w_{t} \) ) of vertices, the intersection of labels \textit{L}( \( w_{s} \) ) and \textit{L}( \( w_{t} \) ) must contain at least one vertex of the shortest path from  \( w_{s} \)  to  \( w_{t} \). During the query phase, the distance ( \( w_{s} \),  \( w_{t} \) ) can be determined in linear time by evaluating the distances between hub vertices present in the intersection of labels \textit{L}( \( w_{s} \) ) and \textit{L}( \( w_{t} \) ). HL has the fastest known queries in road networks, at the cost of high space usage. 

The HL-$\infty$ algorithm \citep{item200} exploited the relationship between hub labelings and vertex orderings. It uses preprocessing algorithms for computing orderings that yield small labels. The iterative range optimization algorithm for vertex ordering \citep{item200} makes the query time of HL-$\infty$ algorithm twice as fast as HL. It starts with some vertex ordering (e.g., the one given by CH) and proceeds in a given number of iterative steps, each reordering a different range of vertices in decreasing order of importance. The Hub Label Compression (HLC) algorithm \citep{item201} reduces space usage by an order of magnitude, at the cost of higher query times, by combining common substructures that appear in multiple labels.

Another bounded-hop algorithm is Transit Node Routing (TNR) \citep{item202}, which uses distance tables on a subset of the vertices. During the preprocessing phase, it selects a small set of vertices as transit nodes and computes all pairwise distances between them. From transit nodes, for each vertex  \( w_{u} \), it computes a set of access nodes. If there is a shortest path from  \( w_{u} \), such that  \( w_{v} \)  is the first transit node in it, then the transit node  \( w_{v} \)  is an access node of  \( w_{u} \). It also computes distances between each vertex and its access nodes. A natural approach for selecting the transit node set is to select vertex separators or boundary vertices of arc separators as transit nodes. During query phase, the distance table is used to select the path from the source vertex  \( w_{s} \) to the target vertex  \( w_{t} \)  that minimizes the combined distance  \( w_{s} \) –\textit{a}( \( w_{s} \) )–\textit{a}( \( w_{t} \) ) –\(  w_{t} \), where \textit{a}( \( w_{s} \) ) and \textit{a}( \( w_{t} \) ) are access nodes. If the shortest path does not contain a transit node, then a local query is performed (typically CH).

\subsubsection{Combinations}

Individual techniques can be combined into hybrid algorithms that exploit different graph properties. The REAL algorithm \citep{item203} combines REACH and ALT. The ReachFlags algorithm \citep{item204} combines REACH and Arc Flags. The SHARC algorithm \citep{item205} combines the computation of shortcuts with multilevel Arc Flags. The CHASE algorithm \citep{item204} combines CH with Arc Flags. The TNR+AF algorithm \citep{item204} combines TNR and Arc Flags. The PHAST algorithm \citep{item206} can be combined with several techniques in order to accelerate them by exploiting parallelism of multiple core CPUs and GPUs.

\citet{item192} evaluated experimentally many of the route planning techniques described here, using the well-known continent sized benchmark Western Europe 
and a simplified model of real-world road networks. Table 1 shows the results of their experimental analysis. For each technique, Table 1 presents the total memory space usage, total preprocessing time, number of vertices scanned by a query on average, and average query time.

\begin{table*}[htb]
\centering
\caption{Performance of Route Planning Techniques on Western Europe \citep{item192}}
\begin{tabular}{|c|c|c|c|c|}
\hline
\multirow{2}{*}{Algorithm} & \multicolumn{2}{c|}{Data Structures} & \multicolumn{2}{c|}{Queries} \\ \cline{2-5} 
 & Space (GB) & Time (h:m) & Scanned Vertices & Time ($\mu$s)\\ \hline                 
Dijstra           & 0.4       & --     & 9,326,696 & 2,195,080 \\ \hline
Bidirect. Dijstra & 0.4       & --     & 4,914,804 & 1,205,660 \\ \hline
CRP               & 0.9       & 1:00   & 2,766     & 1,650 \\ \hline
Arc Flags         & 0.6       & 0:20   & 2,646     & 408 \\ \hline
CH                & 0.4       & 0:05   & 280       & 110 \\ \hline
CHASE             & 0.6       & 0:30   & 28        & 5.76 \\ \hline
HLC               & 1.8       & 0:50   & --        & 2.55 \\ \hline
TNR               & 2.5       & 0.22   & --        & 2.09 \\ \hline
TNR+AF            & 5.4       & 1:24   & --        & 0.70 \\ \hline
HL                & 18.8      & 0.37   & --        & 0.56 \\ \hline
HL-$\infty$       & 17.7      & 60:00  & --        & 0.25 \\ \hline
Table Lookup      & 1,208,358 & 145:30 & --        & 0.06 \\ \hline
\end{tabular}
\end{table*}

\subsection{Path Planning}

The \textbf{Path Planner} subsystem computes a set of Paths,  \( = \{ P_{1}, P_{2}, \ldots, P_{ \vert P \vert } \)  $ \} $, considering the current route, the self-driving car’s State, the internal representation of the environment, as well as traffic rules. A Path  \( P_{j}= \{ p_{1}, p_{2}, \ldots, p_{ \vert P \vert } \}  \) is a sequence of poses  \( p_{i} = \left( x_{i}, y_{i}, \theta _{i} \right)  \), which are car positions and respective orientations in the Offline Maps. The first poses of the paths of  \( P \)  is the current self-driving car position and orientation, and these paths extend some tens or hundreds of meters away from the self-driving car current position. 

Various methods for path planning have been proposed in the literature. We review those that were evaluated experimentally using real-world self-driving cars. For more comprehensive reviews on these methods readers are referred to \citet{item207} and \citet{item59}.

Methods for path planning can be mainly categorized into two classes: graph search based and interpolating curve based \citep{item207,item59}.

\subsubsection{Graph Search Based Techniques}

Graph search based techniques searches for the best paths between the car’s current state and a goal state in a state space represented as a graph. The goal state is a pose near a way point  \( w_{i} \)  of the current route  \( W \). These techniques discretize the search space imposing a graph on an occupancy grid map with centers of cells acting as neighbors in the search graph. The most common graph search based techniques for path planning of self-driving cars are Dijkstra, A-star, and A-star variants. 

The Dijkstra algorithm \citep{item190} finds the shortest path between the initial node and goal node of a graph. Dijkstra repeatedly examines the closest not yet examined node, adding its neighbors to the set of nodes to be examined, and stops when the goal node is attained. \citet{item208} used the Dijkstra algorithm to compute a path (and also a route; see Section 4.1) for the self-driving car Verdino. \citet{item11} employ Dijkstra to construct a path for the self-driving car Odin  to navigate toward and reversing out of a parking spot. \citet{item209} used Dijkstra to generate a path (and also a route), which were tested only in computer simulations.

The A-star algorithm  \citep{item191} is an extension of Dijkstra, which performs fast graph search by assigning weights to nodes based on a heuristic-estimated cost to the goal node. \citet{item210}  employed the A-star algorithm to build a path for the self-driving car Rocky, which participated in the 2005 DARPA Grand Challenge.  \citet{item211}  proposed a path planning that combines A-star with two different heuristic cost functions, namely Rotation Translation Rotation (RTR) metric and Voronoj graph, to compute a path. The first one accounts for kinematics constraints of the car, while the second one incorporates knowledge of shapes and positions of obstacles. The method was tested in the robotic car AnnieWAY, which participated of the 2007 DARPA Urban Challenge.

Other authors propose variations of the A-star algorithm for path planning.  \citet{item9}  proposed the anytime D-star to compute a path for the self-driving car Boss (Carnegie Mellon University’s car that claimed first place in the 2007 DARPA Urban Challenge).  \citet{item212}  proposed the hybrid-state A-star to compute a path for the robotic car Junior  \citep{item10}  (Stanford University’s car that finished in second place in the 2007 DARPA Urban Challenge). Both anytime D-star and hybrid-state A-star algorithms merge two heuristics – one non-holonomic, which disregards obstacles, and the other holonomic, which considers obstacles – and were used for path planning in an unstructured environment (parking lot). \citet{item213} proposed a variation of A-star to build a path that considers car’s kinematic constraints, which ignores the resolution of grid cells and creates a smooth path. \citet{item214} proposed a variation of A-star to compute a path that accounts for kinematic constraints of the autonomous vehicle Kaist.

\subsubsection{Interpolating Curve Based Techniques}

Interpolating curve based techniques take a previously known set of points (e.g., waypoints of a route) and generate a new set of points that depicts a smooth path. The most usual interpolating curve based techniques for path planning of self-driving cars are spline curves. 

A spline is a piecewise polynomial parametric curve divided in sub-intervals that can be defined as polynomial curves. The junction between each sub-segment is called knot (or control points), which commonly possess a high degree of smoothness constraint. This kind of curve has low computational cost, because its behavior is defined by the knots. However, its result might not be optimal, because it focuses more on achieving continuity between the parts instead of meeting road’s constraints \citep{item207}. \citet{item215} and 
\citet{item216} use cubic spline curves for path planning. Both of them construct a center line from a route extracted from a road network. They generate a series of parametric cubic splines, which represent possible path candidates, using the arc length and offset to the center line. The optimal path is selected based on a function that takes into account the safety and confort.

\subsection{Behavior Selection}

The \textbf{Behavior Selector} subsystem is responsible for choosing the current driving behavior, such as lane keeping, intersection handling, traffic light handling, etc., by selecting a Path,  \( P_{j} \), in  \( P \), a pose,  \( p_{g} \), in  \( P_{j} \), and the desired velocity at this pose. The pair  \( p_{g} \)  and associated velocity is called a  \( Goal_{g}= \left( p_{g},v_{g} \right)  \). The estimated time between the current state and the  \( Goal_{g} \)  is the \textit{decision horizon}. The Behavior Selector chooses a Goal considering the current driving behavior and avoiding collisions with static and moving obstacles in the environment within the decision horizon time frame.

The Behavior Selector raises complex discussions about ethical decisions. What the system will choose to do in an inevitable accident? The priority is the passenger's safety or pedestrian's safety? These questions are not covered here because, in the literature, the approaches for real world self-driving cars didn’t achieve the autonomy level to consider these problems. Nevertheless, there is some research on the subject. Readers can refer to \citep{item217, item218, item219}.

A self-driving car has to deal with various road and urban traffic situations, and some of them simultaneously (e.g. yield and merge intersections). Regarding that, in the literature the behavior selection problem can be divided according to the different traffic scenarios in order to solve them in parts. At DARPA Urban Challenge \citep{item8} the principal methods used for different driving scenarios were combination of heuristics \citep{item9}, decision trees \citep{item220}, and Finite State Machines (FSM) \citep{item10}. These methods perform well in limited, simple scenarios. Moreover, State Machines based methods have been improved and fused with other methods to cope with a larger variety of real urban traffic scenario \citep{item72, item109, item221, item71}. Ontologies can also be used as a tool to model traffic scenarios \citep{item222,item223}. Several approaches for behavior selection consider uncertainty in the intentions and trajectories of moving objects. For that, they use probabilistic methods, such as Markov Decision Processes (MDPs) \citep{item24}, and Partially-Observable Markov decision process (POMDP) \citep{item225, item226, item227}. 

\subsubsection{FSM-Based Techniques}

In FSM methods, a rule-based decision process is applied to choose actions under different traffic scenarios. The behavior is represented by states and transitions are based on discrete rules stem from perception information. The current state defines the car’s present behavior. The major drawback of this method is the difficulty to model the uncertainties and complex urban traffic scenarios. One of the successful uses of FSM for urban road traffic was that employed by the Junior team in DARPA Urban Challenge \citep{item8}. They used a FSM that had states for several urban traffic scenarios. However, that competition presented a limited set of urban scenarios, if compared with real world traffic, i.e. fewer traffic participants, no traffic lights and controlled traffic jams. In a more complex scenario, the vehicle A1 \citep{item72} used a FSM for selecting a driving strategy among predefined rules so as to follow traffic regulations (e.g., lane keeping, obeying traffic lights, and keeping the car under speed limits). 

In the IARA’s behavior selector subsystem, a FSM is used for each scenario. It receives as input: the map; the car’s current state; the current path; the perception system information, such as traffic light state \citep{item228}, moving objects detections, pedestrians in crosswalk \citep{item229, item230}, etc.; and a set of map annotations, such as speed bumps, security barriers, crosswalks and speed limits. For each scenario, a finite state machine checks some state transition rules and defines the appropriated next state. Then, considering the current state and using several heuristics, it defines a goal state in the path within the decision horizon and adjusts the goal state’s velocity in order to make the car behave properly according to the scenario, i.e. stopping on red traffic lights, reducing the velocity or stopping in busy crosswalks, etc. Figure 7 shows the IARA’s FSM for handling pedestrians in crosswalks \citep{item229}.

\begin{figure}[htb!]
\centering
\includegraphics[width=0.4\textwidth]{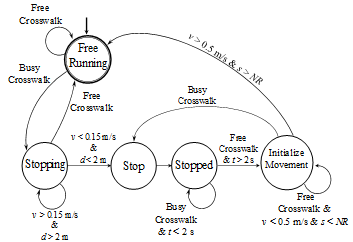}
\caption{IARA’s behavior selector state machine for handling pedestrians in crosswalks \citep{item229}. At the image, \textit{v} is the car’s velocity, \textit{d} is the distance between the car and the crosswalk stop line, \textit{t} is the time while the crosswalk is free, \textit{s} is the car’s current position, and \textit{NR} is the non-return point that is after the crosswalk stop line.}
\end{figure}

\citet{item109} modeled the behavior selection process as a network of hybrid deterministic automata and decision-trees. In order to reduce the complexity of the model, the driving task is divided into a finite set of lateral and longitudinal guidance states. The lateral guidance states consist of lane keeping and lane change states. The longitudinal guidance consists of cruise controls states and a single critical control state. The behavior selection process runs through several hierarchical levels evaluating the present situation and selecting the appropriate driving maneuver. The system was tested in a real world scenario by the BMW Group Research and Technology.

\citet{item221} combine FSM with a support vector machine (SVM) to build a classifier for a high-level behavior selector process in roundabouts situations. The decision process consists of two levels. First the SVM classifier maps the current robot state and perception data to an action, then a FSM process the action in order to output high-level commands. The SVM was trained by demonstration: a human driver annotates the situations where the driver must stop or go through the roundabout considering the predictions of other vehicles behavior. They performed experiments in the real world with an overall success rate of 98.2$\%$. The authors claim that this approach can also be applied in similar autonomous driving situations. 

To cope with more complex scenarios, \citet{item71} select the behavior using a hierarchical, concurrent state machine. This state machine was modeled using a state chart notation (Harel state chart \citep{item231}) that allows concurrent states. Their behavior selector subsystem generates a series of constraints that are derived from these concurrent state machines and used as input of a trajectory optimization problem. These constraints are formulated by the behavior selector subsystem considering information such as the characteristics of the driving corridor, the presence of static and moving objects, and the yield and merge rules. This approach was tested in a real autonomous car, named Bertha, that traveled 104 km fully autonomously through urban areas and country roads.

\subsubsection{Ontology-Based Techniques}

Ontologies are frameworks for knowledge representation that can be used to model concepts and their relationships. \citet{item222} used an ontology-based Knowledge Base to model traffic regulations and sensor data in order to help self-driving cars to understand the world. To build their decision-making system, they constructed the ontology-based Knowledge Base manually. They focused on traffic situations that occur in intersections and narrow roads. The system makes decisions considering regulations such as Right-of-Way rules, and sends decisions such as $``$Stop, $``$ToLeft$"$, or $``$Give Way$"$  to a path planning system to change the route or stop to avoid collisions. The disadvantage of this approach is the need to design an accurate world model composed of items such as mapped lanes and traffic rules at each location, which is usually done manually by humans.

In a more recent work, \citet{item223} improved on their previous work in order use only a small portion of the Knowledge Base, making it 1/20$ \sim $ 1/10 smaller than in the previous work – this increased the performance in terms of time making the system almost 10 times faster.

\subsubsection{Markov Decision Processes Based Techniques }

The Partially Observable Markov Decision Process (POMDP) framework not only addresses the uncertainty in actions transition between states, but also the uncertainty in perception. This algorithm generalizes the value iteration algorithm to estimate an optimal control policy \citep{item61}. \citet{item225} apply online POMDP to behavior selection in order to perform lane changes while driving in urban environments. To reduce the complexity of the POMDP and allow real-time decision making, the proposed algorithm was divided into two steps. In the first step, the situation is evaluated by a signal processing network. This network is a graph that considers relative distances, velocities and time to collisions with objects around the vehicle and outputs whether a lane change is possible or not. With the network output, the POMDP decision making algorithm plans online only for the current belief state. This is made by searching only the belief states that are reachable from the current state. Hence, the POMDP model only needs to have eight states. The algorithm was evaluated in real world inner city traffic showing decision coherency while maintaining a high level of decision quality.

\citet{item226} used a continuous POMDP approach to reason about potentially hidden objects and observation uncertainty, considering the interactions of road participants. They used the idea of assuming finite sets of policies to speed up planning. In the first step, a reward function aims to optimize comfort and efficiency by returning a cost for acceleration and deceleration to reach a goal area. This step only depends on the vehicle’s state and a previously defined goal. In the second step, the other traffic participants are considered by adding a higher cost for collision with other road users. The two steps costs are summed up in a reward function with a scalar value that represents the driving objective. Their approach was evaluated in a merging scenario. The policy for this scenario and all possible outcomes of it were pre-computed offline.

\citet{item227} proposed an integrated inference and behavior selection approach that models vehicle behavior and nearby vehicles as a discrete set of policies. In this work, they used a set of hand-engineering policies to in-lane and intersection driving situations. They used Bayesian changepoint detection on the history of other vehicles to infer possible future actions, estimating a distribution over potential policies that each nearby vehicle might be executing. Then, the behavior selection algorithm executes the policy with the maximum reward value by approximating the POMDP solution that evaluates the policies of the predicted vehicle through forwarding simulation. However, they assume that majority of the driving participants behave in a regular, predictable manner, following traffic rules. The experiments were carried out using an autonomous car platform.

Instead of a set of policies, \citet{item289} proposed an intersection behavior selection for Autonomous Vehicles using multiple online decision-components with interacting actions (MODIA). MODIA also consider the traffic participants but model them as separate individual Markov decision processes (MDP). Each MDP maintains its own belief and proposed action to take at each time step, generating a set of estimated actions. A lexicographic executor action function (LEAF) only executes the best (in terms of preference) action from this set (e.g. stop actions have preference). Actions are simply stop, edge or go, and encode movements by assigning desired velocity and goal points along the AV’s trajectory. MODIA remains tractable by growing linearly in the number of decisions process instantiated. This method was tested in a real autonomous car in intersection scenarios and compared with an ignorant and naive baseline algorithm successfully solving the autonomous vehicle interaction in intersection scenarios.

\subsection{Motion Planning}

The \textbf{Motion Planner} subsystem is responsible for computing a Trajectory,  \( T \), from the current self-driving car’s State to the current Goal. This trajectory must follow the Path,  \( P_{j} \), defined by the Behavior Selector subsystem as close as possible, while satisfying car’s kinematic and dynamic constraints, and providing safety and comfort to the passengers. There are two basic forms of defining a trajectory. It may be defined as (i) a sequence of commands, i.e.  \( T^{c} = \{ c_{1}, c_{2}, \ldots, c_{ \vert T \vert } \}  \), where each command  \( c_{k}= \left( v_{k}, \varphi _{k}, \Delta t_{k} \right)  \)  and, for each  \( c_{k} \),  \( v_{k} \)  is the desired velocity at time  \( k \),  \(  \varphi _{k} \)  is the desired steering angle at time  \( k \), and  \(  \Delta t_{k} \)  is the duration of  \( c_{k} \); or it may be defined as (ii) a sequence of states  \( T^{s}= \{ s_{1}, s_{2}, \ldots, s_{ \vert T \vert } \}  \), where each state  \( s_{k}= \left( p_{k}, t_{k} \right)  \)  and, for each  \( s_{k} \),  \( p_{k} \)  is a pose, and  \( t_{k} \)  is the absolute time in which this pose is expected to be achieved. A Trajectory takes the car from its current State to the current Goal smoothly and safely.

Several techniques for motion planning have been proposed in the literature. We review those that were designed for on-road motion planning and were evaluated experimentally using real-world self-driving cars. On-road motion planning aims at planning trajectories that follow the route, which differs from unstructured motion planning, in which there are no lanes and, thus, trajectories are far less constrained. For more comprehensive reviews on methods for motion planning readers are referred to \citet{item207} and \citet{item59}.

Methods for motion planning can be mainly categorized into four classes: graph search based, sampling based, interpolating curve based, and numerical optimization based  methods \citep{item207,item59}. 

\subsubsection{Graph Search Based Techniques}

Graph search based techniques for trajectory planning extend those for path planning (Section 4.2) in order to specify the evolution of car’s state over time. The most common graph search based trajectory planning techniques for self-driving cars are state lattice, Elastic Band (EB), and A-star. 

A state lattice \citep{item232} is a search graph in which vertices denote states and edges denote paths that connect states satisfying the robot’s kinematic constraints. Vertices are placed in a regular fashion, such that the same paths, although rigidly translated and rotated, can be used to build a feasible path from each of the vertices to the goal. In this way, a trajectory to the goal is likely to be represented as a sequence of edges in the graph. State lattices are able to handle several dimensions, such as position, velocity, and acceleration, and are suitable for dynamic environments. However, they have high computational cost, because they evaluate every possible solution in the graph \citep{item207}. 

\citet{item233} propose a conformal spatiotemporal state lattice for trajectory planning. They construct the state lattice around a centerline path. They define nodes on the road at a lateral offset from the centerline and compute edges between nodes using an optimization algorithm. This optimization algorithm finds the parameters of a polynomial function that define edges connecting any pairs of nodes. They assign to each node a state vector that contains a pose, acceleration profile, and ranges of times and velocities. The acceleration profile increases trajectory diversity at a less cost than would the finer discretization of time and velocity intervals. Furthermore, the ranges of times and velocities reduce computational cost by allowing the assignment of times and velocities to the graph search phase, instead of graph construction phase. \citet{item234} propose an iterative optimization that is applied to the resulting trajectory derived from the state lattice trajectory planning proposed by \citet{item233}, which reduces the planning time and improve the trajectory quality. \citet{item235} propose a planning method that fuses the state lattice trajectory planning proposed by \citet{item233} with a tactical reasoning. A set of candidate trajectories is sampled from the state lattice, from which different maneuvers are extracted. The final trajectory is obtained by selecting a maneuver (e.g., keep lane or change lane) and choosing the candidate trajectory associated with the selected maneuver. \citet{item236} builds a state lattice by generating candidate paths along a route using a cubic polynomial curve. A velocity profile is also computed to be assigned to poses of the generated paths. The resulting trajectories are evaluated by a cost function and the optimal one is selected. 

Another graph search based technique is the elastic band, which was first used in path planning and later in trajectory. An elastic band is a search graph with elastic vertices and edges. An elastic vertex is defined by augmenting the spatial vertex with an in-edge and out-edge that connect the neighboring spatial vertices. The path is found by an optimization algorithm that balances two forces – repulsive forces generated by external obstacles and contractive forces generated by the neighboring points to remove band slackness. This technique demonstrates continuities and stability, it has non-deterministic run-time, but requires a collision-free initial path. 

\citet{item237} propose a decoupled space-time trajectory planning method. The trajectory planning is divided into three phases. In the first phase, a collision-free feasible path is extracted from an elastic band, considering road and obstacles constraints, using a pure-pursuit controller and a kinematic car model (see Section 4.6.2). In the second phase, a velocity profile is suggested under several constraints, namely speed limit, obstacle proximity, lateral acceleration and longitudinal acceleration. Finally, given the path and velocity profile, trajectories are computed using parametric path spirals. Trajectories are evaluated against all static and moving obstacles by simulating their future movements.

The A-star algorithm is commonly used for path planning (Section 4.2). \citet{item238} propose two novel A-star node expansion schemes for trajectory planning. The first scheme tries to find a trajectory that connects the car’s current node directly to the goal node using numerical optimization. The second scheme uses a pure-pursuit controller to generate short edges (i.e., short motion primitives) that guide the car along the global reference path.

\subsubsection{Sampling Based Techniques}

Sampling based techniques randomly sample the state space looking for a connection between the car’s current state and the next goal state. The most used sampling based technique for trajectory planning of self-driving cars is Rapidly-exploring Random Tree (RRT).

RRT methods for trajectory generation \citep{item239} incrementally build a search tree from the car’s current state using random samples from the state space. For each random state, a control command is applied to the nearest vertex of the tree in order to create a new state as close as possible to the random state. Each vertex of the tree represents a state and each directed edge represents a command that was applied to extend a state. Candidate trajectories are evaluated according to various criteria. 
Figure 8 shows an example of a trajectory generated by a RRT method from the current car’s State to a random state. 
RRT methods have low computational cost for high-dimensional spaces and always find a solution, if there is one and it is given enough time. However, its result is not continuous and jerky \citep{item207}. 

\begin{figure}[htb!]
\centering
\includegraphics[width=0.5\textwidth]{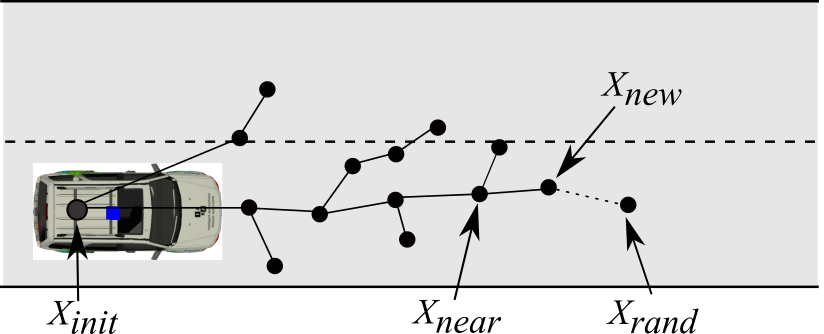}
\caption{Example of a trajectory generated by a RRT method, where  \( X_{init} \)  is the car’s State,  \( X_{rand} \)  is the random state,  \( X_{near}  \) is the nearest vertex of the tree to  \( X_{rand} \)  and  \( X_{new} \)  is the new state as close as possible to  \( X_{rand} \)}
\end{figure}

\citet{item240} propose a RRT method for trajectory planning of the self-driving car IARA. They present new variants to the standard RRT in order to bias the location of random states toward the path, select the most promising control commands to extend states, discard non-promising states, and choose the set of states that constitutes the best trajectory. It also reuses part of the trajectory built in the previous planning cycle. \citet{item241}  propose a RRT method that uses the drivers’ visual search behavior on roads to guide the state sampling of RRT. Drivers use both a $``$near point$"$  and a $``$far point$"$  to drive through curved roads. They exploit this characteristic of drivers’ visual search behavior shown on curved roads to guide the RRT method. Furthermore, they adopt a post-processing, based on B-splines, to generate smooth, continuous, and feasible trajectories.

\subsubsection{Interpolating Curve Based Techniques}

Interpolating curve based techniques interpolate a previously known set of poses (e.g., the path’s poses) and build a smoother trajectory that considers car’s kinematic and dynamic constraints, comfort, obstacles, among other parameters. The most common interpolating curve based techniques for trajectory planning of self-driving cars are clothoid curves \citep{item207}.

A clothoid curve allows defining trajectories with linear variable curvature, so that transitions between straight to curved segments are smooth. However, a clothoid curve has high computational cost, because of the integrals that define it, and it depends on global waypoints \citep{item207}. \citet{item242} use clothoid tentacles for trajectory planning. Tentacles are computed for different speeds and different initial steering angles, starting from the car’s center of gravity and taking the form of clothoids. Tentacles are classified as navigable or not navigable using a path tracking controller that derives the steering angle commands from the tentacles’ geometric parameters and running a collision-check on an occupancy grid map. Among the navigable tentacles, the best tentacle is chosen based on several criteria. \citet{item243,item244} also use clothoid tentacles for trajectory planning. However, they use an approach inspired in Markov Decision Process (MDP) to choose the best tentacle.

\subsubsection{Numerical Optimization Based Techniques}

Numerical optimization based techniques minimize or maximize a function with constrained variables. The most common numerical optimization based techniques for trajectory planning of self-driving cars are function optimization and model-predictive methods.

Function optimization methods find a trajectory by minimizing a cost function that considers trajectory’s constraints, such as position, velocity, acceleration, and jerk. These methods can easily consider car’s kinematic and dynamic constraints and environment’s constraints. However, they have high computational cost, because the optimization process takes place at each motion state, and depend on global waypoints \citep{item207}. \citet{item245} uses a function optimization method for trajectory planning of the self-driving car Bertha. They find the optimal trajectory that minimizes a cost function and obeys trajectory’s constraints. The cost function is composed of a set of terms that make the trajectory to follow the middle of the driving corridor (which is equivalent to a path) at a specified velocity, penalize strong acceleration, dampen rapid changes in acceleration, and attenuate high yaw rates. 

Model predictive methods for trajectory planning \citep{item246} produce dynamically feasible control commands between car’s current state and next goal state by using models that predict future states. They can be used to solve the problem of generating parameterized control commands that satisfy state constraints whose dynamics can be expressed by differential equations for example. \citet{item247} uses a model-predictive method for trajectory planning of the self-driving car Boss (Carnegie Mellon University’s car that claimed first place in the 2007 DARPA Urban Challenge) \citep{item9}. They generate trajectories to a set of goal states derived from the centerline path. To compute each trajectory, they use an optimization algorithm that gradually modifies an initial approximation of the trajectory control parameters until the trajectory end point error is within an acceptable bound. The trajectory control parameters are the trajectory length and three knot points of a spline curve that defines the curvature profile. A velocity profile is generated for each trajectory based on several factors, including the velocity limit of the current road, maximum feasible velocity, and goal state velocity. The best trajectory is selected according to their proximity to obstacles, distance to the centerline path, smoothness, end point error, and velocity error.

\citet{item248} use a state sampling-based trajectory planning scheme that samples goal states along a route. A model-predictive path planning method is applied to produce paths that connect the car’s current state to the sampled goal states. A velocity profile is used to assign a velocity to each of the states along generated paths. A cost function that considers safety and comfort is employed to select the best trajectory. 

\citet{item55} used a model-predictive method for trajectory planning of the self-driving car IARA. To compute the trajectory, they use an optimization algorithm that finds the trajectory control parameters that minimize the distance to the goal state, distance to the path, and proximity to obstacles. The trajectory control parameters are the total time of the trajectory, \(  tt \), and three knot points,  \( k_{1} \),  \( k_{2} \)  and  \( k_{3} \), that define a cubic spline curve which specifies the steering angle profile, i.e., the evolution of the steering angle during  \( tt \). Figure 9 shows an example of a trajectory from the current car’s State (in blue) to the current Goal (in yellow), that follows the path (in white) and avoids obstacles (in shades of grey). Figure 10 shows the steering angle profile of the trajectory illustrated in Figure 9. 

\begin{figure}[htb!]
\centering
\includegraphics[width=0.4\textwidth]{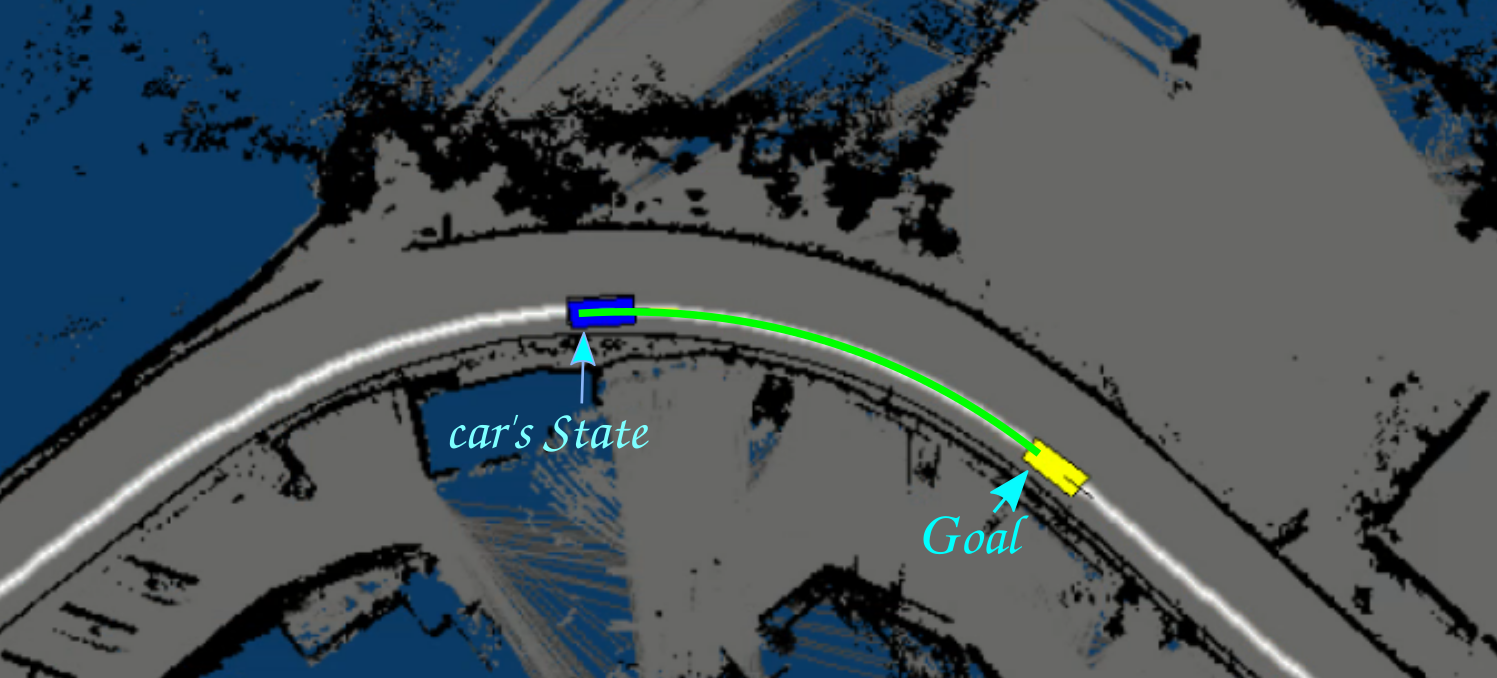}
\caption{Example of a trajectory from the current car’s State (blue rectangle), to the Goal (yellow rectangle), which is illustrated in an excerpt of an OGM. The path is indicated in white and the trajectory in green. In the OGM, the shades of grey indicate the occupancy probability: black corresponds to occupied with high certainty and white to free with high certainty; the light blue indicates unknown or non-sensed regions.}
\end{figure}

\begin{figure}[htb!]
\centering
\includegraphics[width=0.4\textwidth]{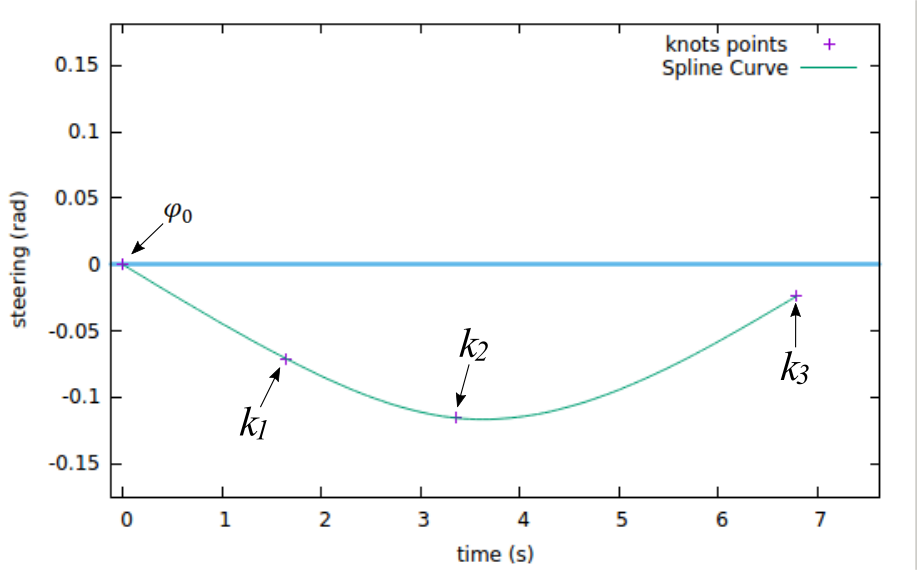}
\caption{Spline curve, defined by  \(  \varphi _{0} \), and  \( k_{1} \),  \( k_{2} \)  and  \( k_{3} \), that specifies the evolution of the steering angle during the total time of the trajectory,  \( tt=6.7 \)  s. The knot point  \( k_{1} \)  is defined at  \( t=tt/4 \),  \( k_{2} \)  at  \( t=tt/2 \)  and  \( k_{3} \)  at  \( t=tt \).}
\end{figure}

\subsection{Obstacle Avoidance}

The \textbf{Obstacle Avoider} subsystem receives the Trajectory computed by the Motion Planner subsystem and changes it (typically reducing the velocity), if necessary, to avoid collisions. There is no much literature on methods for performing the functions of the Obstacle Avoider subsystem. 

\citet{item56} proposes an obstacle avoider subsystem that, at each motion plan cycle, receives an online map that represents the environment around the car, the current state of the self-driving car in the online map, and the trajectory planned by the motion planner subsystem. The obstacle avoider subsystem simulates the trajectory and, if a crash is predicted to occur in the trajectory, the obstacle avoider decreases the linear velocity of the self-driving car to prevent the crash. Figure 11 shows the block diagram of the obstacle avoider.

\begin{figure}[htb!]
\centering
\includegraphics[width=0.4\textwidth]{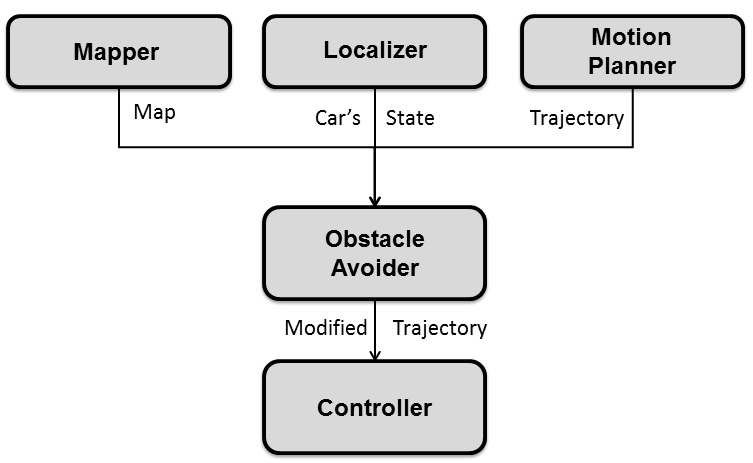}
\caption{Block diagram of the obstacle avoider system proposed by \citet{item56}. The system receives as input the online map, current car’s state and the trajectory and simulates the trajectory. If any crash occurs, the obstacle avoider system decreases the linear velocity of the self-driving car.}
\end{figure}

\citet{item249} uses RADAR data to calculate the distance between a self-driving car and the obstacles in the environment. After that the time to collision (TTC) is calculated and compared with a distance threshold. If TTC exceeds the threshold the system trigger a decision-maker subsystem that decides the self-driving car behavior (lane keeping, lane changing, speeding up, adaptive following and emergency braking).

\citet{item250} proposed a system to avoid collisions based on a hierarchical control architecture that consists of a decision making layer, which contains a dynamic threat assessment model that continuously analyses the risk associated with collision, and a path planner that calculates a collision-free path that considers the kinematics and dynamics constraints of the self-driving car when an emergency situation happens.

\subsection{Control}

The \textbf{Controller} subsystem receives the trajectory generated by the Motion Planner subsystem, eventually modified by the Obstacle Avoider subsystem, and computes and sends Effort commands to the actuators of the steering wheel, throttle and brakes of the self-driving car in order to make the car execute the trajectory as best as the physical world allows. As mentioned in Section 4.4, there are two basic forms of defining a trajectory: (i) as a sequence of commands, i.e. the trajectory  \( T^{c}= \{ c_{1},c_{2}, \ldots,c_{ \vert T^{c} \vert } \}  \), where each command  \( c_{k}= \left( v_{k}, \varphi _{k}, \Delta t_{k} \right)  \)  and, for each  \( c_{k} \),  \( v_{k} \)  is the desired velocity at time  \( k \),  \(  \varphi _{k} \)  is the desired steering angle at time  \( k \), and  \(  \Delta t_{k} \)  is the duration of  \( c_{k} \); or (ii) as a sequence of states  \( T^{s}= \{ s_{1},s_{2}, \ldots,s_{ \vert T^{s} \vert } \}  \), where each state  \( s_{k}= \left( p_{k},t_{k} \right)  \)  and, for each  \( s_{k} \),  \( p_{k} \)  is a pose, and  \( t_{k} \)  is the absolute time in which this pose is expected to be achieved. The implementations of the Controller subsystem that receives a  \( T^{c} \)  trajectory can be classified as direct hardware actuation control methods, while those that receive a  \( T^{s} \)  trajectory can be classified as path tracking methods.

\subsubsection{Direct Hardware Actuation Control Methods}

Direct hardware actuation control methods compute effort inputs to the steering, throttle, and brakes actuators of the car directly from the trajectory computed by the motion planner subsystem (Section 4.2), and try and mitigate inaccuracies caused primarily by the model of how the efforts influence the self-driving car velocity, \(  v \), and steering wheel angle, \(   \varphi  \).

One of the most common direct hardware actuation control methods for self-driving cars is feedback control. It involves applying the efforts, observing  \( v \)  and  \(  \varphi  \), and adjusting future efforts to correct errors (the difference between the current  \( T^{c} \)   \( v_{k} \)  and  \(  \varphi _{k} \)  and the measured  \( v \)  and  \(  \varphi  \) ). Figure 12 shows the diagram of feedback control, where the controller block may be composed by a variety of controllers and the system block is composed by the system plant to be controlled.

\begin{figure}[htb!]
\centering
\includegraphics[width=0.4\textwidth]{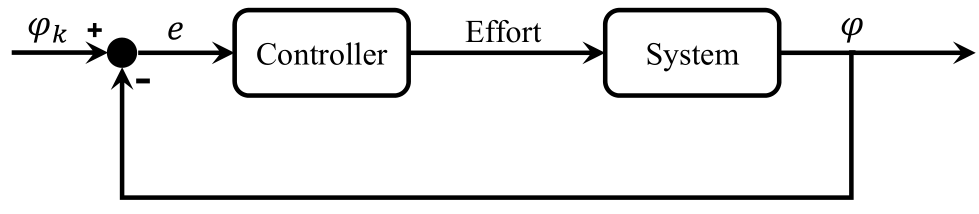}
\caption{Diagram of the feedback control}
\end{figure}

\citet{item251} used a feedback control method to implement the controller subsystem of an Audi TTS. The hardware structures of the Audi TTS were designed to achieve hard real-time control at 200 Hz. This high-speed hardware enabled their controller subsystem to drive the car at up to 160 km/h. \citet{item122} employed a feedback control method for the controller subsystem of Bertha. Bertha’s controller subsystem was able to drive the car at up to 100 km/h through the 103 km long Bertha Benz Memorial Route. \citet{item248} adopted the same control method in a Toyota Land Cruiser, which was able to drive the car at up to 25 km/h. 

Another popular hardware actuation method for self-driving cars is the Proportional Integral Derivative (PID) control method. It involves specifying a desired hardware input and an error measure that accounts for how far the hardware output is from the desired hardware input \citep{item252}. In possession of this information, a PID control method computes the hardware input efforts, which is  \( K_{p} \)  times the error measure (proportional to the error according to  \( K_{p} \) ), plus  \( K_{i} \)  times the integral of the error, plus  \( K_{d} \)  times the derivative of the error, where  \( K_{p} \),  \( K_{i} \)  and  \( K_{d} \)  are parameters of the PID control method. Similar to the feedback control technique, the PID control method can deal with the current error, besides the accumulated past errors over time \citep{item57}. \citet{item253} used an adaptive PID method based on the generalized minimum variance technique to implement the controller subsystem of the self-driving car Intelligent Pionner. Intelligent Pionner’s controller subsystem was able to drive the car at up to 60 km/h. 

\citet{item254} employed a mixture of a Model Predictive Control (MPC) method and a feedforward PID control technique for the controller subsystem of the robotic car Junior \citep{item10} (Stanford University’s car that finished in second place in the 2007 DARPA Urban Challenge). The controller subsystem receives as input  \( T^{s} \)  and computes the efforts to actuate on the throttle, braking and steering. MPC methods use models of how self-driving cars respond to efforts over time to predict future values of  \( v \)  or  \(  \varphi  \). They use these predictions to try and decide which efforts should be sent to the car now to better reduce current and future errors in  \( v \)  or  \(  \varphi  \). \citet{item254} mixture of MPC and PID was able to drive Junior up to 56 km/h. \citet{item57} proposed a Neural based Model Predictive Control (N-MPC) method to tackle delays in the steering wheel hardware of the self-driving car IARA. They used the MPC method to reduce the impact of steering hardware delays by anticipating efforts that would timely move the car according to the current trajectory. They modeled IARA’s steering hardware using a neural network and employed the neural-based steering model in the N-MPC steering control method. The proposed solution was compared to the standard solution based on the PID control method. The PID controller subsystem worked well for speeds of up to 25 km/h. However, above this speed, delays of IARA’s steering hardware were too high to allow proper operation. The N-MPC subsystem reduced the impact of IARA’s steering hardware delays, which allowed its autonomous operation at speeds of up to 37 km/h – an increase of 48$\%$  in previous maximum stable speed.

\subsubsection{Path Tracking Methods}

Path tracking methods stabilize (i.e., reduce the error in) the execution of the motion plan computed by the motion planner subsystem (Section 4.2), in order to reduce inaccuracies caused mainly by car’s motion model. They can be considered simplified trajectory planning techniques, although they do not deal with obstacles. The pure pursuit method \citep{item59} is broadly used in self-driving cars for path tracking, because of its simple implementation. It consists of finding a point in the path at some look-ahead distance from the current path and turning the front wheel so that a circular arc connects the center of the rear axle with the point in the path \citep{item59}, as shown in Figure 13. There is a set of variants proposed to improve the pure pursuit method. \citet{item288} propose to use the rear wheel position as the hardware output to stabilize the execution of the motion plan. \citet{item7} present the control approach of the self-driving car Stanley, which consists of taking the front wheel position as the regulated variable. Stanley’s controller subsystem was able to drive the car at up to 60 km/h.

\begin{figure}[htb!]
\centering
\includegraphics[width=0.4\textwidth]{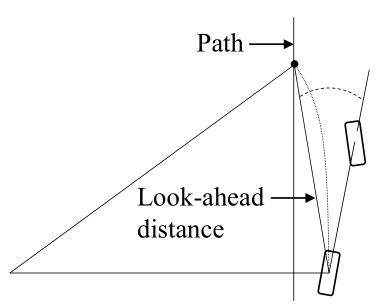}
\caption{Pure pursuit approach geometry}
\end{figure}

The Model Predictive Control (MPC) method is widely employed in self-driving cars. It consists of selecting control command inputs that will lead to desired hardware outputs, using the car’s motion model to simulate and optimize outputs over a prediction horizon in the future \citep{item57}. \citet{item255} use the MPC method for path tracking of a small electrical car. They employ the standard bicycle model to predict car’s motion. The electrical car’s controller subsystem was able to drive the car at up to 20 km/h. \citet{item256} present the control approach of an Audi TTS, in which a dynamic motion model of the car \citep{item257,item258} is used to enable path tracking at the limits of handling. The car’s dynamic motion model considers the deviation from the path plan due to tire slippage. Experimental results demonstrated that the Audi TTS could achieve a speed of up to 160 km/h.

\section{Architecture of the UFES’s Car, IARA}

In this section, to put everything in context, we present a detailed description of the architecture of a research self-driving car, the Intelligent Autonomous Robotic Automobile (IARA). IARA (Figure 14) follows the typical architecture of self-driving cars. It was developed at the Laboratory of High Performance Computing (Laboratório de Computação de Alto Desempenho – LCAD) of the Federal University of Espírito Santo (Universidade Federal do Espírito Santo – UFES). IARA was the first Brazilian self-driving car to travel autonomously 74 km on urban roads and highways. The 74 km route from Vitória to Guarapari was travelled by IARA at the night of May 12, 2017. 

\begin{figure}[htb!]
\centering
\includegraphics[width=0.4\textwidth]{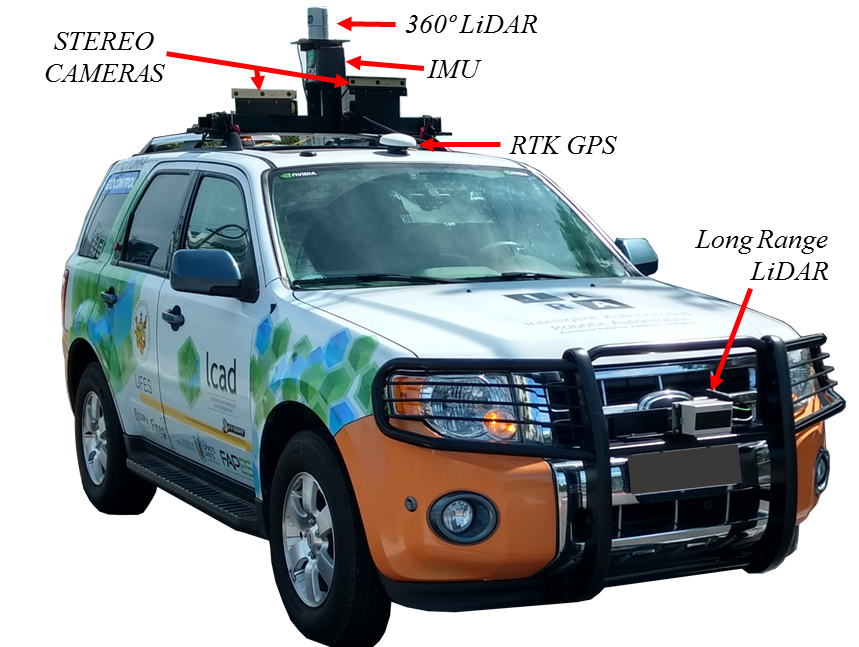}
\caption{Intelligent and Autonomous Robotic Automobile (IARA) that was the first Brazilian self-driving car to travel autonomously 74 km on urban roads and highways. A video that shows IARA operating autonomously is available at \url{https://youtu.be/iyKZV0ICysc}.}
\end{figure}

\subsection{Architecture of IARA’s Hardware}

IARA is based on a 2011 Ford Escape Hybrid, which was adapted for autonomous driving. The actuation on the steering wheel is made by the factory-installed electric power steering system, which was modified so that an added electronic module can send signals equivalent to those sent by the original torque sensor to the electric power steering system. The electronic module also sends signals equivalent to the signals the throttle pedal sends to the car’s electronics. The actuation on the brakes is made by an electric linear actuator attached to the brakes’ pedal. The breaks’ pedal can be used normally even in autonomous mode, though, thanks to the way the linear actuator is attached to it (the actuator only touches the pedal and can push it). The gear stick is originally attached to a series of electric switches that informs the car’s electronics the current gear. A series of relays we have installed allows either connecting our autonomy system to the car’s electronics or the original switches controlled by the gear stick. Similarly, the electric power steering wires and throttle pedal wires can be connected to the original car’s wiring or to our autonomy system. At the press of a button, we can select the original manual factory operation or autonomous operation. 

To power the computers and sensors that implement the autonomy system, we take advantage of the 330 Volts battery used by the electric powertrain of the Ford Escape Hybrid. We use a DC-DC converter to reduce it to 54 Volts; this DC-DC converter charges the battery of standard no-breaks that provides 120V AC inside the car. These no-breaks can also be connected to the 120V AC mains, which allow powering the computers and sensors of IARA while it is in the LCAD garage without requiring to turn on the Ford Escape Hybrid’s engine. 

IARA’s computers and sensors comprise a workstation (Dell Precision R5500, with 2 Xeon X5690 six-core 3.4GHz processors and one NVIDIA GeForce GTX-1030), networking gear, two LIDARs (a Velodyne HDL-32E and a SICK LD-MRS), three cameras (two Bumblebee XB3 and one ZED), an IMU (Xsens MTi), and a dual RTK GPS (based on the Trimble BD982 receiver).

\subsection{Architecture of IARA’s Software}

Following the typical system architecture of self-driving cars, IARA’s software architecture is organized into perception and decision making systems, as shown in Figure 1. Each subsystem of Figure 1 is implemented in IARA as one or more software modules. Figure 15 shows, as a block diagram, the software modules that implement the software architecture of IARA. IARA’s software receives data from the sensors (in yellow in Figure 15) that are made available through drivers modules (in red in Figure 15) and sends them to the filter modules (green in Figure 15). The filter modules receive as input data from sensors and / or other filter modules, and generate as output processed versions of this data. The filter modules implement, individually or in groups, the subsystems of Figure 1.

\begin{figure*}[htb!]
\centering
\includegraphics[width=0.9\textwidth]{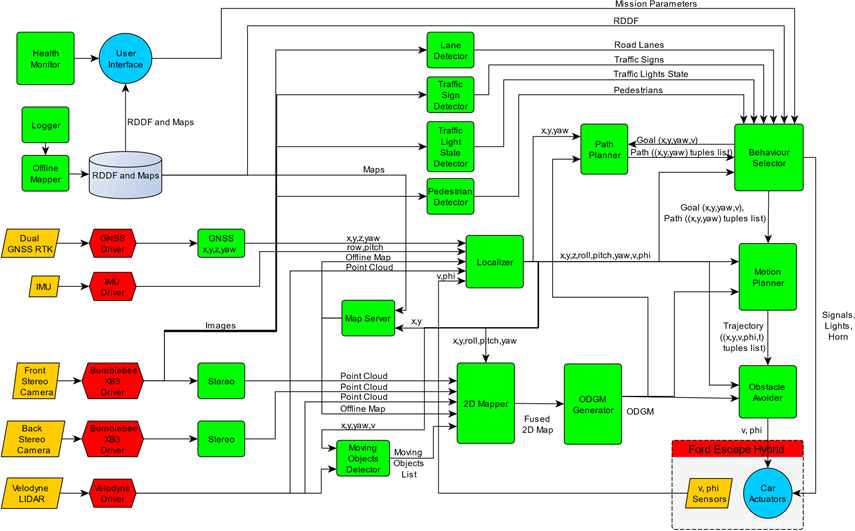}
\caption{Block diagram of the software modules that implement the software architecture of IARA}
\end{figure*}

The \textbf{GNSS} module transforms latitude and longitude data of the IARA Dual RTK GNSS into UTM coordinates (x, y, z) and synchronizes this information with the IARA orientation data (yaw - a Dual RTK GNSS has two antennas and centimeter accuracy, which makes it possible to obtain IARA’s yaw with respect to the true north of Earth). The data in this module is used by the Localizer subsystem when the global localization and by the Mapper subsystem when generating the offline map. 
The \textbf{Stereo} module \citep{item260} produces 3D depth maps (Point Cloud) from stereo images. It can be used by the 2D Mapper module as an alternative to LIDARs. The \textbf{Map Server} module \citep{item54} provides the current Offline Map, given the current IARA’s pose, extracted from the Offline Map database previously computed using the GraphSLAM algorithm. It is part of the Mapper subsystem of Figure 1. 
The \textbf{Localizer} module \citep{item62,item66} localizes IARA on the current Offline Map from IARA (v, phi) and Point Clouds data from LIDAR Velodyne, and provides IARA status with 8 dimensions (x, y, z, roll, pitch, yaw, v, phi). Note that the pose provided by GNSS is only used when IARA is initialized (during the global localization process, when it is not done by imaging \citep{item261,item77}) and the IMU data is only synchronized with the computed position. The Localizer module implements the Localizer subsystem. The \textbf{Moving Objects Detector} module \citep{item114,item229,item230}, using the Velodyne’s Point Cloud and the pose and velocity of IARA, detects moving obstacles (x, y, yaw and geometry position) and its speed. It implements the MOT subsystem. The \textbf{2D Mapper} module creates an online map of the path and static and moving obstacles around IARA (Fused 2D Map) using IARA’s status, various Point Clouds, the offline map and a list of Moving Objects List (see video at \url{https://youtu.be/cyIfGupar-8}). It is part of the Mapper subsystem. The \textbf{ODGM Generator} module generates ODGM from the Fused 2D Map. It is part of the Mapper subsystem. 

The \textbf{Lane Detector} module \citep{item179} identifies the path markings on the lane (horizontal traffic signaling, or Road Lanes in Figure 15) from the images captured by the front camera (see videos at \url{https://youtu.be/NPU9tiyA8vw} and \url{https://youtu.be/R5wdPJ4ZI5M}). It is part of the TSD subsystem. The \textbf{Traffic Sign Detector} module \citep{item263,item264} detects and recognizes \citep{item265} traffic signs along the path from images captured by the front camera (see video at \url{https://youtu.be/SZ9w1XBWJqE}). It is part of the TSD subsystem. The \textbf{Traffic Light State Detector} module \citep{item228} detects traffic lights along the path and identifies its status from images captured by the front camera (see video at \url{https://youtu.be/VjnCu3YXl2o}). It is part of the TSD subsystem. 

The \textbf{Path Planner} module, on demand of the Behavior Selector module, computes a Path ((x, y, yaw) tuple list) from the current IARA’s position to a goal (Goal (x, y, yaw, v)). It implements the Path Planner subsystem. The \textbf{Behavior Selector} module, considering information provided by several modules, establishes a goal for IARA (Goal, i.e., x, y, yaw, v) and suggests a path to this goal (Path ((x, y, yaw) tuple list)). It implements the Behavior Selector subsystem. The \textbf{Motion Planner} module \citep{item55}, from a goal (Goal (x, y, yaw, v)), a suggested path (Path ((x, y, yaw) y, z, roll, pitch, yaw, v, phi), constructs a trajectory (Trajectory ((x, y, v, phi, t) tuples list) from the current IARA’s position to the target, but considering any obstacles present in the online map (\url{https://youtu.be/tg8tkciHGzc}) It implements the Motion Planner subsystem. The \textbf{Obstacle Avoider} module \citep{item56} evaluates continuously and at high speed the path generated by the Motion Planner module, changing, if necessary, their velocities, in order to avoid eminent collisions (\url{https://youtu.be/6X\_8q73g2kA}). It implements the Obstacle Avoider subsystem. The \textbf{Ford Escape Hybrid} module \citep{item57} is the driver of the Ford Escape Hybrid car. From the outputs of the Obstacle Avoider and the Behavior Selector modules, it generates the signals that control the car. It implements the Controller subsystem. Finally, the \textbf{Health Monitor} module continuously checks for the proper functioning of all modules, automatically restarts modules that are not operating properly, and tells the user about their actions.

\section{Self-Driving Cars under Development in the Industry}

In this section, we list prominent self-driving car research platforms developed by academia and technology companies, and reported in the media. Several companies demonstrated interest in developing self-driving cars, and/or investing in technology to support and profit from them. Enterprises range from manufacturing cars and creating hardware for sensing and computing to developing software for assisted and autonomous driving, entertainment and in-car advertisement. We provide an overview of companies doing research and development in self-driving cars. The list is not presented in any specific order, since we aim at making it as impartial and complete as possible. The information was acquired by inspecting companies’ websites and news published in other media. 

Torc was one of the pioneers in developing cars with autonomous capabilities. The company was founded in 2005. In 2007, it joined Virginia Tech’s team to participate in the 2007 DARPA Urban Challenge with their self-driving car Odin \citep{item11}, which reached third place in the competition. The technology used in the competition was improved since then and it was successfully applied in a variety of commercial ground vehicles, from large mining trucks to military vehicles. 

Google's self-driving car project began in 2009 and was formerly led by Sebastian Thrun, who also led the Stanford University’s team with their car Stanley \citep{item7}, winner of the 2005 DARPA Grand Challenge. In 2016, Google's self-driving car project became an independent company called Waymo. The company is a subsidiary of the holding Alphabet Inc., which is also Google's parent company. Waymo's self-driving car uses a sensor set composed of LIDARs to create a detailed map of the world around the car, RADARs to detect distant objects and their velocities, and high resolution cameras to acquire visual information, such as whether a traffic signal is red or green. 

Baidu, one of the giant technology companies in China, is developing an open source self-driving car project with codename Apollo. The source code for the project is available in GitHub. It contains modules for perception (detection and tracking of moving obstacles, and detection and recognition of traffic lights), HD map and localization, planning, and control, among others. Several companies are partners of Baidu in the Apollo project, such as TomTom, Velodyne, Bosch, Intel, Daimler, Ford, Nvidia, and Microsoft. One of the Apollo project’s goals is to create a centralized place for Original Equipment Manufacturers (OEMs), startups, suppliers, and research organizations to share and integrate their data and resources. Besides Baidu, Udacity, an educational organization founded by Sebastian Thrun and others, is also developing an open source self-driving car project, which is available for free in GitHub. 

Uber is a ride-hailing service and, in 2015, they partnered with the Carnegie Mellon University to develop self-driving cars. A motivation for Uber's project is to replace associated drivers by autonomous software. 

Lyft is a company that provides ridesharing and on-demand driving services. Like Uber, Lyft is doing research and development in self-driving cars. The company aims at developing cars with level 5 of autonomy. 

Aptiv is one of Lyft's partners. Aptiv was created in a split of Delphi Automotive, a company owned by General Motors. Aptiv’s objective is to build cars with level 4 and, posteriorly, level 5 of autonomy. Besides other products, the company sells short-range communication modules for vehicle-to-vehicle information exchange. Aptiv recently acquired two relevant self-driving car companies, Movimento and nuTonomy.

Didi is a Chinese transportation service that bought Uber's rights in China. Didi's self-driving car project was announced in 2017 and, in February of 2018, they did the first successful demonstration of their technology. In the same month, company’s cars started being tested in USA and China. Within a year, Didi obtained the certificate for HD mapping in China. The company is now negotiating a partnership with Renault, Nissan, and Mitsubishi to build an electric and autonomous ride-sharing service. 

Tesla was founded in 2003 and, in 2012, it started selling its first electric car, the Model S. In 2015, the company enabled the autopilot software for owners of the Model S. The software has been improved since then and its current version, the so called enhanced autopilot, is now able to match speed with traffic conditions, keep within a lane, change lanes, transition from one freeway to another, exit the freeway when the destination is near, self-park when near a parking spot, and be summoned to and from the user’s garage. Tesla's current sensor set does not include LIDARs.

A Chinese company called LeEco is producing self-driving luxury sedans to compete with the Tesla Model S. The company is also backing up Faraday Future for the development of a concept car. LeEco is also partnering Aston Martin for the development of the RapidE electric car. 

Besides developing hardware for general-purpose high-per-formance computing, NVDIA is also developing hardware and software for self-driving cars. Although their solutions rely mostly on artificial intelligence and deep learning, they are also capable of performing sensor fusion, localization in HD maps, and planning. 

Aurora is a new company founded by experienced engineers that worked in Google's self-driving car project, Tesla, and Uber. The company plans to work with automakers and suppliers to develop full-stack solutions for cars with level 4 and, eventually, level 5 of autonomy. Aurora has independent partnerships with Volkswagen group (that owns Volkswagen Passenger Cars, Audi, Bentley, Skoda, and Porsche) and Hyunday. 

Zenuity is a joint venture created by Volvo Cars and Autoliv. Ericsson will aid in the development of Zenuity’s Connected Cloud that will use Ericsson's IoT Accelerator. TomTom also partnered with Zenuity and provided its HD mapping technology. TomTom’s HD maps will be used for localization, perception and path planning in Zenuity's software stack.

Daimler and Bosch are joining forces to advance the development of cars with level 4 and 5 of autonomy by the beginning of the next decade. The companies already have an automated valet parking in Stuttgart  and they also have tested the so called Highway Pilot in trucks in USA and Germany. Besides partnering with Bosch, Daimler also merged its car sharing business, Car2Go, with BMW's ReachNow, from the same business segment, in an effort to stave off competition from other technology companies, such as Waymo and Uber. The new company will include business in car sharing, ride-hailing, valet parking, and electric vehicle charging.

Argo AI founders led self-driving car teams at Google and Uber. The company received an investment of U$\$$  1 billion from Ford with the goal of developing a new software platform for Ford’s fully autonomous car (level 4) coming in 2021. They have partnerships with professors from Carnegie Mellon University and Georgia Tech. 

Renesas Autonomy develops several solutions for Automated Driving Assistant Systems (ADAS) and automated driving. The company has a partnership with the University of Waterloo. 

Honda revealed in 2017 plans for introducing cars with level 4 of autonomy by 2025. The company intends to have cars with level 3 of autonomy by 2020 and it is negotiating a partnership with Waymo. 

Visteon is a technology company that manufactures cockpit electronic products and connectivity solutions for several vehicle manufacturers. In 2018, Visteon introduced its autonomous driving platform capable of level 3 of autonomy and, potentially, higher levels. The company does not aim at producing cars and sensors, but at producing integrated solutions and software. 

AImotive is using low cost components to develop a self-driving car. Its solution relies strongly on computer vision, but it uses additional sensors. The company is already able to perform valet parking and navigate in highways. AImotive also developed a photorealistic simulator for data collection and preliminary system testing, and chips for artificial intelligence-based, latency-critic, and camera-centric systems. 

As AImotive, AutoX is avoiding to use LIDARs in its solution. However, the company is going further than Almotive and trying to develop a level 5 car without using RADARs, ultrasonics, and differential GPS’s. AutoX's approach is to create a full-stack software solution based on artificial intelligence. 

Mobileye is also seeking to develop a self-driving car without using LIDARs, but relying mostly in a single-lensed camera (mono-camera). Mobileye is one of the leading suppliers of software for Advanced Driver Assist Systems (ADAS), with more than 25 partners among automakers. Beyond ADAS, Mobileye is also developing technology to support other key components for autonomous driving, such as perception (detection of free space, driving paths, moving objects, traffic lights, and traffic signs, among others), mapping, and control. The company partnered with BMW and Intel to develop production-ready fully autonomous cars, with production launch planned for 2021.

Ambarella also does not use LIDAR, but only RADARs and stereo cameras. The company joined the autonomous driving race in 2015 by acquiring VisLAB. Different from other companies, Ambarella does not aim at becoming a tier-one supplier or selling fully autonomous driving systems. Instead, they plan to sell chips and software to automakers, suppliers, and software developers. Ambarella’s current research and development guidelines include detection of vehicles, obstacles, pedestrians, and lanes; traffic sign recognition; terrain mapping; and issues related to technology commercialization, such as system calibration, illumination, noise, temperature, and power consumption. 

Pony.ai was founded in December of 2016 and, in July of 2017, it completed its first fully autonomous driving demonstration. The company signed a strategic agreement with one of the biggest Chinese car makers, the Guangzhou Auto Group (GAC).

Navya  and Transdev  are French companies that develop self-driving buses. Navya has several of their buses being tested in Europe, Asia, and Australia. Their sensor set consists of two multi-layer 360º LIDARs, six 180º mono-layer LIDARs, front and rear cameras, odometer (wheels encoder + IMU), and a GNSS RTK. Transdev is also already demonstrating their self-driving buses for the public.

JD is a Chinese e-commerce company interested in building autonomous delivery vehicles. JD's project, started in 2016, is being developed together with Idriverplus, a Chinese self-driving car startup. 

In March, 2018, Toyota announced an investment of U$\$$  2.8 billion in the creation of a new company called the Toyota Research Institute-Advanced Development (TRI-AD) with the goal of developing an electric and self-driving car until 2020. Besides Toyota, other car manufacturers, such as Ford, Volvo, and Mercedes-Benz, have also recently presented their plans for self-driving cars. Ford defined 2021 as a deadline for presenting a fully autonomous car ready for commercial operation.

\section{Conclusion}

In this paper, we surveyed the literature on self-driving cars focusing on research that has being tested in the real world. Since the DARPA challenges of 2004, 2005 and 2007, a large body of research have contributed to the current state of the self-driving cars’ technology. However, much still have to be done to achieve the industry and academy goal of making SAE level 5 self-driving cars available to the public at large. 

\section*{Acknowledgment}

This work was supported in part by Conselho Nacional de Desenvolvimento Cient\'ifico e Tecnol\'ogico (CNPq), Brazil, under Grants 311120/2016-4 and 311504/ 2017-5; Coordena{\c{c}}\~ao de Aperfei{\c{c}}oamento de Pessoal de Nível Superior (CAPES), Brazil, under Finance Code 001; Funda{\c{c}}\~ao de Amparo \`a Pesquisa do Esp\'irito Santo (FAPES), Brazil, under Grant 84412844/2018; Vale company, Brazil, with FAPES, Brazil, under Grant 75537958/16; and Embraer company, Brazil, under Grant GDT0017-18.

\section*{References}

\bibliography{bibliography}

\begin{thebibliography}{233}
\expandafter\ifx\csname natexlab\endcsname\relax\def\natexlab#1{#1}\fi
\providecommand{\url}[1]{\texttt{#1}}
\providecommand{\href}[2]{#2}
\providecommand{\path}[1]{#1}
\providecommand{\DOIprefix}{doi:}
\providecommand{\ArXivprefix}{arXiv:}
\providecommand{\URLprefix}{URL: }
\providecommand{\Pubmedprefix}{pmid:}
\providecommand{\doi}[1]{\href{http://dx.doi.org/#1}{\path{#1}}}
\providecommand{\Pubmed}[1]{\href{pmid:#1}{\path{#1}}}
\providecommand{\bibinfo}[2]{#2}
\ifx\xfnm\relax \def\xfnm[#1]{\unskip,\space#1}\fi
\bibitem[{Abraham et~al.(2012)Abraham, Delling, Goldberg \& Werneck}]{item200}
\bibinfo{author}{Abraham, I.}, \bibinfo{author}{Delling, D.},
  \bibinfo{author}{Goldberg, A.~V.}, \& \bibinfo{author}{Werneck, R.~F.}
  (\bibinfo{year}{2012}).
\newblock \bibinfo{title}{Hierarchical hub labelings for shortest paths}.
\newblock In {\it \bibinfo{booktitle}{European Symposium on Algorithms}\/} (pp.
  \bibinfo{pages}{24--35}).
\newblock \bibinfo{organization}{Springer}.
\bibitem[{Aeberhard et~al.(2015)Aeberhard, Rauch, Bahram, Tanzmeister, Thomas,
  Pilat, Homm, Huber \& Kaempchen}]{item109}
\bibinfo{author}{Aeberhard, M.}, \bibinfo{author}{Rauch, S.},
  \bibinfo{author}{Bahram, M.}, \bibinfo{author}{Tanzmeister, G.},
  \bibinfo{author}{Thomas, J.}, \bibinfo{author}{Pilat, Y.},
  \bibinfo{author}{Homm, F.}, \bibinfo{author}{Huber, W.}, \&
  \bibinfo{author}{Kaempchen, N.} (\bibinfo{year}{2015}).
\newblock \bibinfo{title}{Experience, results and lessons learned from
  automated driving on germany's highways}.
\newblock {\it \bibinfo{journal}{IEEE Intelligent transportation systems
  magazine}\/},  {\it \bibinfo{volume}{7}\/}, \bibinfo{pages}{42--57}.
\bibitem[{Ahmad et~al.(2017)Ahmad, Ilstrup, Emami \& Bebis}]{item184}
\bibinfo{author}{Ahmad, T.}, \bibinfo{author}{Ilstrup, D.},
  \bibinfo{author}{Emami, E.}, \& \bibinfo{author}{Bebis, G.}
  (\bibinfo{year}{2017}).
\newblock \bibinfo{title}{Symbolic road marking recognition using convolutional
  neural networks}.
\newblock In {\it \bibinfo{booktitle}{2017 IEEE Intelligent Vehicles Symposium
  (IV)}\/} (pp. \bibinfo{pages}{1428--1433}).
\newblock \bibinfo{organization}{IEEE}.
\bibitem[{Alia et~al.(2015)Alia, Gilles, Reine \& Ali}]{item242}
\bibinfo{author}{Alia, C.}, \bibinfo{author}{Gilles, T.},
  \bibinfo{author}{Reine, T.}, \& \bibinfo{author}{Ali, C.}
  (\bibinfo{year}{2015}).
\newblock \bibinfo{title}{Local trajectory planning and tracking of autonomous
  vehicles, using clothoid tentacles method}.
\newblock In {\it \bibinfo{booktitle}{2015 IEEE Intelligent Vehicles Symposium
  (IV)}\/} (pp. \bibinfo{pages}{674--679}).
\newblock \bibinfo{organization}{IEEE}.
\bibitem[{Amaral et~al.(2015)Amaral, Badue, Oliveira-Santos \&
  De~Souza}]{item114}
\bibinfo{author}{Amaral, E.}, \bibinfo{author}{Badue, C.},
  \bibinfo{author}{Oliveira-Santos, T.}, \& \bibinfo{author}{De~Souza, A.~F.}
  (\bibinfo{year}{2015}).
\newblock \bibinfo{title}{Detec{\c{c}}\~ao e rastreamento de ve\'iculos em
  movimento para autom\'oveis rob\'oticos aut\^onomos}.
\newblock In {\it \bibinfo{booktitle}{XII Simp\'osio Brasileiro de
  Automa{\c{c}}\~ao Inteligente (SBAI)}\/} (pp. \bibinfo{pages}{801--806}).
\bibitem[{Arnay et~al.(2016)Arnay, Morales, Morell, Hernandez-Aceituno, Perea,
  Toledo, Hamilton, Sanchez-Medina \& Acosta}]{item208}
\bibinfo{author}{Arnay, R.}, \bibinfo{author}{Morales, N.},
  \bibinfo{author}{Morell, A.}, \bibinfo{author}{Hernandez-Aceituno, J.},
  \bibinfo{author}{Perea, D.}, \bibinfo{author}{Toledo, J.~T.},
  \bibinfo{author}{Hamilton, A.}, \bibinfo{author}{Sanchez-Medina, J.~J.}, \&
  \bibinfo{author}{Acosta, L.} (\bibinfo{year}{2016}).
\newblock \bibinfo{title}{Safe and reliable path planning for the autonomous
  vehicle verdino}.
\newblock {\it \bibinfo{journal}{IEEE Intelligent Transportation Systems
  Magazine}\/},  {\it \bibinfo{volume}{8}\/}, \bibinfo{pages}{22--32}.
\bibitem[{Arz et~al.(2013)Arz, Luxen \& Sanders}]{item202}
\bibinfo{author}{Arz, J.}, \bibinfo{author}{Luxen, D.}, \&
  \bibinfo{author}{Sanders, P.} (\bibinfo{year}{2013}).
\newblock \bibinfo{title}{Transit node routing reconsidered}.
\newblock In {\it \bibinfo{booktitle}{International Symposium on Experimental
  Algorithms}\/} (pp. \bibinfo{pages}{55--66}).
\newblock \bibinfo{organization}{Springer}.
\bibitem[{Astr{\"o}m \& Murray(2010)}]{item252}
\bibinfo{author}{Astr{\"o}m, K.~J.}, \& \bibinfo{author}{Murray, R.~M.}
  (\bibinfo{year}{2010}).
\newblock {\it \bibinfo{title}{Feedback systems: an introduction for scientists
  and engineers}\/}.
\newblock \bibinfo{publisher}{Princeton university press}.
\bibitem[{Awad et~al.(2018)Awad, Dsouza, Kim, Schulz, Henrich, Shariff,
  Bonnefon \& Rahwan}]{item219}
\bibinfo{author}{Awad, E.}, \bibinfo{author}{Dsouza, S.}, \bibinfo{author}{Kim,
  R.}, \bibinfo{author}{Schulz, J.}, \bibinfo{author}{Henrich, J.},
  \bibinfo{author}{Shariff, A.}, \bibinfo{author}{Bonnefon, J.-F.}, \&
  \bibinfo{author}{Rahwan, I.} (\bibinfo{year}{2018}).
\newblock \bibinfo{title}{The moral machine experiment}.
\newblock {\it \bibinfo{journal}{Nature}\/},  {\it \bibinfo{volume}{563}\/},
  \bibinfo{pages}{59}.
\bibitem[{Azim \& Aycard(2014)}]{item125}
\bibinfo{author}{Azim, A.}, \& \bibinfo{author}{Aycard, O.}
  (\bibinfo{year}{2014}).
\newblock \bibinfo{title}{Layer-based supervised classification of moving
  objects in outdoor dynamic environment using 3d laser scanner}.
\newblock In {\it \bibinfo{booktitle}{2014 IEEE Intelligent Vehicles
  Symposium}\/} (pp. \bibinfo{pages}{1408--1414}).
\newblock \bibinfo{organization}{IEEE}.
\bibitem[{Bacha et~al.(2008)Bacha, Bauman, Faruque, Fleming, Terwelp,
  Reinholtz, Hong, Wicks, Alberi, Anderson et~al.}]{item11}
\bibinfo{author}{Bacha, A.}, \bibinfo{author}{Bauman, C.},
  \bibinfo{author}{Faruque, R.}, \bibinfo{author}{Fleming, M.},
  \bibinfo{author}{Terwelp, C.}, \bibinfo{author}{Reinholtz, C.},
  \bibinfo{author}{Hong, D.}, \bibinfo{author}{Wicks, A.},
  \bibinfo{author}{Alberi, T.}, \bibinfo{author}{Anderson, D.} et~al.
  (\bibinfo{year}{2008}).
\newblock \bibinfo{title}{Odin: Team victortango's entry in the darpa urban
  challenge}.
\newblock {\it \bibinfo{journal}{Journal of Field Robotics}\/},  {\it
  \bibinfo{volume}{25}\/}, \bibinfo{pages}{467--492}.
\bibitem[{Bailo et~al.(2017)Bailo, Lee, Rameau, Yoon \& Kweon}]{item183}
\bibinfo{author}{Bailo, O.}, \bibinfo{author}{Lee, S.},
  \bibinfo{author}{Rameau, F.}, \bibinfo{author}{Yoon, J.~S.}, \&
  \bibinfo{author}{Kweon, I.~S.} (\bibinfo{year}{2017}).
\newblock \bibinfo{title}{Robust road marking detection and recognition using
  density-based grouping and machine learning techniques}.
\newblock In {\it \bibinfo{booktitle}{2017 IEEE Winter Conference on
  Applications of Computer Vision (WACV)}\/} (pp. \bibinfo{pages}{760--768}).
\newblock \bibinfo{organization}{IEEE}.
\bibitem[{Barnes et~al.(2008)Barnes, Zelinsky \& Fletcher}]{item165}
\bibinfo{author}{Barnes, N.}, \bibinfo{author}{Zelinsky, A.}, \&
  \bibinfo{author}{Fletcher, L.~S.} (\bibinfo{year}{2008}).
\newblock \bibinfo{title}{Real-time speed sign detection using the radial
  symmetry detector}.
\newblock {\it \bibinfo{journal}{IEEE Transactions on Intelligent
  Transportation Systems}\/},  {\it \bibinfo{volume}{9}\/},
  \bibinfo{pages}{322--332}.
\bibitem[{Bast et~al.(2015)Bast, Delling, Goldberg, M{\"u}ller-Hannemann,
  Pajor, Sanders, Wagner \& Werneck}]{item192}
\bibinfo{author}{Bast, H.}, \bibinfo{author}{Delling, D.},
  \bibinfo{author}{Goldberg, A.}, \bibinfo{author}{M{\"u}ller-Hannemann, M.},
  \bibinfo{author}{Pajor, T.}, \bibinfo{author}{Sanders, P.},
  \bibinfo{author}{Wagner, D.}, \& \bibinfo{author}{Werneck, R.~F.}
  (\bibinfo{year}{2015}).
\newblock \bibinfo{title}{Route planning in transportation networks}.
\newblock {\it \bibinfo{journal}{arXiv preprint arXiv:1504.05140}\/}, .
\bibitem[{Bastani et~al.(2018)Bastani, He, Abbar, Alizadeh, Balakrishnan,
  Chawla, Madden \& DeWitt}]{item103}
\bibinfo{author}{Bastani, F.}, \bibinfo{author}{He, S.},
  \bibinfo{author}{Abbar, S.}, \bibinfo{author}{Alizadeh, M.},
  \bibinfo{author}{Balakrishnan, H.}, \bibinfo{author}{Chawla, S.},
  \bibinfo{author}{Madden, S.}, \& \bibinfo{author}{DeWitt, D.}
  (\bibinfo{year}{2018}).
\newblock \bibinfo{title}{Roadtracer: Automatic extraction of road networks
  from aerial images}.
\newblock In {\it \bibinfo{booktitle}{Proceedings of the IEEE Conference on
  Computer Vision and Pattern Recognition}\/} (pp.
  \bibinfo{pages}{4720--4728}).
\bibitem[{Bauer \& Delling(2008)}]{item205}
\bibinfo{author}{Bauer, R.}, \& \bibinfo{author}{Delling, D.}
  (\bibinfo{year}{2008}).
\newblock \bibinfo{title}{Sharc: Fast and robust unidirectional routing}.
\newblock In {\it \bibinfo{booktitle}{2008 Proceedings of the Tenth Workshop on
  Algorithm Engineering and Experiments (ALENEX)}\/} (pp.
  \bibinfo{pages}{13--26}).
\newblock \bibinfo{organization}{SIAM}.
\bibitem[{Bauer et~al.(2010)Bauer, Delling, Sanders, Schieferdecker, Schultes
  \& Wagner}]{item204}
\bibinfo{author}{Bauer, R.}, \bibinfo{author}{Delling, D.},
  \bibinfo{author}{Sanders, P.}, \bibinfo{author}{Schieferdecker, D.},
  \bibinfo{author}{Schultes, D.}, \& \bibinfo{author}{Wagner, D.}
  (\bibinfo{year}{2010}).
\newblock \bibinfo{title}{Combining hierarchical and goal-directed speed-up
  techniques for dijkstra's algorithm}.
\newblock {\it \bibinfo{journal}{ACM Journal of Experimental Algorithmics}\/},
  {\it \bibinfo{volume}{15}\/}.
\bibitem[{Behrendt et~al.(2017)Behrendt, Novak \& Botros}]{item157}
\bibinfo{author}{Behrendt, K.}, \bibinfo{author}{Novak, L.}, \&
  \bibinfo{author}{Botros, R.} (\bibinfo{year}{2017}).
\newblock \bibinfo{title}{A deep learning approach to traffic lights:
  Detection, tracking, and classification}.
\newblock In {\it \bibinfo{booktitle}{2017 IEEE International Conference on
  Robotics and Automation (ICRA)}\/} (pp. \bibinfo{pages}{1370--1377}).
\newblock \bibinfo{organization}{IEEE}.
\bibitem[{Bender et~al.(2014)Bender, Ziegler \& Stiller}]{item102}
\bibinfo{author}{Bender, P.}, \bibinfo{author}{Ziegler, J.}, \&
  \bibinfo{author}{Stiller, C.} (\bibinfo{year}{2014}).
\newblock \bibinfo{title}{Lanelets: Efficient map representation for autonomous
  driving}.
\newblock In {\it \bibinfo{booktitle}{2014 IEEE Intelligent Vehicles Symposium
  Proceedings}\/} (pp. \bibinfo{pages}{420--425}).
\newblock \bibinfo{organization}{IEEE}.
\bibitem[{Berger et~al.(2013)Berger, Forechi, De~Souza, Neto, Veronese, Neves,
  de~Aguiar \& Badue}]{item265}
\bibinfo{author}{Berger, M.}, \bibinfo{author}{Forechi, A.},
  \bibinfo{author}{De~Souza, A.~F.}, \bibinfo{author}{Neto, J. D.~O.},
  \bibinfo{author}{Veronese, L.}, \bibinfo{author}{Neves, V.},
  \bibinfo{author}{de~Aguiar, E.}, \& \bibinfo{author}{Badue, C.}
  (\bibinfo{year}{2013}).
\newblock \bibinfo{title}{Traffic sign recognition with wisard and vg-ram
  weightless neural networks}.
\newblock {\it \bibinfo{journal}{Journal of Network and Innovative
  Computing}\/},  {\it \bibinfo{volume}{1}\/}, \bibinfo{pages}{87--98}.
\bibitem[{Bernini et~al.(2014)Bernini, Bertozzi, Castangia, Patander \&
  Sabbatelli}]{item112}
\bibinfo{author}{Bernini, N.}, \bibinfo{author}{Bertozzi, M.},
  \bibinfo{author}{Castangia, L.}, \bibinfo{author}{Patander, M.}, \&
  \bibinfo{author}{Sabbatelli, M.} (\bibinfo{year}{2014}).
\newblock \bibinfo{title}{Real-time obstacle detection using stereo vision for
  autonomous ground vehicles: A survey}.
\newblock In {\it \bibinfo{booktitle}{17th International IEEE Conference on
  Intelligent Transportation Systems (ITSC)}\/} (pp.
  \bibinfo{pages}{873--878}).
\newblock \bibinfo{organization}{IEEE}.
\bibitem[{Berriel et~al.(2017{\natexlab{a}})Berriel, de~Aguiar, De~Souza \&
  Oliveira-Santos}]{item179}
\bibinfo{author}{Berriel, R.~F.}, \bibinfo{author}{de~Aguiar, E.},
  \bibinfo{author}{De~Souza, A.~F.}, \& \bibinfo{author}{Oliveira-Santos, T.}
  (\bibinfo{year}{2017}{\natexlab{a}}).
\newblock \bibinfo{title}{Ego-lane analysis system (elas): Dataset and
  algorithms}.
\newblock {\it \bibinfo{journal}{Image and Vision Computing}\/},  {\it
  \bibinfo{volume}{68}\/}, \bibinfo{pages}{64--75}.
\bibitem[{Berriel et~al.(2015)Berriel, de~Aguiar, de~Souza~Filho \&
  Oliveira-Santos}]{item178}
\bibinfo{author}{Berriel, R.~F.}, \bibinfo{author}{de~Aguiar, E.},
  \bibinfo{author}{de~Souza~Filho, V.~V.}, \& \bibinfo{author}{Oliveira-Santos,
  T.} (\bibinfo{year}{2015}).
\newblock \bibinfo{title}{A particle filter-based lane marker tracking approach
  using a cubic spline model}.
\newblock In {\it \bibinfo{booktitle}{2015 28th SIBGRAPI Conference on
  Graphics, Patterns and Images}\/} (pp. \bibinfo{pages}{149--156}).
\newblock \bibinfo{organization}{IEEE}.
\bibitem[{Berriel et~al.(2017{\natexlab{b}})Berriel, Rossi, de~Souza \&
  Oliveira-Santos}]{item189}
\bibinfo{author}{Berriel, R.~F.}, \bibinfo{author}{Rossi, F.~S.},
  \bibinfo{author}{de~Souza, A.~F.}, \& \bibinfo{author}{Oliveira-Santos, T.}
  (\bibinfo{year}{2017}{\natexlab{b}}).
\newblock \bibinfo{title}{Automatic large-scale data acquisition via
  crowdsourcing for crosswalk classification: A deep learning approach}.
\newblock {\it \bibinfo{journal}{Computers \& Graphics}\/},  {\it
  \bibinfo{volume}{68}\/}, \bibinfo{pages}{32--42}.
\bibitem[{Brechtel et~al.(2014)Brechtel, Gindele \& Dillmann}]{item226}
\bibinfo{author}{Brechtel, S.}, \bibinfo{author}{Gindele, T.}, \&
  \bibinfo{author}{Dillmann, R.} (\bibinfo{year}{2014}).
\newblock \bibinfo{title}{Probabilistic decision-making under uncertainty for
  autonomous driving using continuous pomdps}.
\newblock In {\it \bibinfo{booktitle}{17th International IEEE Conference on
  Intelligent Transportation Systems (ITSC)}\/} (pp.
  \bibinfo{pages}{392--399}).
\newblock \bibinfo{organization}{IEEE}.
\bibitem[{Broggi et~al.(1999)Broggi, Bertozzi \& Fascioli}]{item4}
\bibinfo{author}{Broggi, A.}, \bibinfo{author}{Bertozzi, M.}, \&
  \bibinfo{author}{Fascioli, A.} (\bibinfo{year}{1999}).
\newblock \bibinfo{title}{Argo and the millemiglia in automatico tour}.
\newblock {\it \bibinfo{journal}{IEEE Intelligent Systems and their
  Applications}\/},  {\it \bibinfo{volume}{14}\/}, \bibinfo{pages}{55--64}.
\bibitem[{Broggi et~al.(2015)Broggi, Cerri, Debattisti, Laghi, Medici,
  Molinari, Panciroli \& Prioletti}]{item287}
\bibinfo{author}{Broggi, A.}, \bibinfo{author}{Cerri, P.},
  \bibinfo{author}{Debattisti, S.}, \bibinfo{author}{Laghi, M.~C.},
  \bibinfo{author}{Medici, P.}, \bibinfo{author}{Molinari, D.},
  \bibinfo{author}{Panciroli, M.}, \& \bibinfo{author}{Prioletti, A.}
  (\bibinfo{year}{2015}).
\newblock \bibinfo{title}{Proud—public road urban driverless-car test}.
\newblock {\it \bibinfo{journal}{IEEE Transactions on Intelligent
  Transportation Systems}\/},  {\it \bibinfo{volume}{16}\/},
  \bibinfo{pages}{3508--3519}.
\bibitem[{Broggi et~al.(2012)Broggi, Cerri, Felisa, Laghi, Mazzei \&
  Porta}]{item15b}
\bibinfo{author}{Broggi, A.}, \bibinfo{author}{Cerri, P.},
  \bibinfo{author}{Felisa, M.}, \bibinfo{author}{Laghi, M.~C.},
  \bibinfo{author}{Mazzei, L.}, \& \bibinfo{author}{Porta, P.~P.}
  (\bibinfo{year}{2012}).
\newblock \bibinfo{title}{The vislab intercontinental autonomous challenge: an
  extensive test for a platoon of intelligent vehicles}.
\newblock {\it \bibinfo{journal}{International Journal of Vehicle Autonomous
  Systems}\/},  {\it \bibinfo{volume}{10}\/}, \bibinfo{pages}{147--164}.
\bibitem[{Brown et~al.(2017)Brown, Funke, Erlien \& Gerdes}]{item257}
\bibinfo{author}{Brown, M.}, \bibinfo{author}{Funke, J.},
  \bibinfo{author}{Erlien, S.}, \& \bibinfo{author}{Gerdes, J.~C.}
  (\bibinfo{year}{2017}).
\newblock \bibinfo{title}{Safe driving envelopes for path tracking in
  autonomous vehicles}.
\newblock {\it \bibinfo{journal}{Control Engineering Practice}\/},  {\it
  \bibinfo{volume}{61}\/}, \bibinfo{pages}{307--316}.
\bibitem[{Brubaker et~al.(2015)Brubaker, Geiger \& Urtasun}]{item70}
\bibinfo{author}{Brubaker, M.~A.}, \bibinfo{author}{Geiger, A.}, \&
  \bibinfo{author}{Urtasun, R.} (\bibinfo{year}{2015}).
\newblock \bibinfo{title}{Map-based probabilistic visual self-localization}.
\newblock {\it \bibinfo{journal}{IEEE transactions on pattern analysis and
  machine intelligence}\/},  {\it \bibinfo{volume}{38}\/},
  \bibinfo{pages}{652--665}.
\bibitem[{Buehler et~al.(2007)Buehler, Iagnemma \& Singh}]{item6}
\bibinfo{author}{Buehler, M.}, \bibinfo{author}{Iagnemma, K.}, \&
  \bibinfo{author}{Singh, S.} (\bibinfo{year}{2007}).
\newblock {\it \bibinfo{title}{The 2005 DARPA grand challenge: the great robot
  race}\/} volume~\bibinfo{volume}{36}.
\newblock \bibinfo{publisher}{Springer}.
\bibitem[{Buehler et~al.(2009)Buehler, Iagnemma \& Singh}]{item8}
\bibinfo{author}{Buehler, M.}, \bibinfo{author}{Iagnemma, K.}, \&
  \bibinfo{author}{Singh, S.} (\bibinfo{year}{2009}).
\newblock {\it \bibinfo{title}{The DARPA urban challenge: autonomous vehicles
  in city traffic}\/} volume~\bibinfo{volume}{56}.
\newblock \bibinfo{publisher}{springer}.
\bibitem[{Cao et~al.(2016)Cao, Song, Huang \& Pan}]{item249}
\bibinfo{author}{Cao, H.}, \bibinfo{author}{Song, X.}, \bibinfo{author}{Huang,
  Z.}, \& \bibinfo{author}{Pan, L.} (\bibinfo{year}{2016}).
\newblock \bibinfo{title}{Simulation research on emergency path planning of an
  active collision avoidance system combined with longitudinal control for an
  autonomous vehicle}.
\newblock {\it \bibinfo{journal}{Proceedings of the Institution of Mechanical
  Engineers, Part D: Journal of automobile engineering}\/},  {\it
  \bibinfo{volume}{230}\/}, \bibinfo{pages}{1624--1653}.
\bibitem[{Cardoso et~al.(2017)Cardoso, Oliveira, Teixeira, Badue, Mutz,
  Oliveira-Santos, Veronese \& De~Souza}]{item55}
\bibinfo{author}{Cardoso, V.}, \bibinfo{author}{Oliveira, J.},
  \bibinfo{author}{Teixeira, T.}, \bibinfo{author}{Badue, C.},
  \bibinfo{author}{Mutz, F.}, \bibinfo{author}{Oliveira-Santos, T.},
  \bibinfo{author}{Veronese, L.}, \& \bibinfo{author}{De~Souza, A.~F.}
  (\bibinfo{year}{2017}).
\newblock \bibinfo{title}{A model-predictive motion planner for the iara
  autonomous car}.
\newblock In {\it \bibinfo{booktitle}{2017 IEEE International Conference on
  Robotics and Automation (ICRA)}\/} (pp. \bibinfo{pages}{225--230}).
\newblock \bibinfo{organization}{IEEE}.
\bibitem[{Carneiro et~al.(2018)Carneiro, Nascimento, Guidolini, Cardoso,
  Oliveira-Santos, Badue \& De~Souza}]{item98}
\bibinfo{author}{Carneiro, R.~V.}, \bibinfo{author}{Nascimento, R.~C.},
  \bibinfo{author}{Guidolini, R.}, \bibinfo{author}{Cardoso, V.~B.},
  \bibinfo{author}{Oliveira-Santos, T.}, \bibinfo{author}{Badue, C.}, \&
  \bibinfo{author}{De~Souza, A.~F.} (\bibinfo{year}{2018}).
\newblock \bibinfo{title}{Mapping road lanes using laser remission and deep
  neural networks}.
\newblock In {\it \bibinfo{booktitle}{2018 International Joint Conference on
  Neural Networks (IJCNN)}\/} (pp. \bibinfo{pages}{1--8}).
\newblock \bibinfo{organization}{IEEE}.
\bibitem[{Cerri et~al.(2011)Cerri, Soprani, Zani, Choi, Lee, Kim, Yi \&
  Broggi}]{item15a}
\bibinfo{author}{Cerri, P.}, \bibinfo{author}{Soprani, G.},
  \bibinfo{author}{Zani, P.}, \bibinfo{author}{Choi, J.}, \bibinfo{author}{Lee,
  J.}, \bibinfo{author}{Kim, D.}, \bibinfo{author}{Yi, K.}, \&
  \bibinfo{author}{Broggi, A.} (\bibinfo{year}{2011}).
\newblock \bibinfo{title}{Computer vision at the hyundai autonomous challenge}.
\newblock In {\it \bibinfo{booktitle}{2011 14th International IEEE Conference
  on Intelligent Transportation Systems (ITSC)}\/} (pp.
  \bibinfo{pages}{777--783}).
\newblock \bibinfo{organization}{IEEE}.
\bibitem[{Chen \& Shen(2017)}]{item93}
\bibinfo{author}{Chen, J.}, \& \bibinfo{author}{Shen, S.}
  (\bibinfo{year}{2017}).
\newblock \bibinfo{title}{Improving octree-based occupancy maps using
  environment sparsity with application to aerial robot navigation}.
\newblock In {\it \bibinfo{booktitle}{2017 IEEE International Conference on
  Robotics and Automation (ICRA)}\/} (pp. \bibinfo{pages}{3656--3663}).
\newblock \bibinfo{organization}{IEEE}.
\bibitem[{Chen et~al.(2017)Chen, Fan, Xie, Huang \& N{\"u}chter}]{item123}
\bibinfo{author}{Chen, L.}, \bibinfo{author}{Fan, L.}, \bibinfo{author}{Xie,
  G.}, \bibinfo{author}{Huang, K.}, \& \bibinfo{author}{N{\"u}chter, A.}
  (\bibinfo{year}{2017}).
\newblock \bibinfo{title}{Moving-object detection from consecutive stereo pairs
  using slanted plane smoothing}.
\newblock {\it \bibinfo{journal}{IEEE Transactions on Intelligent
  Transportation Systems}\/},  {\it \bibinfo{volume}{18}\/},
  \bibinfo{pages}{3093--3102}.
\bibitem[{Cho et~al.(2014)Cho, Seo, Kumar \& Rajkumar}]{item128}
\bibinfo{author}{Cho, H.}, \bibinfo{author}{Seo, Y.-W.},
  \bibinfo{author}{Kumar, B.~V.}, \& \bibinfo{author}{Rajkumar, R.~R.}
  (\bibinfo{year}{2014}).
\newblock \bibinfo{title}{A multi-sensor fusion system for moving object
  detection and tracking in urban driving environments}.
\newblock In {\it \bibinfo{booktitle}{2014 IEEE International Conference on
  Robotics and Automation (ICRA)}\/} (pp. \bibinfo{pages}{1836--1843}).
\newblock \bibinfo{organization}{IEEE}.
\bibitem[{Chu et~al.(2015)Chu, Kim, Jo \& Sunwoo}]{item213}
\bibinfo{author}{Chu, K.}, \bibinfo{author}{Kim, J.}, \bibinfo{author}{Jo, K.},
  \& \bibinfo{author}{Sunwoo, M.} (\bibinfo{year}{2015}).
\newblock \bibinfo{title}{Real-time path planning of autonomous vehicles for
  unstructured road navigation}.
\newblock {\it \bibinfo{journal}{International Journal of Automotive
  Technology}\/},  {\it \bibinfo{volume}{16}\/}, \bibinfo{pages}{653--668}.
\bibitem[{Chu et~al.(2012)Chu, Lee \& Sunwoo}]{item215}
\bibinfo{author}{Chu, K.}, \bibinfo{author}{Lee, M.}, \&
  \bibinfo{author}{Sunwoo, M.} (\bibinfo{year}{2012}).
\newblock \bibinfo{title}{Local path planning for off-road autonomous driving
  with avoidance of static obstacles}.
\newblock {\it \bibinfo{journal}{IEEE Transactions on Intelligent
  Transportation Systems}\/},  {\it \bibinfo{volume}{13}\/},
  \bibinfo{pages}{1599--1616}.
\bibitem[{Cohen et~al.(2003)Cohen, Halperin, Kaplan \& Zwick}]{item199}
\bibinfo{author}{Cohen, E.}, \bibinfo{author}{Halperin, E.},
  \bibinfo{author}{Kaplan, H.}, \& \bibinfo{author}{Zwick, U.}
  (\bibinfo{year}{2003}).
\newblock \bibinfo{title}{Reachability and distance queries via 2-hop labels}.
\newblock {\it \bibinfo{journal}{SIAM Journal on Computing}\/},  {\it
  \bibinfo{volume}{32}\/}, \bibinfo{pages}{1338--1355}.
\bibitem[{Conitzer et~al.(2017)Conitzer, Sinnott-Armstrong, Borg, Deng \&
  Kramer}]{item218}
\bibinfo{author}{Conitzer, V.}, \bibinfo{author}{Sinnott-Armstrong, W.},
  \bibinfo{author}{Borg, J.~S.}, \bibinfo{author}{Deng, Y.}, \&
  \bibinfo{author}{Kramer, M.} (\bibinfo{year}{2017}).
\newblock \bibinfo{title}{Moral decision making frameworks for artificial
  intelligence}.
\newblock In {\it \bibinfo{booktitle}{Thirty-first aaai conference on
  artificial intelligence}\/}.
\bibitem[{Cummins \& Newman(2008)}]{item79}
\bibinfo{author}{Cummins, M.}, \& \bibinfo{author}{Newman, P.}
  (\bibinfo{year}{2008}).
\newblock \bibinfo{title}{Fab-map: Probabilistic localization and mapping in
  the space of appearance}.
\newblock {\it \bibinfo{journal}{The International Journal of Robotics
  Research}\/},  {\it \bibinfo{volume}{27}\/}, \bibinfo{pages}{647--665}.
\bibitem[{Darms et~al.(2009)Darms, Rybski, Baker \& Urmson}]{item127}
\bibinfo{author}{Darms, M.~S.}, \bibinfo{author}{Rybski, P.~E.},
  \bibinfo{author}{Baker, C.}, \& \bibinfo{author}{Urmson, C.}
  (\bibinfo{year}{2009}).
\newblock \bibinfo{title}{Obstacle detection and tracking for the urban
  challenge}.
\newblock {\it \bibinfo{journal}{IEEE Transactions on intelligent
  transportation systems}\/},  {\it \bibinfo{volume}{10}\/},
  \bibinfo{pages}{475--485}.
\bibitem[{DARPA(2005)}]{item99}
\bibinfo{author}{DARPA} (\bibinfo{year}{2005}).
\newblock {\it \bibinfo{title}{DARPA Grand Challenge 2005 Route Data Definition
  File}\/}.
\newblock \bibinfo{type}{Technical Report} Defense Advanced Research Projects
  Agency, Arlington, VA, USA.
\bibitem[{DARPA(2007)}]{item100}
\bibinfo{author}{DARPA} (\bibinfo{year}{2007}).
\newblock {\it \bibinfo{title}{Urban Challenge: Route Network Definition File
  (RNDF) and Mission Data File (MDF) Formats}\/}.
\newblock \bibinfo{type}{Technical Report} Defense Advanced Research Projects
  Agency, Arlington, VA, USA.
\bibitem[{De~Lima \& Pereira(2010)}]{item46}
\bibinfo{author}{De~Lima, D.~A.}, \& \bibinfo{author}{Pereira, G.~A.}
  (\bibinfo{year}{2010}).
\newblock \bibinfo{title}{Um sistema de visao est{\'e}reo para navega{\c{c}}ao
  de um carro aut{\^o}nomo em ambientes com obst{\'a}culos}.
\newblock In {\it \bibinfo{booktitle}{XVIII Congresso Brasileiro de
  Autom{\'a}tica}\/} (pp. \bibinfo{pages}{224--231}).
\bibitem[{De~Lima \& Pereira(2013)}]{item48}
\bibinfo{author}{De~Lima, D.~A.}, \& \bibinfo{author}{Pereira, G. A.~S.}
  (\bibinfo{year}{2013}).
\newblock \bibinfo{title}{Navigation of an autonomous car using vector fields
  and the dynamic window approach}.
\newblock {\it \bibinfo{journal}{Journal of Control, Automation and Electrical
  Systems}\/},  {\it \bibinfo{volume}{24}\/}, \bibinfo{pages}{106--116}.
\bibitem[{De~Souza et~al.(2013{\natexlab{a}})De~Souza, Fontana, Mutz,
  de~Oliveira, Berger, Forechi, de~Oliveira~Neto, de~Aguiar \& Badue}]{item172}
\bibinfo{author}{De~Souza, A.~F.}, \bibinfo{author}{Fontana, C.},
  \bibinfo{author}{Mutz, F.}, \bibinfo{author}{de~Oliveira, T.~A.},
  \bibinfo{author}{Berger, M.}, \bibinfo{author}{Forechi, A.},
  \bibinfo{author}{de~Oliveira~Neto, J.}, \bibinfo{author}{de~Aguiar, E.}, \&
  \bibinfo{author}{Badue, C.} (\bibinfo{year}{2013}{\natexlab{a}}).
\newblock \bibinfo{title}{Traffic sign detection with vg-ram weightless neural
  networks}.
\newblock In {\it \bibinfo{booktitle}{The 2013 International Joint Conference
  on Neural Networks (IJCNN)}\/} (pp. \bibinfo{pages}{1--9}).
\newblock \bibinfo{organization}{IEEE}.
\bibitem[{De~Souza et~al.(2013{\natexlab{b}})De~Souza, Fontana, Mutz,
  de~Oliveira, Berger, Forechi, de~Oliveira~Neto, de~Aguiar \& Badue}]{item263}
\bibinfo{author}{De~Souza, A.~F.}, \bibinfo{author}{Fontana, C.},
  \bibinfo{author}{Mutz, F.}, \bibinfo{author}{de~Oliveira, T.~A.},
  \bibinfo{author}{Berger, M.}, \bibinfo{author}{Forechi, A.},
  \bibinfo{author}{de~Oliveira~Neto, J.}, \bibinfo{author}{de~Aguiar, E.}, \&
  \bibinfo{author}{Badue, C.} (\bibinfo{year}{2013}{\natexlab{b}}).
\newblock \bibinfo{title}{Traffic sign detection with vg-ram weightless neural
  networks}.
\newblock In {\it \bibinfo{booktitle}{The 2013 International Joint Conference
  on Neural Networks (IJCNN)}\/} (pp. \bibinfo{pages}{1--9}).
\newblock \bibinfo{organization}{IEEE}.
\bibitem[{Delling et~al.(2013{\natexlab{a}})Delling, Goldberg, Nowatzyk \&
  Werneck}]{item206}
\bibinfo{author}{Delling, D.}, \bibinfo{author}{Goldberg, A.~V.},
  \bibinfo{author}{Nowatzyk, A.}, \& \bibinfo{author}{Werneck, R.~F.}
  (\bibinfo{year}{2013}{\natexlab{a}}).
\newblock \bibinfo{title}{Phast: Hardware-accelerated shortest path trees}.
\newblock {\it \bibinfo{journal}{Journal of Parallel and Distributed
  Computing}\/},  {\it \bibinfo{volume}{73}\/}, \bibinfo{pages}{940--952}.
\bibitem[{Delling et~al.(2015)Delling, Goldberg, Pajor \& Werneck}]{item196}
\bibinfo{author}{Delling, D.}, \bibinfo{author}{Goldberg, A.~V.},
  \bibinfo{author}{Pajor, T.}, \& \bibinfo{author}{Werneck, R.~F.}
  (\bibinfo{year}{2015}).
\newblock \bibinfo{title}{Customizable route planning in road networks}.
\newblock {\it \bibinfo{journal}{Transportation Science}\/},  {\it
  \bibinfo{volume}{51}\/}, \bibinfo{pages}{566--591}.
\bibitem[{Delling et~al.(2013{\natexlab{b}})Delling, Goldberg \&
  Werneck}]{item201}
\bibinfo{author}{Delling, D.}, \bibinfo{author}{Goldberg, A.~V.}, \&
  \bibinfo{author}{Werneck, R.~F.} (\bibinfo{year}{2013}{\natexlab{b}}).
\newblock \bibinfo{title}{Hub label compression}.
\newblock In {\it \bibinfo{booktitle}{International Symposium on Experimental
  Algorithms}\/} (pp. \bibinfo{pages}{18--29}).
\newblock \bibinfo{organization}{Springer}.
\bibitem[{Delling et~al.(2009)Delling, Holzer, M{\"u}ller, Schulz \&
  Wagner}]{item195}
\bibinfo{author}{Delling, D.}, \bibinfo{author}{Holzer, M.},
  \bibinfo{author}{M{\"u}ller, K.}, \bibinfo{author}{Schulz, F.}, \&
  \bibinfo{author}{Wagner, D.} (\bibinfo{year}{2009}).
\newblock \bibinfo{title}{High-performance multi-level routing}.
\newblock {\it \bibinfo{journal}{The Shortest Path Problem: Ninth DIMACS
  Implementation Challenge}\/},  {\it \bibinfo{volume}{74}\/},
  \bibinfo{pages}{73--92}.
\bibitem[{Dias et~al.(2014)Dias, Pereira \& Palhares}]{item49}
\bibinfo{author}{Dias, J. E.~A.}, \bibinfo{author}{Pereira, G. A.~S.}, \&
  \bibinfo{author}{Palhares, R.~M.} (\bibinfo{year}{2014}).
\newblock \bibinfo{title}{Longitudinal model identification and velocity
  control of an autonomous car}.
\newblock {\it \bibinfo{journal}{IEEE Transactions on Intelligent
  Transportation Systems}\/},  {\it \bibinfo{volume}{16}\/},
  \bibinfo{pages}{776--786}.
\bibitem[{Diaz-Cabrera et~al.(2015)Diaz-Cabrera, Cerri \& Medici}]{item139}
\bibinfo{author}{Diaz-Cabrera, M.}, \bibinfo{author}{Cerri, P.}, \&
  \bibinfo{author}{Medici, P.} (\bibinfo{year}{2015}).
\newblock \bibinfo{title}{Robust real-time traffic light detection and distance
  estimation using a single camera}.
\newblock {\it \bibinfo{journal}{Expert Systems with Applications}\/},  {\it
  \bibinfo{volume}{42}\/}, \bibinfo{pages}{3911--3923}.
\bibitem[{Diaz-Cabrera et~al.(2012)Diaz-Cabrera, Cerri \&
  Sanchez-Medina}]{item138}
\bibinfo{author}{Diaz-Cabrera, M.}, \bibinfo{author}{Cerri, P.}, \&
  \bibinfo{author}{Sanchez-Medina, J.} (\bibinfo{year}{2012}).
\newblock \bibinfo{title}{Suspended traffic lights detection and distance
  estimation using color features}.
\newblock In {\it \bibinfo{booktitle}{2012 15th International IEEE Conference
  on Intelligent Transportation Systems}\/} (pp. \bibinfo{pages}{1315--1320}).
\newblock \bibinfo{organization}{IEEE}.
\bibitem[{Dijkstra(1959)}]{item190}
\bibinfo{author}{Dijkstra, E.~W.} (\bibinfo{year}{1959}).
\newblock \bibinfo{title}{A note on two problems in connexion with graphs}.
\newblock {\it \bibinfo{journal}{Numerische mathematik}\/},  {\it
  \bibinfo{volume}{1}\/}, \bibinfo{pages}{269--271}.
\bibitem[{Doherty et~al.(2016)Doherty, Wang \& Englot}]{item95}
\bibinfo{author}{Doherty, K.}, \bibinfo{author}{Wang, J.}, \&
  \bibinfo{author}{Englot, B.} (\bibinfo{year}{2016}).
\newblock \bibinfo{title}{Probabilistic map fusion for fast, incremental
  occupancy mapping with 3d hilbert maps}.
\newblock In {\it \bibinfo{booktitle}{2016 IEEE International Conference on
  Robotics and Automation (ICRA)}\/} (pp. \bibinfo{pages}{1011--1018}).
\newblock \bibinfo{organization}{IEEE}.
\bibitem[{Dolgov et~al.(2010)Dolgov, Thrun, Montemerlo \& Diebel}]{item212}
\bibinfo{author}{Dolgov, D.}, \bibinfo{author}{Thrun, S.},
  \bibinfo{author}{Montemerlo, M.}, \& \bibinfo{author}{Diebel, J.}
  (\bibinfo{year}{2010}).
\newblock \bibinfo{title}{Path planning for autonomous vehicles in unknown
  semi-structured environments}.
\newblock {\it \bibinfo{journal}{The International Journal of Robotics
  Research}\/},  {\it \bibinfo{volume}{29}\/}, \bibinfo{pages}{485--501}.
\bibitem[{Droeschel et~al.(2017)Droeschel, Schwarz \& Behnke}]{item94}
\bibinfo{author}{Droeschel, D.}, \bibinfo{author}{Schwarz, M.}, \&
  \bibinfo{author}{Behnke, S.} (\bibinfo{year}{2017}).
\newblock \bibinfo{title}{Continuous mapping and localization for autonomous
  navigation in rough terrain using a 3d laser scanner}.
\newblock {\it \bibinfo{journal}{Robotics and Autonomous Systems}\/},  {\it
  \bibinfo{volume}{88}\/}, \bibinfo{pages}{104--115}.
\bibitem[{Du et~al.(2016)Du, Mei, Liang, Chen, Huang \& Zhao}]{item241}
\bibinfo{author}{Du, M.}, \bibinfo{author}{Mei, T.}, \bibinfo{author}{Liang,
  H.}, \bibinfo{author}{Chen, J.}, \bibinfo{author}{Huang, R.}, \&
  \bibinfo{author}{Zhao, P.} (\bibinfo{year}{2016}).
\newblock \bibinfo{title}{Drivers’ visual behavior-guided rrt motion planner
  for autonomous on-road driving}.
\newblock {\it \bibinfo{journal}{Sensors}\/},  {\it \bibinfo{volume}{16}\/},
  \bibinfo{pages}{102}.
\bibitem[{Englund et~al.(2016)Englund, Chen, Ploeg, Semsar-Kazerooni, Voronov,
  Bengtsson \& Didoff}]{item16}
\bibinfo{author}{Englund, C.}, \bibinfo{author}{Chen, L.},
  \bibinfo{author}{Ploeg, J.}, \bibinfo{author}{Semsar-Kazerooni, E.},
  \bibinfo{author}{Voronov, A.}, \bibinfo{author}{Bengtsson, H.~H.}, \&
  \bibinfo{author}{Didoff, J.} (\bibinfo{year}{2016}).
\newblock \bibinfo{title}{The grand cooperative driving challenge 2016:
  boosting the introduction of cooperative automated vehicles}.
\newblock {\it \bibinfo{journal}{IEEE Wireless Communications}\/},  {\it
  \bibinfo{volume}{23}\/}, \bibinfo{pages}{146--152}.
\bibitem[{Ess et~al.(2010)Ess, Schindler, Leibe \& Van~Gool}]{item121}
\bibinfo{author}{Ess, A.}, \bibinfo{author}{Schindler, K.},
  \bibinfo{author}{Leibe, B.}, \& \bibinfo{author}{Van~Gool, L.}
  (\bibinfo{year}{2010}).
\newblock \bibinfo{title}{Object detection and tracking for autonomous
  navigation in dynamic environments}.
\newblock {\it \bibinfo{journal}{The International Journal of Robotics
  Research}\/},  {\it \bibinfo{volume}{29}\/}, \bibinfo{pages}{1707--1725}.
\bibitem[{Fassbender et~al.(2016)Fassbender, Heinrich \& Wuensche}]{item238}
\bibinfo{author}{Fassbender, D.}, \bibinfo{author}{Heinrich, B.~C.}, \&
  \bibinfo{author}{Wuensche, H.-J.} (\bibinfo{year}{2016}).
\newblock \bibinfo{title}{Motion planning for autonomous vehicles in highly
  constrained urban environments}.
\newblock In {\it \bibinfo{booktitle}{2016 IEEE/RSJ International Conference on
  Intelligent Robots and Systems (IROS)}\/} (pp. \bibinfo{pages}{4708--4713}).
\newblock \bibinfo{organization}{IEEE}.
\bibitem[{Ferguson et~al.(2008)Ferguson, Howard \& Likhachev}]{item247}
\bibinfo{author}{Ferguson, D.}, \bibinfo{author}{Howard, T.~M.}, \&
  \bibinfo{author}{Likhachev, M.} (\bibinfo{year}{2008}).
\newblock \bibinfo{title}{Motion planning in urban environments}.
\newblock {\it \bibinfo{journal}{Journal of Field Robotics}\/},  {\it
  \bibinfo{volume}{25}\/}, \bibinfo{pages}{939--960}.
\bibitem[{Fernandes et~al.(2014)Fernandes, Souza, Pessin, Shinzato, Sales,
  Mendes, Prado, Klaser, Magalhaes, Hata et~al.}]{item50}
\bibinfo{author}{Fernandes, L.~C.}, \bibinfo{author}{Souza, J.~R.},
  \bibinfo{author}{Pessin, G.}, \bibinfo{author}{Shinzato, P.~Y.},
  \bibinfo{author}{Sales, D.}, \bibinfo{author}{Mendes, C.},
  \bibinfo{author}{Prado, M.}, \bibinfo{author}{Klaser, R.},
  \bibinfo{author}{Magalhaes, A.~C.}, \bibinfo{author}{Hata, A.} et~al.
  (\bibinfo{year}{2014}).
\newblock \bibinfo{title}{Carina intelligent robotic car: architectural design
  and applications}.
\newblock {\it \bibinfo{journal}{Journal of Systems Architecture}\/},  {\it
  \bibinfo{volume}{60}\/}, \bibinfo{pages}{372--392}.
\bibitem[{Forechi et~al.(2018)Forechi, Oliveira-Santos, Badue \&
  De~Souza}]{item81}
\bibinfo{author}{Forechi, A.}, \bibinfo{author}{Oliveira-Santos, T.},
  \bibinfo{author}{Badue, C.}, \& \bibinfo{author}{De~Souza, A.~F.}
  (\bibinfo{year}{2018}).
\newblock \bibinfo{title}{Visual global localization with a hybrid wnn-cnn
  approach}.
\newblock In {\it \bibinfo{booktitle}{2018 International Joint Conference on
  Neural Networks (IJCNN)}\/} (pp. \bibinfo{pages}{1--9}).
\newblock \bibinfo{organization}{IEEE}.
\bibitem[{Foucher et~al.(2011)Foucher, Sebsadji, Tarel, Charbonnier \&
  Nicolle}]{item188}
\bibinfo{author}{Foucher, P.}, \bibinfo{author}{Sebsadji, Y.},
  \bibinfo{author}{Tarel, J.-P.}, \bibinfo{author}{Charbonnier, P.}, \&
  \bibinfo{author}{Nicolle, P.} (\bibinfo{year}{2011}).
\newblock \bibinfo{title}{Detection and recognition of urban road markings
  using images}.
\newblock In {\it \bibinfo{booktitle}{2011 14th International IEEE Conference
  on Intelligent Transportation Systems (ITSC)}\/} (pp.
  \bibinfo{pages}{1747--1752}).
\newblock \bibinfo{organization}{IEEE}.
\bibitem[{Funke et~al.(2012)Funke, Theodosis, Hindiyeh, Stanek, Kritatakirana,
  Gerdes, Langer, Hernandez, M{\"u}ller-Bessler \& Huhnke}]{item251}
\bibinfo{author}{Funke, J.}, \bibinfo{author}{Theodosis, P.},
  \bibinfo{author}{Hindiyeh, R.}, \bibinfo{author}{Stanek, G.},
  \bibinfo{author}{Kritatakirana, K.}, \bibinfo{author}{Gerdes, C.},
  \bibinfo{author}{Langer, D.}, \bibinfo{author}{Hernandez, M.},
  \bibinfo{author}{M{\"u}ller-Bessler, B.}, \& \bibinfo{author}{Huhnke, B.}
  (\bibinfo{year}{2012}).
\newblock \bibinfo{title}{Up to the limits: Autonomous audi tts}.
\newblock In {\it \bibinfo{booktitle}{2012 IEEE Intelligent Vehicles
  Symposium}\/} (pp. \bibinfo{pages}{541--547}).
\newblock \bibinfo{organization}{IEEE}.
\bibitem[{Galceran et~al.(2017)Galceran, Cunningham, Eustice \&
  Olson}]{item227}
\bibinfo{author}{Galceran, E.}, \bibinfo{author}{Cunningham, A.~G.},
  \bibinfo{author}{Eustice, R.~M.}, \& \bibinfo{author}{Olson, E.}
  (\bibinfo{year}{2017}).
\newblock \bibinfo{title}{Multipolicy decision-making for autonomous driving
  via changepoint-based behavior prediction: Theory and experiment}.
\newblock {\it \bibinfo{journal}{Autonomous Robots}\/},  {\it
  \bibinfo{volume}{41}\/}, \bibinfo{pages}{1367--1382}.
\bibitem[{Gao et~al.(2006)Gao, Podladchikova, Shaposhnikov, Hong \&
  Shevtsova}]{item164}
\bibinfo{author}{Gao, X.~W.}, \bibinfo{author}{Podladchikova, L.},
  \bibinfo{author}{Shaposhnikov, D.}, \bibinfo{author}{Hong, K.}, \&
  \bibinfo{author}{Shevtsova, N.} (\bibinfo{year}{2006}).
\newblock \bibinfo{title}{Recognition of traffic signs based on their colour
  and shape features extracted using human vision models}.
\newblock {\it \bibinfo{journal}{Journal of Visual Communication and Image
  Representation}\/},  {\it \bibinfo{volume}{17}\/}, \bibinfo{pages}{675--685}.
\bibitem[{Garcia-Fidalgo \& Ortiz(2015)}]{item84}
\bibinfo{author}{Garcia-Fidalgo, E.}, \& \bibinfo{author}{Ortiz, A.}
  (\bibinfo{year}{2015}).
\newblock \bibinfo{title}{Vision-based topological mapping and localization
  methods: A survey}.
\newblock {\it \bibinfo{journal}{Robotics and Autonomous Systems}\/},  {\it
  \bibinfo{volume}{64}\/}, \bibinfo{pages}{1--20}.
\bibitem[{Ge et~al.(2017)Ge, Wang, Wang \& Chen}]{item126}
\bibinfo{author}{Ge, Z.}, \bibinfo{author}{Wang, P.}, \bibinfo{author}{Wang,
  J.}, \& \bibinfo{author}{Chen, Z.} (\bibinfo{year}{2017}).
\newblock \bibinfo{title}{A 2.5 d grids based moving detection methodology for
  autonomous vehicle}.
\newblock In {\it \bibinfo{booktitle}{2017 2nd International Conference on
  Robotics and Automation Engineering (ICRAE)}\/} (pp.
  \bibinfo{pages}{403--407}).
\newblock \bibinfo{organization}{IEEE}.
\bibitem[{Geisberger et~al.(2012)Geisberger, Sanders, Schultes \&
  Vetter}]{item197}
\bibinfo{author}{Geisberger, R.}, \bibinfo{author}{Sanders, P.},
  \bibinfo{author}{Schultes, D.}, \& \bibinfo{author}{Vetter, C.}
  (\bibinfo{year}{2012}).
\newblock \bibinfo{title}{Exact routing in large road networks using
  contraction hierarchies}.
\newblock {\it \bibinfo{journal}{Transportation Science}\/},  {\it
  \bibinfo{volume}{46}\/}, \bibinfo{pages}{388--404}.
\bibitem[{Girao et~al.(2016)Girao, Asvadi, Peixoto \& Nunes}]{item113}
\bibinfo{author}{Girao, P.}, \bibinfo{author}{Asvadi, A.},
  \bibinfo{author}{Peixoto, P.}, \& \bibinfo{author}{Nunes, U.}
  (\bibinfo{year}{2016}).
\newblock \bibinfo{title}{3d object tracking in driving environment: a short
  review and a benchmark dataset}.
\newblock In {\it \bibinfo{booktitle}{2016 IEEE 19th International Conference
  on Intelligent Transportation Systems (ITSC)}\/} (pp.
  \bibinfo{pages}{7--12}).
\newblock \bibinfo{organization}{IEEE}.
\bibitem[{Goldberg \& Harrelson(2005)}]{item193}
\bibinfo{author}{Goldberg, A.~V.}, \& \bibinfo{author}{Harrelson, C.}
  (\bibinfo{year}{2005}).
\newblock \bibinfo{title}{Computing the shortest path: A search meets graph
  theory}.
\newblock In {\it \bibinfo{booktitle}{Proceedings of the sixteenth annual
  ACM-SIAM symposium on Discrete algorithms}\/} (pp.
  \bibinfo{pages}{156--165}).
\newblock \bibinfo{organization}{Society for Industrial and Applied
  Mathematics}.
\bibitem[{Goldberg et~al.(2006)Goldberg, Kaplan \& Werneck}]{item203}
\bibinfo{author}{Goldberg, A.~V.}, \bibinfo{author}{Kaplan, H.}, \&
  \bibinfo{author}{Werneck, R.~F.} (\bibinfo{year}{2006}).
\newblock \bibinfo{title}{Reach for a*: Shortest path algorithms with
  preprocessing.}
\newblock In {\it \bibinfo{booktitle}{The Shortest Path Problem}\/} (pp.
  \bibinfo{pages}{93--140}).
\bibitem[{Gomez et~al.(2014)Gomez, Alencar, Prado, Osorio \& Wolf}]{item145}
\bibinfo{author}{Gomez, A.~E.}, \bibinfo{author}{Alencar, F.~A.},
  \bibinfo{author}{Prado, P.~V.}, \bibinfo{author}{Osorio, F.~S.}, \&
  \bibinfo{author}{Wolf, D.~F.} (\bibinfo{year}{2014}).
\newblock \bibinfo{title}{Traffic lights detection and state estimation using
  hidden markov models}.
\newblock In {\it \bibinfo{booktitle}{2014 IEEE Intelligent Vehicles Symposium
  Proceedings}\/} (pp. \bibinfo{pages}{750--755}).
\newblock \bibinfo{organization}{IEEE}.
\bibitem[{Gong et~al.(2010)Gong, Jiang, Xiong, Guan, Tao \& Chen}]{item148}
\bibinfo{author}{Gong, J.}, \bibinfo{author}{Jiang, Y.},
  \bibinfo{author}{Xiong, G.}, \bibinfo{author}{Guan, C.},
  \bibinfo{author}{Tao, G.}, \& \bibinfo{author}{Chen, H.}
  (\bibinfo{year}{2010}).
\newblock \bibinfo{title}{The recognition and tracking of traffic lights based
  on color segmentation and camshift for intelligent vehicles}.
\newblock In {\it \bibinfo{booktitle}{2010 IEEE Intelligent Vehicles
  Symposium}\/} (pp. \bibinfo{pages}{431--435}).
\newblock \bibinfo{organization}{Ieee}.
\bibitem[{Gonz{\'a}lez et~al.(2015)Gonz{\'a}lez, P{\'e}rez, Milan{\'e}s \&
  Nashashibi}]{item207}
\bibinfo{author}{Gonz{\'a}lez, D.}, \bibinfo{author}{P{\'e}rez, J.},
  \bibinfo{author}{Milan{\'e}s, V.}, \& \bibinfo{author}{Nashashibi, F.}
  (\bibinfo{year}{2015}).
\newblock \bibinfo{title}{A review of motion planning techniques for automated
  vehicles}.
\newblock {\it \bibinfo{journal}{IEEE Transactions on Intelligent
  Transportation Systems}\/},  {\it \bibinfo{volume}{17}\/},
  \bibinfo{pages}{1135--1145}.
\bibitem[{Greene et~al.(2016)Greene, Rossi, Tasioulas, Venable \&
  Williams}]{item217}
\bibinfo{author}{Greene, J.}, \bibinfo{author}{Rossi, F.},
  \bibinfo{author}{Tasioulas, J.}, \bibinfo{author}{Venable, K.~B.}, \&
  \bibinfo{author}{Williams, B.} (\bibinfo{year}{2016}).
\newblock \bibinfo{title}{Embedding ethical principles in collective decision
  support systems}.
\newblock In {\it \bibinfo{booktitle}{Thirtieth AAAI Conference on Artificial
  Intelligence}\/}.
\bibitem[{Greenhalgh \& Mirmehdi(2015)}]{item186}
\bibinfo{author}{Greenhalgh, J.}, \& \bibinfo{author}{Mirmehdi, M.}
  (\bibinfo{year}{2015}).
\newblock \bibinfo{title}{Detection and recognition of painted road surface
  markings}.
\newblock In {\it \bibinfo{booktitle}{Proceedings of the International
  Conference on Pattern Recognition Applications and Methods-Volume 1}\/} (pp.
  \bibinfo{pages}{130--138}).
\newblock \bibinfo{organization}{SCITEPRESS-Science and Technology
  Publications, Lda}.
\bibitem[{Gregor et~al.(2002)Gregor, Lutzeler, Pellkofer, Siedersberger \&
  Dickmanns}]{item5}
\bibinfo{author}{Gregor, R.}, \bibinfo{author}{Lutzeler, M.},
  \bibinfo{author}{Pellkofer, M.}, \bibinfo{author}{Siedersberger, K.-H.}, \&
  \bibinfo{author}{Dickmanns, E.~D.} (\bibinfo{year}{2002}).
\newblock \bibinfo{title}{Ems-vision: A perceptual system for autonomous
  vehicles}.
\newblock {\it \bibinfo{journal}{IEEE Transactions on Intelligent
  Transportation Systems}\/},  {\it \bibinfo{volume}{3}\/},
  \bibinfo{pages}{48--59}.
\bibitem[{Gu et~al.(2015)Gu, Atwood, Dong, Dolan \& Lee}]{item237}
\bibinfo{author}{Gu, T.}, \bibinfo{author}{Atwood, J.}, \bibinfo{author}{Dong,
  C.}, \bibinfo{author}{Dolan, J.~M.}, \& \bibinfo{author}{Lee, J.-W.}
  (\bibinfo{year}{2015}).
\newblock \bibinfo{title}{Tunable and stable real-time trajectory planning for
  urban autonomous driving}.
\newblock In {\it \bibinfo{booktitle}{2015 IEEE/RSJ International Conference on
  Intelligent Robots and Systems (IROS)}\/} (pp. \bibinfo{pages}{250--256}).
\newblock \bibinfo{organization}{IEEE}.
\bibitem[{Gu et~al.(2016)Gu, Dolan \& Lee}]{item235}
\bibinfo{author}{Gu, T.}, \bibinfo{author}{Dolan, J.~M.}, \&
  \bibinfo{author}{Lee, J.-W.} (\bibinfo{year}{2016}).
\newblock \bibinfo{title}{Automated tactical maneuver discovery, reasoning and
  trajectory planning for autonomous driving}.
\newblock In {\it \bibinfo{booktitle}{2016 IEEE/RSJ International Conference on
  Intelligent Robots and Systems (IROS)}\/} (pp. \bibinfo{pages}{5474--5480}).
\newblock \bibinfo{organization}{IEEE}.
\bibitem[{Gudigar et~al.(2016)Gudigar, Chokkadi \& Raghavendra}]{item163}
\bibinfo{author}{Gudigar, A.}, \bibinfo{author}{Chokkadi, S.}, \&
  \bibinfo{author}{Raghavendra, U.} (\bibinfo{year}{2016}).
\newblock \bibinfo{title}{A review on automatic detection and recognition of
  traffic sign}.
\newblock {\it \bibinfo{journal}{Multimedia Tools and Applications}\/},  {\it
  \bibinfo{volume}{75}\/}, \bibinfo{pages}{333--364}.
\bibitem[{Guidolini et~al.(2016)Guidolini, Badue, Berger, de~Paula~Veronese \&
  De~Souza}]{item56}
\bibinfo{author}{Guidolini, R.}, \bibinfo{author}{Badue, C.},
  \bibinfo{author}{Berger, M.}, \bibinfo{author}{de~Paula~Veronese, L.}, \&
  \bibinfo{author}{De~Souza, A.~F.} (\bibinfo{year}{2016}).
\newblock \bibinfo{title}{A simple yet effective obstacle avoider for the iara
  autonomous car}.
\newblock In {\it \bibinfo{booktitle}{2016 IEEE 19th International Conference
  on Intelligent Transportation Systems (ITSC)}\/} (pp.
  \bibinfo{pages}{1914--1919}).
\newblock \bibinfo{organization}{IEEE}.
\bibitem[{Guidolini et~al.(2017)Guidolini, De~Souza, Mutz \& Badue}]{item57}
\bibinfo{author}{Guidolini, R.}, \bibinfo{author}{De~Souza, A.~F.},
  \bibinfo{author}{Mutz, F.}, \& \bibinfo{author}{Badue, C.}
  (\bibinfo{year}{2017}).
\newblock \bibinfo{title}{Neural-based model predictive control for tackling
  steering delays of autonomous cars}.
\newblock In {\it \bibinfo{booktitle}{2017 International Joint Conference on
  Neural Networks (IJCNN)}\/} (pp. \bibinfo{pages}{4324--4331}).
\newblock \bibinfo{organization}{IEEE}.
\bibitem[{Guidolini et~al.(2018)Guidolini, Scart, Jesus, Cardoso, Badue \&
  Oliveira-Santos}]{item229}
\bibinfo{author}{Guidolini, R.}, \bibinfo{author}{Scart, L.~G.},
  \bibinfo{author}{Jesus, L.~F.}, \bibinfo{author}{Cardoso, V.~B.},
  \bibinfo{author}{Badue, C.}, \& \bibinfo{author}{Oliveira-Santos, T.}
  (\bibinfo{year}{2018}).
\newblock \bibinfo{title}{Handling pedestrians in crosswalks using deep neural
  networks in the iara autonomous car}.
\newblock In {\it \bibinfo{booktitle}{2018 International Joint Conference on
  Neural Networks (IJCNN)}\/} (pp. \bibinfo{pages}{1--8}).
\newblock \bibinfo{organization}{IEEE}.
\bibitem[{Gurghian et~al.(2016)Gurghian, Koduri, Bailur, Carey \&
  Murali}]{item180}
\bibinfo{author}{Gurghian, A.}, \bibinfo{author}{Koduri, T.},
  \bibinfo{author}{Bailur, S.~V.}, \bibinfo{author}{Carey, K.~J.}, \&
  \bibinfo{author}{Murali, V.~N.} (\bibinfo{year}{2016}).
\newblock \bibinfo{title}{Deeplanes: End-to-end lane position estimation using
  deep neural networksa}.
\newblock In {\it \bibinfo{booktitle}{Proceedings of the IEEE Conference on
  Computer Vision and Pattern Recognition Workshops}\/} (pp.
  \bibinfo{pages}{38--45}).
\bibitem[{Gutman(2004)}]{item198}
\bibinfo{author}{Gutman, R.~J.} (\bibinfo{year}{2004}).
\newblock \bibinfo{title}{Reach-based routing: A new approach to shortest path
  algorithms optimized for road networks.}
\newblock {\it \bibinfo{journal}{ALENEX/ANALC}\/},  {\it
  \bibinfo{volume}{4}\/}, \bibinfo{pages}{100--111}.
\bibitem[{Haltakov et~al.(2015)Haltakov, Mayr, Unger \& Ilic}]{item149}
\bibinfo{author}{Haltakov, V.}, \bibinfo{author}{Mayr, J.},
  \bibinfo{author}{Unger, C.}, \& \bibinfo{author}{Ilic, S.}
  (\bibinfo{year}{2015}).
\newblock \bibinfo{title}{Semantic segmentation based traffic light detection
  at day and at night}.
\newblock In {\it \bibinfo{booktitle}{German Conference on Pattern
  Recognition}\/} (pp. \bibinfo{pages}{446--457}).
\newblock \bibinfo{organization}{Springer}.
\bibitem[{Harel(1987)}]{item231}
\bibinfo{author}{Harel, D.} (\bibinfo{year}{1987}).
\newblock \bibinfo{title}{Statecharts: A visual formalism for complex systems}.
\newblock {\it \bibinfo{journal}{Science of computer programming}\/},  {\it
  \bibinfo{volume}{8}\/}, \bibinfo{pages}{231--274}.
\bibitem[{Hart et~al.(1968)Hart, Nilsson \& Raphael}]{item191}
\bibinfo{author}{Hart, P.~E.}, \bibinfo{author}{Nilsson, N.~J.}, \&
  \bibinfo{author}{Raphael, B.} (\bibinfo{year}{1968}).
\newblock \bibinfo{title}{A formal basis for the heuristic determination of
  minimum cost paths}.
\newblock {\it \bibinfo{journal}{IEEE transactions on Systems Science and
  Cybernetics}\/},  {\it \bibinfo{volume}{4}\/}, \bibinfo{pages}{100--107}.
\bibitem[{Hata et~al.(2017)Hata, Ramos \& Wolf}]{item53}
\bibinfo{author}{Hata, A.~Y.}, \bibinfo{author}{Ramos, F.~T.}, \&
  \bibinfo{author}{Wolf, D.~F.} (\bibinfo{year}{2017}).
\newblock \bibinfo{title}{Monte carlo localization on gaussian process
  occupancy maps for urban environments}.
\newblock {\it \bibinfo{journal}{IEEE Transactions on Intelligent
  Transportation Systems}\/},  {\it \bibinfo{volume}{19}\/},
  \bibinfo{pages}{2893--2902}.
\bibitem[{Hata \& Wolf(2015)}]{item63}
\bibinfo{author}{Hata, A.~Y.}, \& \bibinfo{author}{Wolf, D.~F.}
  (\bibinfo{year}{2015}).
\newblock \bibinfo{title}{Feature detection for vehicle localization in urban
  environments using a multilayer lidar}.
\newblock {\it \bibinfo{journal}{IEEE Transactions on Intelligent
  Transportation Systems}\/},  {\it \bibinfo{volume}{17}\/},
  \bibinfo{pages}{420--429}.
\bibitem[{Hata et~al.(2016)Hata, Wolf \& Ramos}]{item96}
\bibinfo{author}{Hata, A.~Y.}, \bibinfo{author}{Wolf, D.~F.}, \&
  \bibinfo{author}{Ramos, F.~T.} (\bibinfo{year}{2016}).
\newblock \bibinfo{title}{Particle filter localization on continuous occupancy
  maps}.
\newblock In {\it \bibinfo{booktitle}{International symposium on experimental
  robotics}\/} (pp. \bibinfo{pages}{742--751}).
\newblock \bibinfo{organization}{Springer}.
\bibitem[{He et~al.(2016)He, Takeuchi, Ninomiya \& Kato}]{item118}
\bibinfo{author}{He, M.}, \bibinfo{author}{Takeuchi, E.},
  \bibinfo{author}{Ninomiya, Y.}, \& \bibinfo{author}{Kato, S.}
  (\bibinfo{year}{2016}).
\newblock \bibinfo{title}{Precise and efficient model-based vehicle tracking
  method using rao-blackwellized and scaling series particle filters}.
\newblock In {\it \bibinfo{booktitle}{2016 IEEE/RSJ International Conference on
  Intelligent Robots and Systems (IROS)}\/} (pp. \bibinfo{pages}{117--124}).
\newblock \bibinfo{organization}{IEEE}.
\bibitem[{He et~al.(2019)He, Liu, Lv, Ji \& Liu}]{item250}
\bibinfo{author}{He, X.}, \bibinfo{author}{Liu, Y.}, \bibinfo{author}{Lv, C.},
  \bibinfo{author}{Ji, X.}, \& \bibinfo{author}{Liu, Y.}
  (\bibinfo{year}{2019}).
\newblock \bibinfo{title}{Emergency steering control of autonomous vehicle for
  collision avoidance and stabilisation}.
\newblock {\it \bibinfo{journal}{Vehicle System Dynamics}\/},  {\it
  \bibinfo{volume}{57}\/}, \bibinfo{pages}{1163--1187}.
\bibitem[{Held et~al.(2016)Held, Thrun \& Savarese}]{item136}
\bibinfo{author}{Held, D.}, \bibinfo{author}{Thrun, S.}, \&
  \bibinfo{author}{Savarese, S.} (\bibinfo{year}{2016}).
\newblock \bibinfo{title}{Learning to track at 100 fps with deep regression
  networks}.
\newblock In {\it \bibinfo{booktitle}{European Conference on Computer
  Vision}\/} (pp. \bibinfo{pages}{749--765}).
\newblock \bibinfo{organization}{Springer}.
\bibitem[{Hess et~al.(2016)Hess, Kohler, Rapp \& Andor}]{item91}
\bibinfo{author}{Hess, W.}, \bibinfo{author}{Kohler, D.},
  \bibinfo{author}{Rapp, H.}, \& \bibinfo{author}{Andor, D.}
  (\bibinfo{year}{2016}).
\newblock \bibinfo{title}{Real-time loop closure in 2d lidar slam}.
\newblock In {\it \bibinfo{booktitle}{2016 IEEE International Conference on
  Robotics and Automation (ICRA)}\/} (pp. \bibinfo{pages}{1271--1278}).
\newblock \bibinfo{organization}{IEEE}.
\bibitem[{Hilger et~al.(2009)Hilger, K{\"o}hler, M{\"o}hring \&
  Schilling}]{item194}
\bibinfo{author}{Hilger, M.}, \bibinfo{author}{K{\"o}hler, E.},
  \bibinfo{author}{M{\"o}hring, R.~H.}, \& \bibinfo{author}{Schilling, H.}
  (\bibinfo{year}{2009}).
\newblock \bibinfo{title}{Fast point-to-point shortest path computations with
  arc-flags}.
\newblock {\it \bibinfo{journal}{The Shortest Path Problem: Ninth DIMACS
  Implementation Challenge}\/},  {\it \bibinfo{volume}{74}\/},
  \bibinfo{pages}{41--72}.
\bibitem[{Hillel et~al.(2014)Hillel, Lerner, Levi \& Raz}]{item108}
\bibinfo{author}{Hillel, A.~B.}, \bibinfo{author}{Lerner, R.},
  \bibinfo{author}{Levi, D.}, \& \bibinfo{author}{Raz, G.}
  (\bibinfo{year}{2014}).
\newblock \bibinfo{title}{Recent progress in road and lane detection: a
  survey}.
\newblock {\it \bibinfo{journal}{Machine vision and applications}\/},  {\it
  \bibinfo{volume}{25}\/}, \bibinfo{pages}{727--745}.
\bibitem[{Hornung et~al.(2013)Hornung, Wurm, Bennewitz, Stachniss \&
  Burgard}]{item82}
\bibinfo{author}{Hornung, A.}, \bibinfo{author}{Wurm, K.~M.},
  \bibinfo{author}{Bennewitz, M.}, \bibinfo{author}{Stachniss, C.}, \&
  \bibinfo{author}{Burgard, W.} (\bibinfo{year}{2013}).
\newblock \bibinfo{title}{Octomap: An efficient probabilistic 3d mapping
  framework based on octrees}.
\newblock {\it \bibinfo{journal}{Autonomous robots}\/},  {\it
  \bibinfo{volume}{34}\/}, \bibinfo{pages}{189--206}.
\bibitem[{Houben et~al.(2013)Houben, Stallkamp, Salmen, Schlipsing \&
  Igel}]{item170}
\bibinfo{author}{Houben, S.}, \bibinfo{author}{Stallkamp, J.},
  \bibinfo{author}{Salmen, J.}, \bibinfo{author}{Schlipsing, M.}, \&
  \bibinfo{author}{Igel, C.} (\bibinfo{year}{2013}).
\newblock \bibinfo{title}{Detection of traffic signs in real-world images: The
  german traffic sign detection benchmark}.
\newblock In {\it \bibinfo{booktitle}{The 2013 international joint conference
  on neural networks (IJCNN)}\/} (pp. \bibinfo{pages}{1--8}).
\newblock \bibinfo{organization}{IEEE}.
\bibitem[{Howard \& Kelly(2007)}]{item246}
\bibinfo{author}{Howard, T.~M.}, \& \bibinfo{author}{Kelly, A.}
  (\bibinfo{year}{2007}).
\newblock \bibinfo{title}{Optimal rough terrain trajectory generation for
  wheeled mobile robots}.
\newblock {\it \bibinfo{journal}{The International Journal of Robotics
  Research}\/},  {\it \bibinfo{volume}{26}\/}, \bibinfo{pages}{141--166}.
\bibitem[{Hu et~al.(2018)Hu, Chen, Tang, Cao \& He}]{item216}
\bibinfo{author}{Hu, X.}, \bibinfo{author}{Chen, L.}, \bibinfo{author}{Tang,
  B.}, \bibinfo{author}{Cao, D.}, \& \bibinfo{author}{He, H.}
  (\bibinfo{year}{2018}).
\newblock \bibinfo{title}{Dynamic path planning for autonomous driving on
  various roads with avoidance of static and moving obstacles}.
\newblock {\it \bibinfo{journal}{Mechanical Systems and Signal Processing}\/},
  {\it \bibinfo{volume}{100}\/}, \bibinfo{pages}{482--500}.
\bibitem[{Huval et~al.(2015)Huval, Wang, Tandon, Kiske, Song, Pazhayampallil,
  Andriluka, Rajpurkar, Migimatsu, Cheng-Yue et~al.}]{item133}
\bibinfo{author}{Huval, B.}, \bibinfo{author}{Wang, T.},
  \bibinfo{author}{Tandon, S.}, \bibinfo{author}{Kiske, J.},
  \bibinfo{author}{Song, W.}, \bibinfo{author}{Pazhayampallil, J.},
  \bibinfo{author}{Andriluka, M.}, \bibinfo{author}{Rajpurkar, P.},
  \bibinfo{author}{Migimatsu, T.}, \bibinfo{author}{Cheng-Yue, R.} et~al.
  (\bibinfo{year}{2015}).
\newblock \bibinfo{title}{An empirical evaluation of deep learning on highway
  driving}.
\newblock {\it \bibinfo{journal}{arXiv preprint arXiv:1504.01716}\/}, .
\bibitem[{Hwang et~al.(2016)Hwang, Kim, Choi, Lee \& Kweon}]{item116}
\bibinfo{author}{Hwang, S.}, \bibinfo{author}{Kim, N.}, \bibinfo{author}{Choi,
  Y.}, \bibinfo{author}{Lee, S.}, \& \bibinfo{author}{Kweon, I.~S.}
  (\bibinfo{year}{2016}).
\newblock \bibinfo{title}{Fast multiple objects detection and tracking fusing
  color camera and 3d lidar for intelligent vehicles}.
\newblock In {\it \bibinfo{booktitle}{2016 13th International Conference on
  Ubiquitous Robots and Ambient Intelligence (URAI)}\/} (pp.
  \bibinfo{pages}{234--239}).
\newblock \bibinfo{organization}{IEEE}.
\bibitem[{Hyeon et~al.(2016)Hyeon, Lee, Jung, Kim \& Seo}]{item185}
\bibinfo{author}{Hyeon, D.}, \bibinfo{author}{Lee, S.}, \bibinfo{author}{Jung,
  S.}, \bibinfo{author}{Kim, S.-W.}, \& \bibinfo{author}{Seo, S.-W.}
  (\bibinfo{year}{2016}).
\newblock \bibinfo{title}{Robust road marking detection using convex grouping
  method in around-view monitoring system}.
\newblock In {\it \bibinfo{booktitle}{2016 IEEE Intelligent Vehicles Symposium
  (IV)}\/} (pp. \bibinfo{pages}{1004--1009}).
\newblock \bibinfo{organization}{IEEE}.
\bibitem[{Ivanchenko et~al.(2008)Ivanchenko, Coughlan \& Shen}]{item187}
\bibinfo{author}{Ivanchenko, V.}, \bibinfo{author}{Coughlan, J.}, \&
  \bibinfo{author}{Shen, H.} (\bibinfo{year}{2008}).
\newblock \bibinfo{title}{Detecting and locating crosswalks using a camera
  phone}.
\newblock In {\it \bibinfo{booktitle}{2008 IEEE Computer Society Conference on
  Computer Vision and Pattern Recognition Workshops}\/} (pp.
  \bibinfo{pages}{1--8}).
\newblock \bibinfo{organization}{IEEE}.
\bibitem[{Jang et~al.(2014)Jang, Kim, Kim, Lee \& Sunwoo}]{item147}
\bibinfo{author}{Jang, C.}, \bibinfo{author}{Kim, C.}, \bibinfo{author}{Kim,
  D.}, \bibinfo{author}{Lee, M.}, \& \bibinfo{author}{Sunwoo, M.}
  (\bibinfo{year}{2014}).
\newblock \bibinfo{title}{Multiple exposure images based traffic light
  recognition}.
\newblock In {\it \bibinfo{booktitle}{2014 IEEE Intelligent Vehicles Symposium
  Proceedings}\/} (pp. \bibinfo{pages}{1313--1318}).
\newblock \bibinfo{organization}{IEEE}.
\bibitem[{Jensen et~al.(2017)Jensen, Nasrollahi \& Moeslund}]{item154}
\bibinfo{author}{Jensen, M.~B.}, \bibinfo{author}{Nasrollahi, K.}, \&
  \bibinfo{author}{Moeslund, T.~B.} (\bibinfo{year}{2017}).
\newblock \bibinfo{title}{Evaluating state-of-the-art object detector on
  challenging traffic light data}.
\newblock In {\it \bibinfo{booktitle}{2017 IEEE Conference on Computer Vision
  and Pattern Recognition Workshops (CVPRW)}\/} (pp.
  \bibinfo{pages}{882--888}).
\newblock \bibinfo{organization}{IEEE}.
\bibitem[{Jensen et~al.(2016{\natexlab{a}})Jensen, Philipsen, M{\o}gelmose,
  Moeslund \& Trivedi}]{item137}
\bibinfo{author}{Jensen, M.~B.}, \bibinfo{author}{Philipsen, M.~P.},
  \bibinfo{author}{M{\o}gelmose, A.}, \bibinfo{author}{Moeslund, T.~B.}, \&
  \bibinfo{author}{Trivedi, M.~M.} (\bibinfo{year}{2016}{\natexlab{a}}).
\newblock \bibinfo{title}{Vision for looking at traffic lights: Issues, survey,
  and perspectives}.
\newblock {\it \bibinfo{journal}{IEEE Transactions on Intelligent
  Transportation Systems}\/},  {\it \bibinfo{volume}{17}\/},
  \bibinfo{pages}{1800--1815}.
\bibitem[{Jensen et~al.(2016{\natexlab{b}})Jensen, Philipsen, M{\o}gelmose,
  Moeslund \& Trivedi}]{item155}
\bibinfo{author}{Jensen, M.~B.}, \bibinfo{author}{Philipsen, M.~P.},
  \bibinfo{author}{M{\o}gelmose, A.}, \bibinfo{author}{Moeslund, T.~B.}, \&
  \bibinfo{author}{Trivedi, M.~M.} (\bibinfo{year}{2016}{\natexlab{b}}).
\newblock \bibinfo{title}{Vision for looking at traffic lights: Issues, survey,
  and perspectives}.
\newblock {\it \bibinfo{journal}{IEEE Transactions on Intelligent
  Transportation Systems}\/},  {\it \bibinfo{volume}{17}\/},
  \bibinfo{pages}{1800--1815}.
\bibitem[{Jo et~al.(2015)Jo, Jo, Suhr, Jung \& Sunwoo}]{item72}
\bibinfo{author}{Jo, K.}, \bibinfo{author}{Jo, Y.}, \bibinfo{author}{Suhr,
  J.~K.}, \bibinfo{author}{Jung, H.~G.}, \& \bibinfo{author}{Sunwoo, M.}
  (\bibinfo{year}{2015}).
\newblock \bibinfo{title}{Precise localization of an autonomous car based on
  probabilistic noise models of road surface marker features using multiple
  cameras}.
\newblock {\it \bibinfo{journal}{IEEE Transactions on Intelligent
  Transportation Systems}\/},  {\it \bibinfo{volume}{16}\/},
  \bibinfo{pages}{3377--3392}.
\bibitem[{Jung \& Kelber(2005)}]{item177}
\bibinfo{author}{Jung, C.~R.}, \& \bibinfo{author}{Kelber, C.~R.}
  (\bibinfo{year}{2005}).
\newblock \bibinfo{title}{Lane following and lane departure using a
  linear-parabolic model}.
\newblock {\it \bibinfo{journal}{Image and Vision Computing}\/},  {\it
  \bibinfo{volume}{23}\/}, \bibinfo{pages}{1192--1202}.
\bibitem[{Kala \& Warwick(2013)}]{item209}
\bibinfo{author}{Kala, R.}, \& \bibinfo{author}{Warwick, K.}
  (\bibinfo{year}{2013}).
\newblock \bibinfo{title}{Multi-level planning for semi-autonomous vehicles in
  traffic scenarios based on separation maximization}.
\newblock {\it \bibinfo{journal}{Journal of Intelligent \& Robotic Systems}\/},
   {\it \bibinfo{volume}{72}\/}, \bibinfo{pages}{559--590}.
\bibitem[{Kim \& Kim(2013)}]{item86}
\bibinfo{author}{Kim, S.}, \& \bibinfo{author}{Kim, J.} (\bibinfo{year}{2013}).
\newblock \bibinfo{title}{Continuous occupancy maps using overlapping local
  gaussian processes}.
\newblock In {\it \bibinfo{booktitle}{2013 IEEE/RSJ International Conference on
  Intelligent Robots and Systems}\/} (pp. \bibinfo{pages}{4709--4714}).
\newblock \bibinfo{organization}{IEEE}.
\bibitem[{Koga et~al.(2016)Koga, Okuda, Tazaki, Suzuki, Haraguchi \&
  Kang}]{item255}
\bibinfo{author}{Koga, A.}, \bibinfo{author}{Okuda, H.},
  \bibinfo{author}{Tazaki, Y.}, \bibinfo{author}{Suzuki, T.},
  \bibinfo{author}{Haraguchi, K.}, \& \bibinfo{author}{Kang, Z.}
  (\bibinfo{year}{2016}).
\newblock \bibinfo{title}{Realization of different driving characteristics for
  autonomous vehicle by using model predictive control}.
\newblock In {\it \bibinfo{booktitle}{2016 IEEE Intelligent Vehicles Symposium
  (IV)}\/} (pp. \bibinfo{pages}{722--728}).
\newblock \bibinfo{organization}{IEEE}.
\bibitem[{Kohlbrecher et~al.(2011)Kohlbrecher, Von~Stryk, Meyer \&
  Klingauf}]{item92}
\bibinfo{author}{Kohlbrecher, S.}, \bibinfo{author}{Von~Stryk, O.},
  \bibinfo{author}{Meyer, J.}, \& \bibinfo{author}{Klingauf, U.}
  (\bibinfo{year}{2011}).
\newblock \bibinfo{title}{A flexible and scalable slam system with full 3d
  motion estimation}.
\newblock In {\it \bibinfo{booktitle}{2011 IEEE International Symposium on
  Safety, Security, and Rescue Robotics}\/} (pp. \bibinfo{pages}{155--160}).
\newblock \bibinfo{organization}{IEEE}.
\bibitem[{Koukoumidis et~al.(2011)Koukoumidis, Martonosi \& Peh}]{item143}
\bibinfo{author}{Koukoumidis, E.}, \bibinfo{author}{Martonosi, M.}, \&
  \bibinfo{author}{Peh, L.-S.} (\bibinfo{year}{2011}).
\newblock \bibinfo{title}{Leveraging smartphone cameras for collaborative road
  advisories}.
\newblock {\it \bibinfo{journal}{IEEE Transactions on mobile computing}\/},
  {\it \bibinfo{volume}{11}\/}, \bibinfo{pages}{707--723}.
\bibitem[{Kritayakirana \& Gerdes(2012)}]{item256}
\bibinfo{author}{Kritayakirana, K.}, \& \bibinfo{author}{Gerdes, J.~C.}
  (\bibinfo{year}{2012}).
\newblock {\it \bibinfo{title}{Autonomous vehicle control at the limits of
  handling}\/}.
\newblock Ph.D. thesis Stanford University Stanford, CA.
\bibitem[{Lafuente-Arroyo et~al.(2010)Lafuente-Arroyo, Salcedo-Sanz,
  Maldonado-Basc{\'o}n, Portilla-Figueras \& L{\'o}pez-Sastre}]{item166}
\bibinfo{author}{Lafuente-Arroyo, S.}, \bibinfo{author}{Salcedo-Sanz, S.},
  \bibinfo{author}{Maldonado-Basc{\'o}n, S.},
  \bibinfo{author}{Portilla-Figueras, J.~A.}, \&
  \bibinfo{author}{L{\'o}pez-Sastre, R.~J.} (\bibinfo{year}{2010}).
\newblock \bibinfo{title}{A decision support system for the automatic
  management of keep-clear signs based on support vector machines and
  geographic information systems}.
\newblock {\it \bibinfo{journal}{Expert Systems with applications}\/},  {\it
  \bibinfo{volume}{37}\/}, \bibinfo{pages}{767--773}.
\bibitem[{Larsson \& Felsberg(2011)}]{item173}
\bibinfo{author}{Larsson, F.}, \& \bibinfo{author}{Felsberg, M.}
  (\bibinfo{year}{2011}).
\newblock \bibinfo{title}{Using fourier descriptors and spatial models for
  traffic sign recognition}.
\newblock In {\it \bibinfo{booktitle}{Scandinavian conference on image
  analysis}\/} (pp. \bibinfo{pages}{238--249}).
\newblock \bibinfo{organization}{Springer}.
\bibitem[{Laurense et~al.(2017)Laurense, Goh \& Gerdes}]{item258}
\bibinfo{author}{Laurense, V.~A.}, \bibinfo{author}{Goh, J.~Y.}, \&
  \bibinfo{author}{Gerdes, J.~C.} (\bibinfo{year}{2017}).
\newblock \bibinfo{title}{Path-tracking for autonomous vehicles at the limit of
  friction}.
\newblock In {\it \bibinfo{booktitle}{2017 American Control Conference
  (ACC)}\/} (pp. \bibinfo{pages}{5586--5591}).
\newblock \bibinfo{organization}{IEEE}.
\bibitem[{LaValle \& Kuffner~Jr(2001)}]{item239}
\bibinfo{author}{LaValle, S.~M.}, \& \bibinfo{author}{Kuffner~Jr, J.~J.}
  (\bibinfo{year}{2001}).
\newblock \bibinfo{title}{Randomized kinodynamic planning}.
\newblock {\it \bibinfo{journal}{The international journal of robotics
  research}\/},  {\it \bibinfo{volume}{20}\/}, \bibinfo{pages}{378--400}.
\bibitem[{Lee et~al.(2017)Lee, Kim, Yoon, Shin, Bailo, Kim, Lee, Hong, Han \&
  Kweon}]{item182}
\bibinfo{author}{Lee, S.}, \bibinfo{author}{Kim, J.}, \bibinfo{author}{Yoon,
  J.~S.}, \bibinfo{author}{Shin, S.}, \bibinfo{author}{Bailo, O.},
  \bibinfo{author}{Kim, N.}, \bibinfo{author}{Lee, T.-H.},
  \bibinfo{author}{Hong, H.~S.}, \bibinfo{author}{Han, S.-H.}, \&
  \bibinfo{author}{Kweon, I.~S.} (\bibinfo{year}{2017}).
\newblock \bibinfo{title}{Vpgnet: Vanishing point guided network for lane and
  road marking detection and recognition}.
\newblock In {\it \bibinfo{booktitle}{2017 IEEE International Conference on
  Computer Vision (ICCV)}\/} (pp. \bibinfo{pages}{1965--1973}).
\newblock \bibinfo{organization}{IEEE}.
\bibitem[{Lee et~al.(2018)Lee, Jung, Jung \& Shim}]{item110}
\bibinfo{author}{Lee, U.}, \bibinfo{author}{Jung, J.}, \bibinfo{author}{Jung,
  S.}, \& \bibinfo{author}{Shim, D.~H.} (\bibinfo{year}{2018}).
\newblock \bibinfo{title}{Development of a self-driving car that can handle the
  adverse weather}.
\newblock {\it \bibinfo{journal}{International journal of automotive
  technology}\/},  {\it \bibinfo{volume}{19}\/}, \bibinfo{pages}{191--197}.
\bibitem[{Leedy et~al.(2007)Leedy, Putney, Bauman, Cacciola, Webster \&
  Reinholtz}]{item210}
\bibinfo{author}{Leedy, B.~M.}, \bibinfo{author}{Putney, J.~S.},
  \bibinfo{author}{Bauman, C.}, \bibinfo{author}{Cacciola, S.},
  \bibinfo{author}{Webster, J.~M.}, \& \bibinfo{author}{Reinholtz, C.~F.}
  (\bibinfo{year}{2007}).
\newblock \bibinfo{title}{Virginia tech’s twin contenders: A comparative
  study of reactive and deliberative navigation}.
\newblock In {\it \bibinfo{booktitle}{The 2005 DARPA Grand Challenge}\/} (pp.
  \bibinfo{pages}{155--182}).
\newblock \bibinfo{publisher}{Springer}.
\bibitem[{Levinson et~al.(2011)Levinson, Askeland, Becker, Dolson, Held,
  Kammel, Kolter, Langer, Pink, Pratt et~al.}]{item254}
\bibinfo{author}{Levinson, J.}, \bibinfo{author}{Askeland, J.},
  \bibinfo{author}{Becker, J.}, \bibinfo{author}{Dolson, J.},
  \bibinfo{author}{Held, D.}, \bibinfo{author}{Kammel, S.},
  \bibinfo{author}{Kolter, J.~Z.}, \bibinfo{author}{Langer, D.},
  \bibinfo{author}{Pink, O.}, \bibinfo{author}{Pratt, V.} et~al.
  (\bibinfo{year}{2011}).
\newblock \bibinfo{title}{Towards fully autonomous driving: Systems and
  algorithms}.
\newblock In {\it \bibinfo{booktitle}{2011 IEEE Intelligent Vehicles Symposium
  (IV)}\/} (pp. \bibinfo{pages}{163--168}).
\newblock \bibinfo{organization}{IEEE}.
\bibitem[{Levinson \& Thrun(2010)}]{item60}
\bibinfo{author}{Levinson, J.}, \& \bibinfo{author}{Thrun, S.}
  (\bibinfo{year}{2010}).
\newblock \bibinfo{title}{Robust vehicle localization in urban environments
  using probabilistic maps}.
\newblock In {\it \bibinfo{booktitle}{2010 IEEE International Conference on
  Robotics and Automation}\/} (pp. \bibinfo{pages}{4372--4378}).
\newblock \bibinfo{organization}{IEEE}.
\bibitem[{Li et~al.(2015)Li, Sun, Cao, He \& Zhu}]{item236}
\bibinfo{author}{Li, X.}, \bibinfo{author}{Sun, Z.}, \bibinfo{author}{Cao, D.},
  \bibinfo{author}{He, Z.}, \& \bibinfo{author}{Zhu, Q.}
  (\bibinfo{year}{2015}).
\newblock \bibinfo{title}{Real-time trajectory planning for autonomous urban
  driving: Framework, algorithms, and verifications}.
\newblock {\it \bibinfo{journal}{IEEE/ASME Transactions on Mechatronics}\/},
  {\it \bibinfo{volume}{21}\/}, \bibinfo{pages}{740--753}.
\bibitem[{Li et~al.(2017)Li, Sun, Cao, Liu \& He}]{item248}
\bibinfo{author}{Li, X.}, \bibinfo{author}{Sun, Z.}, \bibinfo{author}{Cao, D.},
  \bibinfo{author}{Liu, D.}, \& \bibinfo{author}{He, H.}
  (\bibinfo{year}{2017}).
\newblock \bibinfo{title}{Development of a new integrated local trajectory
  planning and tracking control framework for autonomous ground vehicles}.
\newblock {\it \bibinfo{journal}{Mechanical Systems and Signal Processing}\/},
  {\it \bibinfo{volume}{87}\/}, \bibinfo{pages}{118--137}.
\bibitem[{Lindner et~al.(2004)Lindner, Kressel \& Kaelberer}]{item146}
\bibinfo{author}{Lindner, F.}, \bibinfo{author}{Kressel, U.}, \&
  \bibinfo{author}{Kaelberer, S.} (\bibinfo{year}{2004}).
\newblock \bibinfo{title}{Robust recognition of traffic signals}.
\newblock In {\it \bibinfo{booktitle}{IEEE Intelligent Vehicles Symposium,
  2004}\/} (pp. \bibinfo{pages}{49--53}).
\newblock \bibinfo{organization}{IEEE}.
\bibitem[{Liu et~al.(2016)Liu, Anguelov, Erhan, Szegedy, Reed, Fu \&
  Berg}]{item152}
\bibinfo{author}{Liu, W.}, \bibinfo{author}{Anguelov, D.},
  \bibinfo{author}{Erhan, D.}, \bibinfo{author}{Szegedy, C.},
  \bibinfo{author}{Reed, S.}, \bibinfo{author}{Fu, C.-Y.}, \&
  \bibinfo{author}{Berg, A.~C.} (\bibinfo{year}{2016}).
\newblock \bibinfo{title}{Ssd: Single shot multibox detector}.
\newblock In {\it \bibinfo{booktitle}{European conference on computer
  vision}\/} (pp. \bibinfo{pages}{21--37}).
\newblock \bibinfo{organization}{Springer}.
\bibitem[{Lyrio et~al.(2015)Lyrio, Oliveira-Santos, Badue \& De~Souza}]{item77}
\bibinfo{author}{Lyrio, L.~J.}, \bibinfo{author}{Oliveira-Santos, T.},
  \bibinfo{author}{Badue, C.}, \& \bibinfo{author}{De~Souza, A.~F.}
  (\bibinfo{year}{2015}).
\newblock \bibinfo{title}{Image-based mapping, global localization and position
  tracking using vg-ram weightless neural networks}.
\newblock In {\it \bibinfo{booktitle}{2015 IEEE International Conference on
  Robotics and Automation (ICRA)}\/} (pp. \bibinfo{pages}{3603--3610}).
\newblock \bibinfo{organization}{IEEE}.
\bibitem[{Lyrio et~al.(2014)Lyrio, Oliveira-Santos, Forechi, Veronese, Badue \&
  De~Souza}]{item261}
\bibinfo{author}{Lyrio, L.~J.}, \bibinfo{author}{Oliveira-Santos, T.},
  \bibinfo{author}{Forechi, A.}, \bibinfo{author}{Veronese, L.},
  \bibinfo{author}{Badue, C.}, \& \bibinfo{author}{De~Souza, A.~F.}
  (\bibinfo{year}{2014}).
\newblock \bibinfo{title}{Image-based global localization using vg-ram
  weightless neural networks}.
\newblock In {\it \bibinfo{booktitle}{2014 International Joint Conference on
  Neural Networks (IJCNN)}\/} (pp. \bibinfo{pages}{3363--3370}).
\newblock \bibinfo{organization}{IEEE}.
\bibitem[{Massera~Filho et~al.(2014)Massera~Filho, Wolf, Grassi \&
  Os{\'o}rio}]{item51}
\bibinfo{author}{Massera~Filho, C.}, \bibinfo{author}{Wolf, D.~F.},
  \bibinfo{author}{Grassi, V.}, \& \bibinfo{author}{Os{\'o}rio, F.~S.}
  (\bibinfo{year}{2014}).
\newblock \bibinfo{title}{Longitudinal and lateral control for autonomous
  ground vehicles}.
\newblock In {\it \bibinfo{booktitle}{2014 IEEE Intelligent Vehicles Symposium
  Proceedings}\/} (pp. \bibinfo{pages}{588--593}).
\newblock \bibinfo{organization}{IEEE}.
\bibitem[{Mathias et~al.(2013)Mathias, Timofte, Benenson \& Van~Gool}]{item171}
\bibinfo{author}{Mathias, M.}, \bibinfo{author}{Timofte, R.},
  \bibinfo{author}{Benenson, R.}, \& \bibinfo{author}{Van~Gool, L.}
  (\bibinfo{year}{2013}).
\newblock \bibinfo{title}{Traffic sign recognition—how far are we from the
  solution?}
\newblock In {\it \bibinfo{booktitle}{The 2013 international joint conference
  on Neural networks (IJCNN)}\/} (pp. \bibinfo{pages}{1--8}).
\newblock \bibinfo{organization}{IEEE}.
\bibitem[{McCall \& Trivedi(2006)}]{item176}
\bibinfo{author}{McCall, J.}, \& \bibinfo{author}{Trivedi, M.}
  (\bibinfo{year}{2006}).
\newblock \bibinfo{title}{Video-based lane estimation and tracking for driver
  assistance: survey, system, and evaluation}.
\newblock {\it \bibinfo{journal}{IEEE Transactions on Intelligent
  Transportation Systems}\/},  {\it \bibinfo{volume}{1}\/},
  \bibinfo{pages}{20--37}.
\bibitem[{McNaughton et~al.(2011)McNaughton, Urmson, Dolan \& Lee}]{item233}
\bibinfo{author}{McNaughton, M.}, \bibinfo{author}{Urmson, C.},
  \bibinfo{author}{Dolan, J.~M.}, \& \bibinfo{author}{Lee, J.-W.}
  (\bibinfo{year}{2011}).
\newblock \bibinfo{title}{Motion planning for autonomous driving with a
  conformal spatiotemporal lattice}.
\newblock In {\it \bibinfo{booktitle}{2011 IEEE International Conference on
  Robotics and Automation}\/} (pp. \bibinfo{pages}{4889--4895}).
\newblock \bibinfo{organization}{IEEE}.
\bibitem[{Mertz et~al.(2013)Mertz, Navarro-Serment, MacLachlan, Rybski,
  Steinfeld, Supp{\'e}, Urmson, Vandapel, Hebert, Thorpe et~al.}]{item129}
\bibinfo{author}{Mertz, C.}, \bibinfo{author}{Navarro-Serment, L.~E.},
  \bibinfo{author}{MacLachlan, R.}, \bibinfo{author}{Rybski, P.},
  \bibinfo{author}{Steinfeld, A.}, \bibinfo{author}{Supp{\'e}, A.},
  \bibinfo{author}{Urmson, C.}, \bibinfo{author}{Vandapel, N.},
  \bibinfo{author}{Hebert, M.}, \bibinfo{author}{Thorpe, C.} et~al.
  (\bibinfo{year}{2013}).
\newblock \bibinfo{title}{Moving object detection with laser scanners}.
\newblock {\it \bibinfo{journal}{Journal of Field Robotics}\/},  {\it
  \bibinfo{volume}{30}\/}, \bibinfo{pages}{17--43}.
\bibitem[{Milford \& Wyeth(2012)}]{item80}
\bibinfo{author}{Milford, M.~J.}, \& \bibinfo{author}{Wyeth, G.~F.}
  (\bibinfo{year}{2012}).
\newblock \bibinfo{title}{Seqslam: Visual route-based navigation for sunny
  summer days and stormy winter nights}.
\newblock In {\it \bibinfo{booktitle}{2012 IEEE International Conference on
  Robotics and Automation}\/} (pp. \bibinfo{pages}{1643--1649}).
\newblock \bibinfo{organization}{IEEE}.
\bibitem[{Miller et~al.(2008)Miller, Campbell, Huttenlocher, Kline, Nathan,
  Lupashin, Catlin, Schimpf, Moran, Zych et~al.}]{item220}
\bibinfo{author}{Miller, I.}, \bibinfo{author}{Campbell, M.},
  \bibinfo{author}{Huttenlocher, D.}, \bibinfo{author}{Kline, F.-R.},
  \bibinfo{author}{Nathan, A.}, \bibinfo{author}{Lupashin, S.},
  \bibinfo{author}{Catlin, J.}, \bibinfo{author}{Schimpf, B.},
  \bibinfo{author}{Moran, P.}, \bibinfo{author}{Zych, N.} et~al.
  (\bibinfo{year}{2008}).
\newblock \bibinfo{title}{Team cornell's skynet: Robust perception and planning
  in an urban environment}.
\newblock {\it \bibinfo{journal}{Journal of Field Robotics}\/},  {\it
  \bibinfo{volume}{25}\/}, \bibinfo{pages}{493--527}.
\bibitem[{Mnih \& Hinton(2010)}]{item106}
\bibinfo{author}{Mnih, V.}, \& \bibinfo{author}{Hinton, G.~E.}
  (\bibinfo{year}{2010}).
\newblock \bibinfo{title}{Learning to detect roads in high-resolution aerial
  images}.
\newblock In {\it \bibinfo{booktitle}{European Conference on Computer
  Vision}\/} (pp. \bibinfo{pages}{210--223}).
\newblock \bibinfo{organization}{Springer}.
\bibitem[{Mogelmose et~al.(2012)Mogelmose, Trivedi \& Moeslund}]{item162}
\bibinfo{author}{Mogelmose, A.}, \bibinfo{author}{Trivedi, M.~M.}, \&
  \bibinfo{author}{Moeslund, T.~B.} (\bibinfo{year}{2012}).
\newblock \bibinfo{title}{Vision-based traffic sign detection and analysis for
  intelligent driver assistance systems: Perspectives and survey}.
\newblock {\it \bibinfo{journal}{IEEE Transactions on Intelligent
  Transportation Systems}\/},  {\it \bibinfo{volume}{13}\/},
  \bibinfo{pages}{1484--1497}.
\bibitem[{Montemerlo et~al.(2008)Montemerlo, Becker, Bhat, Dahlkamp, Dolgov,
  Ettinger, Haehnel, Hilden, Hoffmann, Huhnke et~al.}]{item10}
\bibinfo{author}{Montemerlo, M.}, \bibinfo{author}{Becker, J.},
  \bibinfo{author}{Bhat, S.}, \bibinfo{author}{Dahlkamp, H.},
  \bibinfo{author}{Dolgov, D.}, \bibinfo{author}{Ettinger, S.},
  \bibinfo{author}{Haehnel, D.}, \bibinfo{author}{Hilden, T.},
  \bibinfo{author}{Hoffmann, G.}, \bibinfo{author}{Huhnke, B.} et~al.
  (\bibinfo{year}{2008}).
\newblock \bibinfo{title}{Junior: The stanford entry in the urban challenge}.
\newblock {\it \bibinfo{journal}{Journal of Field Robotics}\/},  {\it
  \bibinfo{volume}{25}\/}, \bibinfo{pages}{569--597}.
\bibitem[{Moravec \& Elfes(1985)}]{item85}
\bibinfo{author}{Moravec, H.}, \& \bibinfo{author}{Elfes, A.}
  (\bibinfo{year}{1985}).
\newblock \bibinfo{title}{High resolution maps from wide angle sonar}.
\newblock In {\it \bibinfo{booktitle}{Proceedings. 1985 IEEE International
  Conference on Robotics and Automation}\/} (pp. \bibinfo{pages}{116--121}).
\newblock \bibinfo{organization}{IEEE} volume~\bibinfo{volume}{2}.
\bibitem[{Mouhagir et~al.(2017)Mouhagir, Cherfaoui, Talj, Aioun \&
  Guillemard}]{item244}
\bibinfo{author}{Mouhagir, H.}, \bibinfo{author}{Cherfaoui, V.},
  \bibinfo{author}{Talj, R.}, \bibinfo{author}{Aioun, F.}, \&
  \bibinfo{author}{Guillemard, F.} (\bibinfo{year}{2017}).
\newblock \bibinfo{title}{Trajectory planning for autonomous vehicle in
  uncertain environment using evidential grid}.
\newblock {\it \bibinfo{journal}{IFAC-PapersOnLine}\/},  {\it
  \bibinfo{volume}{50}\/}, \bibinfo{pages}{12545--12550}.
\bibitem[{Mouhagir et~al.(2016)Mouhagir, Talj, Cherfaoui, Aioun \&
  Guillemard}]{item243}
\bibinfo{author}{Mouhagir, H.}, \bibinfo{author}{Talj, R.},
  \bibinfo{author}{Cherfaoui, V.}, \bibinfo{author}{Aioun, F.}, \&
  \bibinfo{author}{Guillemard, F.} (\bibinfo{year}{2016}).
\newblock \bibinfo{title}{Integrating safety distances with trajectory planning
  by modifying the occupancy grid for autonomous vehicle navigation}.
\newblock In {\it \bibinfo{booktitle}{2016 IEEE 19th International Conference
  on Intelligent Transportation Systems (ITSC)}\/} (pp.
  \bibinfo{pages}{1114--1119}).
\newblock \bibinfo{organization}{IEEE}.
\bibitem[{Mutz et~al.(2017)Mutz, Cardoso, Teixeira, Jesus, Gol{\c{c}}alves,
  Guidolini, Oliveira, Badue \& De~Souza}]{item135}
\bibinfo{author}{Mutz, F.}, \bibinfo{author}{Cardoso, V.},
  \bibinfo{author}{Teixeira, T.}, \bibinfo{author}{Jesus, L.~F.},
  \bibinfo{author}{Gol{\c{c}}alves, M.~A.}, \bibinfo{author}{Guidolini, R.},
  \bibinfo{author}{Oliveira, J.}, \bibinfo{author}{Badue, C.}, \&
  \bibinfo{author}{De~Souza, A.~F.} (\bibinfo{year}{2017}).
\newblock \bibinfo{title}{Following the leader using a tracking system based on
  pre-trained deep neural networks}.
\newblock In {\it \bibinfo{booktitle}{2017 International Joint Conference on
  Neural Networks (IJCNN)}\/} (pp. \bibinfo{pages}{4332--4339}).
\newblock \bibinfo{organization}{IEEE}.
\bibitem[{Mutz et~al.(2016)Mutz, Veronese, Oliveira-Santos, De~Aguiar, Cheein
  \& De~Souza}]{item54}
\bibinfo{author}{Mutz, F.}, \bibinfo{author}{Veronese, L.~P.},
  \bibinfo{author}{Oliveira-Santos, T.}, \bibinfo{author}{De~Aguiar, E.},
  \bibinfo{author}{Cheein, F. A.~A.}, \& \bibinfo{author}{De~Souza, A.~F.}
  (\bibinfo{year}{2016}).
\newblock \bibinfo{title}{Large-scale mapping in complex field scenarios using
  an autonomous car}.
\newblock {\it \bibinfo{journal}{Expert Systems with Applications}\/},  {\it
  \bibinfo{volume}{46}\/}, \bibinfo{pages}{439--462}.
\bibitem[{Na et~al.(2015)Na, Byun, Roh \& Seo}]{item130}
\bibinfo{author}{Na, K.}, \bibinfo{author}{Byun, J.}, \bibinfo{author}{Roh,
  M.}, \& \bibinfo{author}{Seo, B.} (\bibinfo{year}{2015}).
\newblock \bibinfo{title}{Roadplot-datmo: Moving object tracking and track
  fusion system using multiple sensors}.
\newblock In {\it \bibinfo{booktitle}{2015 International Conference on
  Connected Vehicles and Expo (ICCVE)}\/} (pp. \bibinfo{pages}{142--143}).
\newblock \bibinfo{organization}{IEEE}.
\bibitem[{Nguyen et~al.(2011)Nguyen, Michaelis, Al-Hamadi, Tornow \&
  Meinecke}]{item124}
\bibinfo{author}{Nguyen, T.-N.}, \bibinfo{author}{Michaelis, B.},
  \bibinfo{author}{Al-Hamadi, A.}, \bibinfo{author}{Tornow, M.}, \&
  \bibinfo{author}{Meinecke, M.-M.} (\bibinfo{year}{2011}).
\newblock \bibinfo{title}{Stereo-camera-based urban environment perception
  using occupancy grid and object tracking}.
\newblock {\it \bibinfo{journal}{IEEE Transactions on Intelligent
  Transportation Systems}\/},  {\it \bibinfo{volume}{13}\/},
  \bibinfo{pages}{154--165}.
\bibitem[{Nothdurft et~al.(2011)Nothdurft, Hecker, Ohl, Saust, Maurer, Reschka
  \& B{\"o}hmer}]{item160}
\bibinfo{author}{Nothdurft, T.}, \bibinfo{author}{Hecker, P.},
  \bibinfo{author}{Ohl, S.}, \bibinfo{author}{Saust, F.},
  \bibinfo{author}{Maurer, M.}, \bibinfo{author}{Reschka, A.}, \&
  \bibinfo{author}{B{\"o}hmer, J.~R.} (\bibinfo{year}{2011}).
\newblock \bibinfo{title}{Stadtpilot: First fully autonomous test drives in
  urban traffic}.
\newblock In {\it \bibinfo{booktitle}{2011 14th International IEEE Conference
  on Intelligent Transportation Systems (ITSC)}\/} (pp.
  \bibinfo{pages}{919--924}).
\newblock \bibinfo{organization}{IEEE}.
\bibitem[{Okumura et~al.(2016)Okumura, James, Kanzawa, Derry, Sakai, Nishi \&
  Prokhorov}]{item221}
\bibinfo{author}{Okumura, B.}, \bibinfo{author}{James, M.~R.},
  \bibinfo{author}{Kanzawa, Y.}, \bibinfo{author}{Derry, M.},
  \bibinfo{author}{Sakai, K.}, \bibinfo{author}{Nishi, T.}, \&
  \bibinfo{author}{Prokhorov, D.} (\bibinfo{year}{2016}).
\newblock \bibinfo{title}{Challenges in perception and decision making for
  intelligent automotive vehicles: A case study}.
\newblock {\it \bibinfo{journal}{IEEE Transactions on Intelligent Vehicles}\/},
   {\it \bibinfo{volume}{1}\/}, \bibinfo{pages}{20--32}.
\bibitem[{Oliveira et~al.(2017)Oliveira, Radwan, Burgard \& Brox}]{item78}
\bibinfo{author}{Oliveira, G.~L.}, \bibinfo{author}{Radwan, N.},
  \bibinfo{author}{Burgard, W.}, \& \bibinfo{author}{Brox, T.}
  (\bibinfo{year}{2017}).
\newblock \bibinfo{title}{Topometric localization with deep learning}.
\newblock {\it \bibinfo{journal}{arXiv preprint arXiv:1706.08775}\/}, .
\bibitem[{Omachi \& Omachi(2010)}]{item140}
\bibinfo{author}{Omachi, M.}, \& \bibinfo{author}{Omachi, S.}
  (\bibinfo{year}{2010}).
\newblock \bibinfo{title}{Detection of traffic light using structural
  information}.
\newblock In {\it \bibinfo{booktitle}{IEEE 10th International Conference on
  Signal Processing Proceedings}\/} (pp. \bibinfo{pages}{809--812}).
\newblock \bibinfo{organization}{IEEE}.
\bibitem[{Otsu(1979)}]{item64}
\bibinfo{author}{Otsu, N.} (\bibinfo{year}{1979}).
\newblock \bibinfo{title}{A threshold selection method from gray-level
  histograms}.
\newblock {\it \bibinfo{journal}{IEEE transactions on systems, man, and
  cybernetics}\/},  {\it \bibinfo{volume}{9}\/}, \bibinfo{pages}{62--66}.
\bibitem[{Overett \& Petersson(2011)}]{item168}
\bibinfo{author}{Overett, G.}, \& \bibinfo{author}{Petersson, L.}
  (\bibinfo{year}{2011}).
\newblock \bibinfo{title}{Large scale sign detection using hog feature
  variants}.
\newblock In {\it \bibinfo{booktitle}{2011 IEEE Intelligent Vehicles Symposium
  (IV)}\/} (pp. \bibinfo{pages}{326--331}).
\newblock \bibinfo{organization}{IEEE}.
\bibitem[{Paden et~al.(2016)Paden, {\v{C}}{\'a}p, Yong, Yershov \&
  Frazzoli}]{item59}
\bibinfo{author}{Paden, B.}, \bibinfo{author}{{\v{C}}{\'a}p, M.},
  \bibinfo{author}{Yong, S.~Z.}, \bibinfo{author}{Yershov, D.}, \&
  \bibinfo{author}{Frazzoli, E.} (\bibinfo{year}{2016}).
\newblock \bibinfo{title}{A survey of motion planning and control techniques
  for self-driving urban vehicles}.
\newblock {\it \bibinfo{journal}{IEEE Transactions on intelligent vehicles}\/},
   {\it \bibinfo{volume}{1}\/}, \bibinfo{pages}{33--55}.
\bibitem[{de~Paula~Veronese et~al.(2012)de~Paula~Veronese, Junior, Mutz,
  de~Oliveira~Neto, Azevedo, Berger, De~Souza \& Badue}]{item260}
\bibinfo{author}{de~Paula~Veronese, L.}, \bibinfo{author}{Junior, L. J.~L.},
  \bibinfo{author}{Mutz, F.~W.}, \bibinfo{author}{de~Oliveira~Neto, J.},
  \bibinfo{author}{Azevedo, V.~B.}, \bibinfo{author}{Berger, M.},
  \bibinfo{author}{De~Souza, A.~F.}, \& \bibinfo{author}{Badue, C.}
  (\bibinfo{year}{2012}).
\newblock \bibinfo{title}{Stereo matching with vg-ram weightless neural
  networks}.
\newblock In {\it \bibinfo{booktitle}{2012 12th International Conference on
  Intelligent Systems Design and Applications (ISDA)}\/} (pp.
  \bibinfo{pages}{309--314}).
\newblock \bibinfo{organization}{IEEE}.
\bibitem[{Petrovskaya et~al.(2012)Petrovskaya, Perrollaz, Oliveira, Spinello,
  Triebel, Makris, Yoder, Laugier, Nunes \& Bessiere}]{item111}
\bibinfo{author}{Petrovskaya, A.}, \bibinfo{author}{Perrollaz, M.},
  \bibinfo{author}{Oliveira, L.}, \bibinfo{author}{Spinello, L.},
  \bibinfo{author}{Triebel, R.}, \bibinfo{author}{Makris, A.},
  \bibinfo{author}{Yoder, J.-D.}, \bibinfo{author}{Laugier, C.},
  \bibinfo{author}{Nunes, U.}, \& \bibinfo{author}{Bessiere, P.}
  (\bibinfo{year}{2012}).
\newblock \bibinfo{title}{Awareness of road scene participants for autonomous
  driving}.
\newblock {\it \bibinfo{journal}{Handbook of Intelligent Vehicles}\/},  (pp.
  \bibinfo{pages}{1383--1432}).
\bibitem[{Petrovskaya \& Thrun(2009)}]{item117}
\bibinfo{author}{Petrovskaya, A.}, \& \bibinfo{author}{Thrun, S.}
  (\bibinfo{year}{2009}).
\newblock \bibinfo{title}{Model based vehicle detection and tracking for
  autonomous urban driving}.
\newblock {\it \bibinfo{journal}{Autonomous Robots}\/},  {\it
  \bibinfo{volume}{26}\/}, \bibinfo{pages}{123--139}.
\bibitem[{Pettersson et~al.(2008)Pettersson, Petersson \& Andersson}]{item167}
\bibinfo{author}{Pettersson, N.}, \bibinfo{author}{Petersson, L.}, \&
  \bibinfo{author}{Andersson, L.} (\bibinfo{year}{2008}).
\newblock \bibinfo{title}{The histogram feature-a resource-efficient weak
  classifier}.
\newblock In {\it \bibinfo{booktitle}{2008 IEEE Intelligent Vehicles
  Symposium}\/} (pp. \bibinfo{pages}{678--683}).
\newblock \bibinfo{organization}{IEEE}.
\bibitem[{Pivtoraiko et~al.(2009)Pivtoraiko, Knepper \& Kelly}]{item232}
\bibinfo{author}{Pivtoraiko, M.}, \bibinfo{author}{Knepper, R.~A.}, \&
  \bibinfo{author}{Kelly, A.} (\bibinfo{year}{2009}).
\newblock \bibinfo{title}{Differentially constrained mobile robot motion
  planning in state lattices}.
\newblock {\it \bibinfo{journal}{Journal of Field Robotics}\/},  {\it
  \bibinfo{volume}{26}\/}, \bibinfo{pages}{308--333}.
\bibitem[{Possatti et~al.(2019)Possatti, Guidolini, Cardoso, Berriel,
  Paix{\~a}o, Badue, De~Souza \& Oliveira-Santos}]{item228}
\bibinfo{author}{Possatti, L.~C.}, \bibinfo{author}{Guidolini, R.},
  \bibinfo{author}{Cardoso, V.~B.}, \bibinfo{author}{Berriel, R.~F.},
  \bibinfo{author}{Paix{\~a}o, T.~M.}, \bibinfo{author}{Badue, C.},
  \bibinfo{author}{De~Souza, A.~F.}, \& \bibinfo{author}{Oliveira-Santos, T.}
  (\bibinfo{year}{2019}).
\newblock \bibinfo{title}{Traffic light recognition using deep learning and
  prior maps for autonomous cars}.
\newblock In {\it \bibinfo{booktitle}{2019 International Joint Conference on
  Neural Networks (IJCNN)}\/}.
\newblock \bibinfo{organization}{IEEE}.
\bibitem[{Radaelli et~al.(2014)Radaelli, Badue, Gon{\c{c}}alves,
  Oliveira-Santos \& De~Souza}]{item240}
\bibinfo{author}{Radaelli, R.~R.}, \bibinfo{author}{Badue, C.},
  \bibinfo{author}{Gon{\c{c}}alves, M.~A.}, \bibinfo{author}{Oliveira-Santos,
  T.}, \& \bibinfo{author}{De~Souza, A.~F.} (\bibinfo{year}{2014}).
\newblock \bibinfo{title}{A motion planner for car-like robots based on
  rapidly-exploring random trees}.
\newblock In {\it \bibinfo{booktitle}{Ibero-American Conference on Artificial
  Intelligence}\/} (pp. \bibinfo{pages}{469--480}).
\newblock \bibinfo{organization}{Springer}.
\bibitem[{Radwan et~al.(2016)Radwan, Tipaldi, Spinello \& Burgard}]{item75}
\bibinfo{author}{Radwan, N.}, \bibinfo{author}{Tipaldi, G.~D.},
  \bibinfo{author}{Spinello, L.}, \& \bibinfo{author}{Burgard, W.}
  (\bibinfo{year}{2016}).
\newblock \bibinfo{title}{Do you see the bakery? leveraging geo-referenced
  texts for global localization in public maps}.
\newblock In {\it \bibinfo{booktitle}{2016 IEEE International Conference on
  Robotics and Automation (ICRA)}\/} (pp. \bibinfo{pages}{4837--4842}).
\newblock \bibinfo{organization}{IEEE}.
\bibitem[{Ramm et~al.(2011)Ramm, Topf \& Chilton}]{item101}
\bibinfo{author}{Ramm, F.}, \bibinfo{author}{Topf, J.}, \&
  \bibinfo{author}{Chilton, S.} (\bibinfo{year}{2011}).
\newblock {\it \bibinfo{title}{OpenStreetMap: using and enhancing the free map
  of the world}\/}.
\newblock \bibinfo{publisher}{UIT Cambridge Cambridge}.
\bibitem[{Ramos \& Ott(2016)}]{item97}
\bibinfo{author}{Ramos, F.}, \& \bibinfo{author}{Ott, L.}
  (\bibinfo{year}{2016}).
\newblock \bibinfo{title}{Hilbert maps: scalable continuous occupancy mapping
  with stochastic gradient descent}.
\newblock {\it \bibinfo{journal}{The International Journal of Robotics
  Research}\/},  {\it \bibinfo{volume}{35}\/}, \bibinfo{pages}{1717--1730}.
\bibitem[{Redmon \& Farhadi(2017)}]{item153}
\bibinfo{author}{Redmon, J.}, \& \bibinfo{author}{Farhadi, A.}
  (\bibinfo{year}{2017}).
\newblock \bibinfo{title}{Yolo9000: Better, faster, stronger}.
\newblock In {\it \bibinfo{booktitle}{2017 IEEE Conference on Computer Vision
  and Pattern Recognition (CVPR)}\/} (pp. \bibinfo{pages}{6517--6525}).
\newblock \bibinfo{organization}{IEEE}.
\bibitem[{Ren et~al.(2015)Ren, He, Girshick \& Sun}]{item151}
\bibinfo{author}{Ren, S.}, \bibinfo{author}{He, K.}, \bibinfo{author}{Girshick,
  R.}, \& \bibinfo{author}{Sun, J.} (\bibinfo{year}{2015}).
\newblock \bibinfo{title}{Faster r-cnn: Towards real-time object detection with
  region proposal networks}.
\newblock In {\it \bibinfo{booktitle}{Advances in neural information processing
  systems}\/} (pp. \bibinfo{pages}{91--99}).
\bibitem[{Rohde et~al.(2016)Rohde, Jatzkowski, Mielenz \& Z{\"o}llner}]{item65}
\bibinfo{author}{Rohde, J.}, \bibinfo{author}{Jatzkowski, I.},
  \bibinfo{author}{Mielenz, H.}, \& \bibinfo{author}{Z{\"o}llner, J.~M.}
  (\bibinfo{year}{2016}).
\newblock \bibinfo{title}{Vehicle pose estimation in cluttered urban
  environments using multilayer adaptive monte carlo localization}.
\newblock In {\it \bibinfo{booktitle}{2016 19th International Conference on
  Information Fusion (FUSION)}\/} (pp. \bibinfo{pages}{1774--1779}).
\newblock \bibinfo{organization}{IEEE}.
\bibitem[{Sabbagh et~al.(2010)Sabbagh, Freitas, Castro, Santos, Baleeiro,
  da~Silva, Iscold, Torres \& Pereira}]{item47}
\bibinfo{author}{Sabbagh, V.~B.}, \bibinfo{author}{Freitas, E.~J.},
  \bibinfo{author}{Castro, G.~M.}, \bibinfo{author}{Santos, M.~M.},
  \bibinfo{author}{Baleeiro, M.~F.}, \bibinfo{author}{da~Silva, T.~M.},
  \bibinfo{author}{Iscold, P.}, \bibinfo{author}{Torres, L.~A.}, \&
  \bibinfo{author}{Pereira, G.~A.} (\bibinfo{year}{2010}).
\newblock \bibinfo{title}{Desenvolvimento de um sistema de controle para um
  carro de passeio aut{\^o}nomo}.
\newblock In {\it \bibinfo{booktitle}{XVIII Congresso Brasileiro de
  Autom{\'a}tica}\/}.
\bibitem[{SAE(2018)}]{item58}
\bibinfo{author}{SAE} (\bibinfo{year}{2018}).
\newblock {\it \bibinfo{title}{Taxonomy and definitions for terms related to
  driving automation systems for on-road motor vehicles}\/}.
\newblock \bibinfo{type}{Technical Report} SAE International: Warrendale, PA,
  USA.
\bibitem[{Samson(1995)}]{item288}
\bibinfo{author}{Samson, C.} (\bibinfo{year}{1995}).
\newblock \bibinfo{title}{Control of chained systems application to path
  following and time-varying point-stabilization of mobile robots}.
\newblock {\it \bibinfo{journal}{IEEE transactions on Automatic Control}\/},
  {\it \bibinfo{volume}{40}\/}, \bibinfo{pages}{64--77}.
\bibitem[{Sarcinelli et~al.(2019)Sarcinelli, Guidolini, Cardoso, Paix{\~a}o,
  Berriel, Azevedo, De~Souza, Badue \& Oliveira-Santos}]{item230}
\bibinfo{author}{Sarcinelli, R.}, \bibinfo{author}{Guidolini, R.},
  \bibinfo{author}{Cardoso, V.~B.}, \bibinfo{author}{Paix{\~a}o, T.~M.},
  \bibinfo{author}{Berriel, R.~F.}, \bibinfo{author}{Azevedo, P.},
  \bibinfo{author}{De~Souza, A.~F.}, \bibinfo{author}{Badue, C.}, \&
  \bibinfo{author}{Oliveira-Santos, T.} (\bibinfo{year}{2019}).
\newblock \bibinfo{title}{Handling pedestrians in self-driving cars using image
  tracking and alternative path generation with fren{\'e}t frames}.
\newblock {\it \bibinfo{journal}{Computers \& Graphics}\/}, .
\bibitem[{Schaefer et~al.(2018)Schaefer, Luft \& Burgard}]{item83}
\bibinfo{author}{Schaefer, A.}, \bibinfo{author}{Luft, L.}, \&
  \bibinfo{author}{Burgard, W.} (\bibinfo{year}{2018}).
\newblock \bibinfo{title}{Dct maps: Compact differentiable lidar maps based on
  the cosine transform}.
\newblock {\it \bibinfo{journal}{IEEE Robotics and Automation Letters}\/},
  {\it \bibinfo{volume}{3}\/}, \bibinfo{pages}{1002--1009}.
\bibitem[{Schneider \& Wildermuth(2011)}]{item12}
\bibinfo{author}{Schneider, F.~E.}, \& \bibinfo{author}{Wildermuth, D.}
  (\bibinfo{year}{2011}).
\newblock \bibinfo{title}{Results of the european land robot trial and their
  usability for benchmarking outdoor robot systems}.
\newblock In {\it \bibinfo{booktitle}{Conference Towards Autonomous Robotic
  Systems}\/} (pp. \bibinfo{pages}{408--409}).
\newblock \bibinfo{organization}{Springer}.
\bibitem[{Segal et~al.(2009)Segal, Haehnel \& Thrun}]{item88}
\bibinfo{author}{Segal, A.}, \bibinfo{author}{Haehnel, D.}, \&
  \bibinfo{author}{Thrun, S.} (\bibinfo{year}{2009}).
\newblock \bibinfo{title}{Generalized-icp}.
\newblock In {\it \bibinfo{booktitle}{Robotics: science and systems}\/} (pp.
  \bibinfo{pages}{168--176}).
\newblock \bibinfo{organization}{Seattle, WA} volume~\bibinfo{volume}{5}.
\bibitem[{Sermanet et~al.(2013)Sermanet, Eigen, Mathieu, Zhang, Fergus \&
  Lecun}]{item134}
\bibinfo{author}{Sermanet, P.}, \bibinfo{author}{Eigen, D.},
  \bibinfo{author}{Mathieu, M.}, \bibinfo{author}{Zhang, X.},
  \bibinfo{author}{Fergus, R.}, \& \bibinfo{author}{Lecun, Y.}
  (\bibinfo{year}{2013}).
\newblock \bibinfo{title}{Overfeat detection using deep learning}.
\newblock In {\it \bibinfo{booktitle}{International Conference on Learning
  Representations (ICLR)}\/}.
\newblock volume~\bibinfo{volume}{16}.
\bibitem[{Shinzato et~al.(2016)Shinzato, dos Santos, Rosero, Ridel, Massera,
  Alencar, Batista, Hata, Os{\'o}rio \& Wolf}]{item52}
\bibinfo{author}{Shinzato, P.~Y.}, \bibinfo{author}{dos Santos, T.~C.},
  \bibinfo{author}{Rosero, L.~A.}, \bibinfo{author}{Ridel, D.~A.},
  \bibinfo{author}{Massera, C.~M.}, \bibinfo{author}{Alencar, F.},
  \bibinfo{author}{Batista, M.~P.}, \bibinfo{author}{Hata, A.~Y.},
  \bibinfo{author}{Os{\'o}rio, F.~S.}, \& \bibinfo{author}{Wolf, D.~F.}
  (\bibinfo{year}{2016}).
\newblock \bibinfo{title}{Carina dataset: An emerging-country urban scenario
  benchmark for road detection systems}.
\newblock In {\it \bibinfo{booktitle}{2016 IEEE 19th International Conference
  on Intelligent Transportation Systems (ITSC)}\/} (pp.
  \bibinfo{pages}{41--46}).
\newblock \bibinfo{organization}{IEEE}.
\bibitem[{Sooksatra \& Kondo(2014)}]{item142}
\bibinfo{author}{Sooksatra, S.}, \& \bibinfo{author}{Kondo, T.}
  (\bibinfo{year}{2014}).
\newblock \bibinfo{title}{Red traffic light detection using fast radial
  symmetry transform}.
\newblock In {\it \bibinfo{booktitle}{2014 11th International Conference on
  Electrical Engineering/Electronics, Computer, Telecommunications and
  Information Technology (ECTI-CON)}\/} (pp. \bibinfo{pages}{1--6}).
\newblock \bibinfo{organization}{IEEE}.
\bibitem[{Spangenberg et~al.(2016)Spangenberg, Goehring \& Rojas}]{item76}
\bibinfo{author}{Spangenberg, R.}, \bibinfo{author}{Goehring, D.}, \&
  \bibinfo{author}{Rojas, R.} (\bibinfo{year}{2016}).
\newblock \bibinfo{title}{Pole-based localization for autonomous vehicles in
  urban scenarios}.
\newblock In {\it \bibinfo{booktitle}{2016 IEEE/RSJ International Conference on
  Intelligent Robots and Systems (IROS)}\/} (pp. \bibinfo{pages}{2161--2166}).
\newblock \bibinfo{organization}{IEEE}.
\bibitem[{SparkFun()}]{item14}
\bibinfo{author}{SparkFun} ().
\newblock \bibinfo{title}{Autonomous vehicle competition}.
\newblock \bibinfo{howpublished}{\url{ https://avc.sparkfun.com/}}.
\newblock \bibinfo{note}{Accessed: 22- May- 2018}.
\bibitem[{Stallkamp et~al.(2012)Stallkamp, Schlipsing, Salmen \&
  Igel}]{item169}
\bibinfo{author}{Stallkamp, J.}, \bibinfo{author}{Schlipsing, M.},
  \bibinfo{author}{Salmen, J.}, \& \bibinfo{author}{Igel, C.}
  (\bibinfo{year}{2012}).
\newblock \bibinfo{title}{Man vs. computer: Benchmarking machine learning
  algorithms for traffic sign recognition}.
\newblock {\it \bibinfo{journal}{Neural networks}\/},  {\it
  \bibinfo{volume}{32}\/}, \bibinfo{pages}{323--332}.
\bibitem[{Stentz et~al.(2003)Stentz, Fox \& Montemerlo}]{item90}
\bibinfo{author}{Stentz, A.}, \bibinfo{author}{Fox, D.}, \&
  \bibinfo{author}{Montemerlo, M.} (\bibinfo{year}{2003}).
\newblock \bibinfo{title}{Fastslam: A factored solution to the simultaneous
  localization and mapping problem with unknown data association}.
\newblock In {\it \bibinfo{booktitle}{In Proceedings of the AAAI National
  Conference on Artificial Intelligence}\/}.
\newblock \bibinfo{organization}{Citeseer}.
\bibitem[{Suhr et~al.(2016)Suhr, Jang, Min \& Jung}]{item73}
\bibinfo{author}{Suhr, J.~K.}, \bibinfo{author}{Jang, J.},
  \bibinfo{author}{Min, D.}, \& \bibinfo{author}{Jung, H.~G.}
  (\bibinfo{year}{2016}).
\newblock \bibinfo{title}{Sensor fusion-based low-cost vehicle localization
  system for complex urban environments}.
\newblock {\it \bibinfo{journal}{IEEE Transactions on Intelligent
  Transportation Systems}\/},  {\it \bibinfo{volume}{18}\/},
  \bibinfo{pages}{1078--1086}.
\bibitem[{Teixeira et~al.(2018)Teixeira, Mutz, Cardoso, Veronese, Badue,
  Oliveira-Santos \& De~Souza}]{item89}
\bibinfo{author}{Teixeira, T.}, \bibinfo{author}{Mutz, F.},
  \bibinfo{author}{Cardoso, V.~B.}, \bibinfo{author}{Veronese, L.},
  \bibinfo{author}{Badue, C.}, \bibinfo{author}{Oliveira-Santos, T.}, \&
  \bibinfo{author}{De~Souza, A.~F.} (\bibinfo{year}{2018}).
\newblock \bibinfo{title}{Map memorization and forgetting in the iara
  autonomous car}.
\newblock {\it \bibinfo{journal}{arXiv preprint arXiv:1810.02355}\/}, .
\bibitem[{Thorpe et~al.(1991)Thorpe, Herbert, Kanade \& Shafter}]{item2}
\bibinfo{author}{Thorpe, C.}, \bibinfo{author}{Herbert, M.},
  \bibinfo{author}{Kanade, T.}, \& \bibinfo{author}{Shafter, S.}
  (\bibinfo{year}{1991}).
\newblock \bibinfo{title}{Toward autonomous driving: the cmu navlab. ii.
  architecture and systems}.
\newblock {\it \bibinfo{journal}{IEEE expert}\/},  {\it \bibinfo{volume}{6}\/},
  \bibinfo{pages}{44--52}.
\bibitem[{Thrun(2010)}]{item175}
\bibinfo{author}{Thrun, S.} (\bibinfo{year}{2010}).
\newblock \bibinfo{title}{Toward robotic cars}.
\newblock {\it \bibinfo{journal}{Communications of the ACM}\/},  {\it
  \bibinfo{volume}{53}\/}, \bibinfo{pages}{99--106}.
\bibitem[{Thrun et~al.(2005)Thrun, Burgard \& Fox}]{item61}
\bibinfo{author}{Thrun, S.}, \bibinfo{author}{Burgard, W.}, \&
  \bibinfo{author}{Fox, D.} (\bibinfo{year}{2005}).
\newblock {\it \bibinfo{title}{Probabilistic robotics}\/}.
\newblock \bibinfo{publisher}{MIT press}.
\bibitem[{Thrun \& Montemerlo(2006)}]{item87}
\bibinfo{author}{Thrun, S.}, \& \bibinfo{author}{Montemerlo, M.}
  (\bibinfo{year}{2006}).
\newblock \bibinfo{title}{The graph slam algorithm with applications to
  large-scale mapping of urban structures}.
\newblock {\it \bibinfo{journal}{The International Journal of Robotics
  Research}\/},  {\it \bibinfo{volume}{25}\/}, \bibinfo{pages}{403--429}.
\bibitem[{Thrun et~al.(2006)Thrun, Montemerlo, Dahlkamp, Stavens, Aron, Diebel,
  Fong, Gale, Halpenny, Hoffmann et~al.}]{item7}
\bibinfo{author}{Thrun, S.}, \bibinfo{author}{Montemerlo, M.},
  \bibinfo{author}{Dahlkamp, H.}, \bibinfo{author}{Stavens, D.},
  \bibinfo{author}{Aron, A.}, \bibinfo{author}{Diebel, J.},
  \bibinfo{author}{Fong, P.}, \bibinfo{author}{Gale, J.},
  \bibinfo{author}{Halpenny, M.}, \bibinfo{author}{Hoffmann, G.} et~al.
  (\bibinfo{year}{2006}).
\newblock \bibinfo{title}{Stanley: The robot that won the darpa grand
  challenge}.
\newblock {\it \bibinfo{journal}{Journal of Field Robotics}\/},  {\it
  \bibinfo{volume}{23}\/}, \bibinfo{pages}{661--692}.
\bibitem[{Torres et~al.(2019)Torres, Paix{\~a}o, Berriel, De~Souza, Badue, Sebe
  \& Oliveira-Santos}]{item264}
\bibinfo{author}{Torres, L.~T.}, \bibinfo{author}{Paix{\~a}o, T.~M.},
  \bibinfo{author}{Berriel, R.~F.}, \bibinfo{author}{De~Souza, A.~F.},
  \bibinfo{author}{Badue, C.}, \bibinfo{author}{Sebe, N.}, \&
  \bibinfo{author}{Oliveira-Santos, T.} (\bibinfo{year}{2019}).
\newblock \bibinfo{title}{Effortless deep training for traffic sign detection
  using templates and arbitrary natural images}.
\newblock In {\it \bibinfo{booktitle}{2019 International Joint Conference on
  Neural Networks (IJCNN)}\/}.
\newblock \bibinfo{organization}{IEEE}.
\bibitem[{Trehard et~al.(2014)Trehard, Pollard, Bradai \& Nashashibi}]{item141}
\bibinfo{author}{Trehard, G.}, \bibinfo{author}{Pollard, E.},
  \bibinfo{author}{Bradai, B.}, \& \bibinfo{author}{Nashashibi, F.}
  (\bibinfo{year}{2014}).
\newblock \bibinfo{title}{Tracking both pose and status of a traffic light via
  an interacting multiple model filter}.
\newblock In {\it \bibinfo{booktitle}{17th International Conference on
  Information Fusion (FUSION)}\/} (pp. \bibinfo{pages}{1--7}).
\newblock \bibinfo{organization}{IEEE}.
\bibitem[{Uber()}]{item24}
\bibinfo{author}{Uber} ().
\newblock \bibinfo{title}{Uber: Our road to self-driving vehicles}.
\newblock \bibinfo{howpublished}{\url{
  https://www.uber.com/blog/our-road-to-self-driving-vehicles }}.
\newblock \bibinfo{note}{Accessed: 22- May- 2018}.
\bibitem[{Ulbrich \& Maurer(2013)}]{item225}
\bibinfo{author}{Ulbrich, S.}, \& \bibinfo{author}{Maurer, M.}
  (\bibinfo{year}{2013}).
\newblock \bibinfo{title}{Probabilistic online pomdp decision making for lane
  changes in fully automated driving}.
\newblock In {\it \bibinfo{booktitle}{16th International IEEE Conference on
  Intelligent Transportation Systems (ITSC 2013)}\/} (pp.
  \bibinfo{pages}{2063--2067}).
\newblock \bibinfo{organization}{IEEE}.
\bibitem[{Urmson et~al.(2008)Urmson, Anhalt, Bagnell, Baker, Bittner, Clark,
  Dolan, Duggins, Galatali, Geyer et~al.}]{item9}
\bibinfo{author}{Urmson, C.}, \bibinfo{author}{Anhalt, J.},
  \bibinfo{author}{Bagnell, D.}, \bibinfo{author}{Baker, C.},
  \bibinfo{author}{Bittner, R.}, \bibinfo{author}{Clark, M.},
  \bibinfo{author}{Dolan, J.}, \bibinfo{author}{Duggins, D.},
  \bibinfo{author}{Galatali, T.}, \bibinfo{author}{Geyer, C.} et~al.
  (\bibinfo{year}{2008}).
\newblock \bibinfo{title}{Autonomous driving in urban environments: Boss and
  the urban challenge}.
\newblock {\it \bibinfo{journal}{Journal of Field Robotics}\/},  {\it
  \bibinfo{volume}{25}\/}, \bibinfo{pages}{425--466}.
\bibitem[{Veronese et~al.(2015)Veronese, Aguiar, Nascimento, Guivant, Cheein,
  De~Souza \& Oliveira-Santos}]{item62}
\bibinfo{author}{Veronese, L.}, \bibinfo{author}{Aguiar, E.},
  \bibinfo{author}{Nascimento, R.~C.}, \bibinfo{author}{Guivant, J.},
  \bibinfo{author}{Cheein, F. A.~A.}, \bibinfo{author}{De~Souza, A.~F.}, \&
  \bibinfo{author}{Oliveira-Santos, T.} (\bibinfo{year}{2015}).
\newblock \bibinfo{title}{Re-emission and satellite aerial maps applied to
  vehicle localization on urban environments}.
\newblock In {\it \bibinfo{booktitle}{2015 IEEE/RSJ International Conference on
  Intelligent Robots and Systems (IROS)}\/} (pp. \bibinfo{pages}{4285--4290}).
\newblock \bibinfo{organization}{IEEE}.
\bibitem[{Veronese et~al.(2016)Veronese, Guivant, Cheein, Oliveira-Santos,
  Mutz, de~Aguiar, Badue \& De~Souza}]{item66}
\bibinfo{author}{Veronese, L.}, \bibinfo{author}{Guivant, J.},
  \bibinfo{author}{Cheein, F. A.~A.}, \bibinfo{author}{Oliveira-Santos, T.},
  \bibinfo{author}{Mutz, F.}, \bibinfo{author}{de~Aguiar, E.},
  \bibinfo{author}{Badue, C.}, \& \bibinfo{author}{De~Souza, A.~F.}
  (\bibinfo{year}{2016}).
\newblock \bibinfo{title}{A light-weight yet accurate localization system for
  autonomous cars in large-scale and complex environments}.
\newblock In {\it \bibinfo{booktitle}{2016 IEEE 19th International Conference
  on Intelligent Transportation Systems (ITSC)}\/} (pp.
  \bibinfo{pages}{520--525}).
\newblock \bibinfo{organization}{IEEE}.
\bibitem[{Viswanathan et~al.(2016)Viswanathan, Pires \& Huber}]{item69}
\bibinfo{author}{Viswanathan, A.}, \bibinfo{author}{Pires, B.~R.}, \&
  \bibinfo{author}{Huber, D.} (\bibinfo{year}{2016}).
\newblock \bibinfo{title}{Vision-based robot localization across seasons and in
  remote locations}.
\newblock In {\it \bibinfo{booktitle}{2016 IEEE International Conference on
  Robotics and Automation (ICRA)}\/} (pp. \bibinfo{pages}{4815--4821}).
\newblock \bibinfo{organization}{IEEE}.
\bibitem[{Vivacqua et~al.(2017)Vivacqua, Vassallo \& Martins}]{item74}
\bibinfo{author}{Vivacqua, R.}, \bibinfo{author}{Vassallo, R.}, \&
  \bibinfo{author}{Martins, F.} (\bibinfo{year}{2017}).
\newblock \bibinfo{title}{A low cost sensors approach for accurate vehicle
  localization and autonomous driving application}.
\newblock {\it \bibinfo{journal}{Sensors}\/},  {\it \bibinfo{volume}{17}\/},
  \bibinfo{pages}{2359}.
\bibitem[{Vu \& Aycard(2009)}]{item119}
\bibinfo{author}{Vu, T.-D.}, \& \bibinfo{author}{Aycard, O.}
  (\bibinfo{year}{2009}).
\newblock \bibinfo{title}{Laser-based detection and tracking moving objects
  using data-driven markov chain monte carlo}.
\newblock In {\it \bibinfo{booktitle}{2009 IEEE International Conference on
  Robotics and Automation}\/} (pp. \bibinfo{pages}{3800--3806}).
\newblock \bibinfo{organization}{IEEE}.
\bibitem[{Wang et~al.(2015)Wang, Posner \& Newman}]{item120}
\bibinfo{author}{Wang, D.~Z.}, \bibinfo{author}{Posner, I.}, \&
  \bibinfo{author}{Newman, P.} (\bibinfo{year}{2015}).
\newblock \bibinfo{title}{Model-free detection and tracking of dynamic objects
  with 2d lidar}.
\newblock {\it \bibinfo{journal}{The International Journal of Robotics
  Research}\/},  {\it \bibinfo{volume}{34}\/}, \bibinfo{pages}{1039--1063}.
\bibitem[{Wegner et~al.(2015)Wegner, Montoya-Zegarra \& Schindler}]{item105}
\bibinfo{author}{Wegner, J.~D.}, \bibinfo{author}{Montoya-Zegarra, J.~A.}, \&
  \bibinfo{author}{Schindler, K.} (\bibinfo{year}{2015}).
\newblock \bibinfo{title}{Road networks as collections of minimum cost paths}.
\newblock {\it \bibinfo{journal}{ISPRS Journal of Photogrammetry and Remote
  Sensing}\/},  {\it \bibinfo{volume}{108}\/}, \bibinfo{pages}{128--137}.
\bibitem[{Wei et~al.(2013)Wei, Snider, Kim, Dolan, Rajkumar \&
  Litkouhi}]{item161}
\bibinfo{author}{Wei, J.}, \bibinfo{author}{Snider, J.~M.},
  \bibinfo{author}{Kim, J.}, \bibinfo{author}{Dolan, J.~M.},
  \bibinfo{author}{Rajkumar, R.}, \& \bibinfo{author}{Litkouhi, B.}
  (\bibinfo{year}{2013}).
\newblock \bibinfo{title}{Towards a viable autonomous driving research
  platform}.
\newblock In {\it \bibinfo{booktitle}{2013 IEEE Intelligent Vehicles Symposium
  (IV)}\/} (pp. \bibinfo{pages}{763--770}).
\newblock \bibinfo{organization}{IEEE}.
\bibitem[{Wolcott \& Eustice(2017)}]{item67}
\bibinfo{author}{Wolcott, R.~W.}, \& \bibinfo{author}{Eustice, R.~M.}
  (\bibinfo{year}{2017}).
\newblock \bibinfo{title}{Robust lidar localization using multiresolution
  gaussian mixture maps for autonomous driving}.
\newblock {\it \bibinfo{journal}{The International Journal of Robotics
  Research}\/},  {\it \bibinfo{volume}{36}\/}, \bibinfo{pages}{292--319}.
\bibitem[{Wray et~al.(2017)Wray, Witwicki \& Zilberstein}]{item289}
\bibinfo{author}{Wray, K.~H.}, \bibinfo{author}{Witwicki, S.~J.}, \&
  \bibinfo{author}{Zilberstein, S.} (\bibinfo{year}{2017}).
\newblock \bibinfo{title}{Online decision-making for scalable autonomous
  systems}.
\newblock In {\it \bibinfo{booktitle}{International Joint Conference on
  Artificial Intelligence}\/}.
\bibitem[{Wu \& Ranganathan(2012)}]{item181}
\bibinfo{author}{Wu, T.}, \& \bibinfo{author}{Ranganathan, A.}
  (\bibinfo{year}{2012}).
\newblock \bibinfo{title}{A practical system for road marking detection and
  recognition}.
\newblock In {\it \bibinfo{booktitle}{2012 IEEE Intelligent Vehicles
  Symposium}\/} (pp. \bibinfo{pages}{25--30}).
\newblock \bibinfo{organization}{IEEE}.
\bibitem[{Xin et~al.(2014)Xin, Wang, Zhang \& Zheng}]{item13}
\bibinfo{author}{Xin, J.}, \bibinfo{author}{Wang, C.}, \bibinfo{author}{Zhang,
  Z.}, \& \bibinfo{author}{Zheng, N.} (\bibinfo{year}{2014}).
\newblock \bibinfo{title}{China future challenge: Beyond the intelligent
  vehicle}.
\newblock {\it \bibinfo{journal}{IEEE Intell. Transp. Syst. Soc. Newslett.}\/},
   {\it \bibinfo{volume}{16}\/}, \bibinfo{pages}{8--10}.
\bibitem[{Xu et~al.(2015)Xu, Snider, Wei \& Dolan}]{item131}
\bibinfo{author}{Xu, W.}, \bibinfo{author}{Snider, J.}, \bibinfo{author}{Wei,
  J.}, \& \bibinfo{author}{Dolan, J.~M.} (\bibinfo{year}{2015}).
\newblock \bibinfo{title}{Context-aware tracking of moving objects for distance
  keeping}.
\newblock In {\it \bibinfo{booktitle}{2015 IEEE Intelligent Vehicles Symposium
  (IV)}\/} (pp. \bibinfo{pages}{1380--1385}).
\newblock \bibinfo{organization}{IEEE}.
\bibitem[{Xu et~al.(2012)Xu, Wei, Dolan, Zhao \& Zha}]{item234}
\bibinfo{author}{Xu, W.}, \bibinfo{author}{Wei, J.}, \bibinfo{author}{Dolan,
  J.~M.}, \bibinfo{author}{Zhao, H.}, \& \bibinfo{author}{Zha, H.}
  (\bibinfo{year}{2012}).
\newblock \bibinfo{title}{A real-time motion planner with trajectory
  optimization for autonomous vehicles}.
\newblock In {\it \bibinfo{booktitle}{2012 IEEE International Conference on
  Robotics and Automation}\/} (pp. \bibinfo{pages}{2061--2067}).
\newblock \bibinfo{organization}{IEEE}.
\bibitem[{Xu et~al.(2017)Xu, John, Mita, Tehrani, Ishimaru \& Nishino}]{item68}
\bibinfo{author}{Xu, Y.}, \bibinfo{author}{John, V.}, \bibinfo{author}{Mita,
  S.}, \bibinfo{author}{Tehrani, H.}, \bibinfo{author}{Ishimaru, K.}, \&
  \bibinfo{author}{Nishino, S.} (\bibinfo{year}{2017}).
\newblock \bibinfo{title}{3d point cloud map based vehicle localization using
  stereo camera}.
\newblock In {\it \bibinfo{booktitle}{2017 IEEE Intelligent Vehicles Symposium
  (IV)}\/} (pp. \bibinfo{pages}{487--492}).
\newblock \bibinfo{organization}{IEEE}.
\bibitem[{Xue et~al.(2017)Xue, Wang, Du, Cui, Huang \& Zheng}]{item132}
\bibinfo{author}{Xue, J.-r.}, \bibinfo{author}{Wang, D.}, \bibinfo{author}{Du,
  S.-y.}, \bibinfo{author}{Cui, D.-x.}, \bibinfo{author}{Huang, Y.}, \&
  \bibinfo{author}{Zheng, N.-n.} (\bibinfo{year}{2017}).
\newblock \bibinfo{title}{A vision-centered multi-sensor fusing approach to
  self-localization and obstacle perception for robotic cars}.
\newblock {\it \bibinfo{journal}{Frontiers of Information Technology \&
  Electronic Engineering}\/},  {\it \bibinfo{volume}{18}\/},
  \bibinfo{pages}{122--138}.
\bibitem[{Yenikaya et~al.(2013)Yenikaya, Yenikaya \& D{\"u}ven}]{item107}
\bibinfo{author}{Yenikaya, S.}, \bibinfo{author}{Yenikaya, G.}, \&
  \bibinfo{author}{D{\"u}ven, E.} (\bibinfo{year}{2013}).
\newblock \bibinfo{title}{Keeping the vehicle on the road: A survey on on-road
  lane detection systems}.
\newblock {\it \bibinfo{journal}{ACM Computing Surveys (CSUR)}\/},  {\it
  \bibinfo{volume}{46}\/}, \bibinfo{pages}{1--43}.
\bibitem[{Yoon et~al.(2015)Yoon, Yoon, Lee \& Shim}]{item214}
\bibinfo{author}{Yoon, S.}, \bibinfo{author}{Yoon, S.-E.},
  \bibinfo{author}{Lee, U.}, \& \bibinfo{author}{Shim, D.~H.}
  (\bibinfo{year}{2015}).
\newblock \bibinfo{title}{Recursive path planning using reduced states for
  car-like vehicles on grid maps}.
\newblock {\it \bibinfo{journal}{IEEE Transactions on Intelligent
  Transportation Systems}\/},  {\it \bibinfo{volume}{16}\/},
  \bibinfo{pages}{2797--2813}.
\bibitem[{Yu et~al.(2018)Yu, Xian, Chen, Liu, Liao, Madhavan \&
  Darrell}]{item158}
\bibinfo{author}{Yu, F.}, \bibinfo{author}{Xian, W.}, \bibinfo{author}{Chen,
  Y.}, \bibinfo{author}{Liu, F.}, \bibinfo{author}{Liao, M.},
  \bibinfo{author}{Madhavan, V.}, \& \bibinfo{author}{Darrell, T.}
  (\bibinfo{year}{2018}).
\newblock \bibinfo{title}{Bdd100k: A diverse driving video database with
  scalable annotation tooling}.
\newblock {\it \bibinfo{journal}{arXiv preprint arXiv:1805.04687}\/}, .
\bibitem[{Zhang et~al.(2013)Zhang, Li, Li, Mao \& N{\"u}chter}]{item115}
\bibinfo{author}{Zhang, L.}, \bibinfo{author}{Li, Q.}, \bibinfo{author}{Li,
  M.}, \bibinfo{author}{Mao, Q.}, \& \bibinfo{author}{N{\"u}chter, A.}
  (\bibinfo{year}{2013}).
\newblock \bibinfo{title}{Multiple vehicle-like target tracking based on the
  velodyne lidar}.
\newblock {\it \bibinfo{journal}{IFAC Proceedings Volumes}\/},  {\it
  \bibinfo{volume}{46}\/}, \bibinfo{pages}{126--131}.
\bibitem[{Zhang et~al.(2014)Zhang, Xue, Zhang, Zhang \& Zheng}]{item144}
\bibinfo{author}{Zhang, Y.}, \bibinfo{author}{Xue, J.}, \bibinfo{author}{Zhang,
  G.}, \bibinfo{author}{Zhang, Y.}, \& \bibinfo{author}{Zheng, N.}
  (\bibinfo{year}{2014}).
\newblock \bibinfo{title}{A multi-feature fusion based traffic light
  recognition algorithm for intelligent vehicles}.
\newblock In {\it \bibinfo{booktitle}{Proceedings of the 33rd Chinese Control
  Conference}\/} (pp. \bibinfo{pages}{4924--4929}).
\newblock \bibinfo{organization}{IEEE}.
\bibitem[{Zhao et~al.(2017)Zhao, Ichise, Liu, Mita \& Sasaki}]{item223}
\bibinfo{author}{Zhao, L.}, \bibinfo{author}{Ichise, R.}, \bibinfo{author}{Liu,
  Z.}, \bibinfo{author}{Mita, S.}, \& \bibinfo{author}{Sasaki, Y.}
  (\bibinfo{year}{2017}).
\newblock \bibinfo{title}{Ontology-based driving decision making: A feasibility
  study at uncontrolled intersections}.
\newblock {\it \bibinfo{journal}{IEICE TRANSACTIONS on Information and
  Systems}\/},  {\it \bibinfo{volume}{100}\/}, \bibinfo{pages}{1425--1439}.
\bibitem[{Zhao et~al.(2015)Zhao, Ichise, Yoshikawa, Naito, Kakinami \&
  Sasaki}]{item222}
\bibinfo{author}{Zhao, L.}, \bibinfo{author}{Ichise, R.},
  \bibinfo{author}{Yoshikawa, T.}, \bibinfo{author}{Naito, T.},
  \bibinfo{author}{Kakinami, T.}, \& \bibinfo{author}{Sasaki, Y.}
  (\bibinfo{year}{2015}).
\newblock \bibinfo{title}{Ontology-based decision making on uncontrolled
  intersections and narrow roads}.
\newblock In {\it \bibinfo{booktitle}{2015 IEEE intelligent vehicles symposium
  (IV)}\/} (pp. \bibinfo{pages}{83--88}).
\newblock \bibinfo{organization}{IEEE}.
\bibitem[{Zhao et~al.(2012)Zhao, Chen, Song, Tao, Xu \& Mei}]{item253}
\bibinfo{author}{Zhao, P.}, \bibinfo{author}{Chen, J.}, \bibinfo{author}{Song,
  Y.}, \bibinfo{author}{Tao, X.}, \bibinfo{author}{Xu, T.}, \&
  \bibinfo{author}{Mei, T.} (\bibinfo{year}{2012}).
\newblock \bibinfo{title}{Design of a control system for an autonomous vehicle
  based on adaptive-pid}.
\newblock {\it \bibinfo{journal}{International Journal of Advanced Robotic
  Systems}\/},  {\it \bibinfo{volume}{9}\/}, \bibinfo{pages}{44}.
\bibitem[{Zhu et~al.(2016)Zhu, Liang, Zhang, Huang, Li \& Hu}]{item174}
\bibinfo{author}{Zhu, Z.}, \bibinfo{author}{Liang, D.}, \bibinfo{author}{Zhang,
  S.}, \bibinfo{author}{Huang, X.}, \bibinfo{author}{Li, B.}, \&
  \bibinfo{author}{Hu, S.} (\bibinfo{year}{2016}).
\newblock \bibinfo{title}{Traffic-sign detection and classification in the
  wild}.
\newblock In {\it \bibinfo{booktitle}{Proceedings of the IEEE Conference on
  Computer Vision and Pattern Recognition}\/} (pp.
  \bibinfo{pages}{2110--2118}).
\bibitem[{Ziegler et~al.(2014{\natexlab{a}})Ziegler, Bender, Dang \&
  Stiller}]{item245}
\bibinfo{author}{Ziegler, J.}, \bibinfo{author}{Bender, P.},
  \bibinfo{author}{Dang, T.}, \& \bibinfo{author}{Stiller, C.}
  (\bibinfo{year}{2014}{\natexlab{a}}).
\newblock \bibinfo{title}{Trajectory planning for bertha—a local, continuous
  method}.
\newblock In {\it \bibinfo{booktitle}{2014 IEEE Intelligent Vehicles
  Symposium}\/} (pp. \bibinfo{pages}{450--457}).
\newblock \bibinfo{organization}{IEEE}.
\bibitem[{Ziegler et~al.(2014{\natexlab{b}})Ziegler, Bender, Schreiber,
  Lategahn, Strauss, Stiller, Dang, Franke, Appenrodt, Keller et~al.}]{item122}
\bibinfo{author}{Ziegler, J.}, \bibinfo{author}{Bender, P.},
  \bibinfo{author}{Schreiber, M.}, \bibinfo{author}{Lategahn, H.},
  \bibinfo{author}{Strauss, T.}, \bibinfo{author}{Stiller, C.},
  \bibinfo{author}{Dang, T.}, \bibinfo{author}{Franke, U.},
  \bibinfo{author}{Appenrodt, N.}, \bibinfo{author}{Keller, C.~G.} et~al.
  (\bibinfo{year}{2014}{\natexlab{b}}).
\newblock \bibinfo{title}{Making bertha drive—an autonomous journey on a
  historic route}.
\newblock {\it \bibinfo{journal}{IEEE Intelligent transportation systems
  magazine}\/},  {\it \bibinfo{volume}{6}\/}, \bibinfo{pages}{8--20}.
\bibitem[{Ziegler et~al.(2014{\natexlab{c}})Ziegler, Lategahn, Schreiber,
  Keller, Kn{\"o}ppel, Hipp, Haueis \& Stiller}]{item71}
\bibinfo{author}{Ziegler, J.}, \bibinfo{author}{Lategahn, H.},
  \bibinfo{author}{Schreiber, M.}, \bibinfo{author}{Keller, C.~G.},
  \bibinfo{author}{Kn{\"o}ppel, C.}, \bibinfo{author}{Hipp, J.},
  \bibinfo{author}{Haueis, M.}, \& \bibinfo{author}{Stiller, C.}
  (\bibinfo{year}{2014}{\natexlab{c}}).
\newblock \bibinfo{title}{Video based localization for bertha}.
\newblock In {\it \bibinfo{booktitle}{2014 IEEE Intelligent Vehicles Symposium
  Proceedings}\/} (pp. \bibinfo{pages}{1231--1238}).
\newblock \bibinfo{organization}{IEEE}.
\bibitem[{Ziegler et~al.(2008)Ziegler, Werling \& Schroder}]{item211}
\bibinfo{author}{Ziegler, J.}, \bibinfo{author}{Werling, M.}, \&
  \bibinfo{author}{Schroder, J.} (\bibinfo{year}{2008}).
\newblock \bibinfo{title}{Navigating car-like robots in unstructured
  environments using an obstacle sensitive cost function}.
\newblock In {\it \bibinfo{booktitle}{2008 IEEE Intelligent Vehicles
  Symposium}\/} (pp. \bibinfo{pages}{787--791}).
\newblock \bibinfo{organization}{IEEE}.
\bibitem[{Zoller(2018)}]{item104}
\bibinfo{author}{Zoller, E.} (\bibinfo{year}{2018}).
\newblock {\it \bibinfo{title}{Location Platform Index: Mapping and Navigation,
  1H18: Key vendor rankings and market trends}\/}.
\newblock \bibinfo{type}{Technical Report} Ovum - TMT Intelligence - Informa.

\end{thebibliography}

\end{document}